\documentclass[final,times]{elsarticle}
\usepackage[margin=2.35cm]{geometry}
\usepackage{graphicx}
\usepackage{multirow}
\usepackage{amsmath,amsfonts,amssymb}
\usepackage{hyperref}
\biboptions{sort&compress}
\usepackage{color}
\usepackage{makecell}
\usepackage{pifont}
\newcommand{\cmark}{\ding{51}}
\newcommand{\xmark}{\ding{55}}
\usepackage[export]{adjustbox}
\usepackage{acronym}

\hypersetup{
    colorlinks=true,
    allcolors={blue},
    pdftitle={Advanced computer vision for extracting georeferenced vehicle trajectories from drone imagery},
    pdfauthor={Robert Fonod; Haechan Cho; Hwasoo Yeo; Nikolas Geroliminis},
    pdfkeywords={Drones; UAVs; Traffic Monitoring; Machine Vision; Computer Vision; Vehicle Detection; Vehicle Tracking; Video Image Processing; Urban Traffic Dataset; Spatio-temporal Sensing}
}

\AtBeginDocument{}
\AtBeginDocument{}
\AtBeginDocument{}
\AtBeginDocument{}
\AtBeginDocument{}

\acrodef{A-KAZE}{accelerated-KAZE}
\acrodef{AMP}{automatic mixed precision}
\acrodef{AV}{autonomous vehicle}
\acrodef{BB}{bounding box}
\acrodefplural{BB}[BBs]{bounding boxes}
\acrodef{BF}{brute force}
\acrodef{BEV}{bird's-eye view}
\acrodef{BoT-SORT}{``bag of tricks'' SORT}
\acrodef{BRISK}{binary robust invariant scalable keypoints}
\acrodef{CLAHE}{contrast limited adaptive histogram equalization}
\acrodef{COTS}{commercial-off-the-shelf}
\acrodef{CS}{coordinate system}
\acrodef{CSV}{comma-separated values}
\acrodef{CV}{computer vision}
\acrodef{DaaS}{Drone as a Service}
\acrodef{DETR}{transformer-based detector}
\acrodef{DFL}{direction field loss}
\acrodef{DL}{deep learning}
\acrodef{EMC}{ego motion compensation}
\acrodef{FLANN}{fast library for approximate nearest neighbors}
\acrodef{FoV}{field of view}
\acrodef{FPS}{frames per second}
\acrodef{GNSS}{global navigation satellite system}
\acrodef{GCP}{ground control point}
\acrodef{GIS}{geographic information system}
\acrodef{GT}{ground truth}
\acrodef{GSD}{ground sample distance}
\acrodef{HEA}{homography estimation accuracy}
\acrodef{ID}{identifier}
\acrodef{IoU}{intersection over union}
\acrodef{ITS}{intelligent transportation systems}
\acrodef{KLT}{Kanade-Lucas-Tomasi}
\acrodef{mAA}{mean Average Accuracy}
\acrodef{mAP}{mean Average Precision}
\acrodef{MIoU}{mean IoU}
\acrodef{MOT}{multi-object tracking}
\acrodef{NMS}{non-maximum suppression}
\acrodef{OC-SORT}{observation-centric SORT}
\acrodef{OF}{optical flow}
\acrodef{ORB}{oriented FAST and rotated BRIEF}
\acrodef{RANSAC}{random sample consensus}
\acrodef{RT-DETR}{real-time DETR}
\acrodef{R-CNN}{region-based convolutional neural network}
\acrodef{RoI}{region of interest}
\acrodef{RSIFT}{root SIFT}
\acrodef{RTK}{real-time kinematic}
\acrodef{RTK-GNSS}{real-time kinematic GNSS}
\acrodef{RTS}{Rauch-Tung-Striebel}
\acrodef{SAHI}{Sliced Aided Hyper Inference}
\acrodef{SCIGC}{Stanford Center at the Incheon Global Campus}
\acrodef{SDP}{separate detection and tracking}
\acrodef{SGD}{stochastic gradient descent}
\acrodef{SSD}{single shot detector}
\acrodef{SIFT}{scale-invariant feature transform}
\acrodef{SNN}{second nearest neighbors}
\acrodef{SORT}{simple online and realtime tracking}
\acrodef{SOTA}{state-of-the-art}
\acrodef{ST}{spatio-temporal}
\acrodef{UAV}{unmanned aerial vehicle}
\acrodef{UHD}{ultra-high-definition}
\acrodef{YOLO}{you only look once}

\makeatletter
\def\ps@pprintTitle{%
 \let\@oddhead\@empty
 \let\@evenhead\@empty
 \let\@oddfoot\@empty
 \let\@evenfoot\@empty
}
\makeatother

\begin{document}
\begin{frontmatter}

\title{Advanced computer vision for extracting georeferenced vehicle trajectories from drone imagery\tnoteref{pub}}

\author[EPFL]{Robert Fonod\corref{cor1}}\ead{robert.fonod@ieee.org}
\author[KAIST]{Haechan Cho}\ead{gkqkemwh@kaist.ac.kr}
\author[KAIST]{Hwasoo Yeo}\ead{hwasoo@kaist.ac.kr}
\author[EPFL]{Nikolas Geroliminis}\ead{nikolas.geroliminis@epfl.ch}

\affiliation[EPFL]{organization={School of Architecture, Civil and Environmental Engineering, École Polytechnique Fédérale de\\ Lausanne~(EPFL)},
            city={Lausanne},
            postcode={CH-1015},
            country={Switzerland}}

\affiliation[KAIST]{organization={Department of Civil and Environmental Engineering, Korea Advanced Institute of Science and\\ Technology~(KAIST)},
            city={Daejeon},
            postcode={34141},
            country={South~Korea}}

\tnotetext[pub]{This is the accepted version of the manuscript, before typesetting and final editorial corrections. The final version is published in \textit{Transportation Research Part C: Emerging Technologies} and is available at \url{https://doi.org/10.1016/j.trc.2025.105205}.}
\cortext[cor1]{Corresponding author.}

\begin{abstract}
This paper presents a comprehensive framework for extracting georeferenced vehicle trajectories from high-altitude drone imagery, addressing key challenges in urban traffic monitoring and the limitations of traditional ground-based systems. Our approach integrates several novel contributions, including a tailored object detector optimized for high-altitude bird’s-eye view perspectives, a unique track stabilization method that uses detected vehicle bounding boxes as exclusion masks during image registration, and an orthophoto and master frame-based georeferencing strategy that enhances consistent alignment across multiple drone viewpoints. Additionally, our framework features robust vehicle dimension estimation and detailed road segmentation, enabling comprehensive traffic dynamics analysis. Conducted in the Songdo International Business District, South Korea, the study utilized a multi-drone experiment covering 20 intersections, capturing approximately 12TB of ultra-high-definition video data over four days. The framework produced two high-quality datasets: the Songdo Traffic dataset, comprising approximately 700,000 unique vehicle trajectories, and the Songdo Vision dataset, containing over 5,000 human-annotated images with about 300,000 vehicle instances categorized into four classes. Comparisons with high-precision sensor data from an instrumented probe vehicle highlight the accuracy and consistency of our extraction pipeline in dense urban environments. The public release of the Songdo Traffic and Songdo Vision datasets, along with the complete source code for the extraction pipeline, establishes new benchmarks in data quality, reproducibility, and scalability in traffic research.  The results demonstrate the potential of integrating drone technology with advanced computer vision methods for precise and cost-effective urban traffic monitoring, providing valuable resources for developing intelligent transportation systems and enhancing traffic management strategies.
\end{abstract}


\end{frontmatter}

\section{Introduction}\label{sec:introduction}
Traditional traffic monitoring methods, such as loop detectors and manual counting, are becoming obsolete in the era of smart cities due to their inflexibility and limited scope~\cite{zhang2018network}. Stationary camera surveillance, although effective, presents economic challenges due to high installation costs and limited \ac{FoV}, leading to blind spots. Data from \ac{GNSS} and connected vehicles also lack the versatility needed for comprehensive traffic monitoring. Several vehicle detection and tracking technologies have emerged, including camera-based~\cite{sivaraman2013looking, yang2018vehicle, ghahremannezhad2023object}, LiDAR-based~\cite{li20173d, zhang2019vehicle}, magnetics-based~\cite{sifuentes2011wireless, wang2017roadside}, and radar-based~\cite{fang2007low, park2003novel} systems. Although each offers certain advantages, they still face limitations in adaptability and coverage. Accurate estimation of traffic states allows adjustments in traffic control strategies~\cite{seo2017traffic}, improving network utilization~\cite{kouvelas2023linear, tsitsokas2023two} and road user experiences, as well as delivering economic and ecological benefits.

The integration of \acp{UAV}, commonly known as drones, and \ac{CV} techniques presents a promising solution to these challenges. Originally developed for military purposes~\cite{hassanalian2017classifications}, \acp{UAV} have since found applications in various civilian domains, including urban traffic monitoring and management~\cite{khan2017unmanned, outay2020applications, espadaler2023continuous}. Advances in drone technology, combined with sophisticated \ac{CV} tools, enable the extraction of detailed vehicle trajectories from high-altitude drone footage, facilitating \acl{ST} analyses of dynamic entities such as vehicles, pedestrians, and cyclists~\cite{Goh2022Involvement, kim2020pedestrian, mahajan2023treating}. 
Compared to traditional sensors, these trajectories offer new opportunities to study complex traffic phenomena, such as large-scale network modeling~\cite{paipuri2021empirical}, on-street parking~\cite{kim2024monitoring}, as well as emission~\cite{barmpounakis2021empirical} and noise~\cite{espadaler2023traffic} estimation. This aerial perspective, free from the constraints of fixed infrastructure, provides an adaptable and cost-effective means of observing urban mobility patterns in congested environments, thus supporting responsive traffic management strategies in smart cities~\cite{jian2019combining, khan2020smart, barmpounakis2020new}.

\begin{figure}[htbp]
  \centering
  \includegraphics[height=0.34\columnwidth, valign=t]{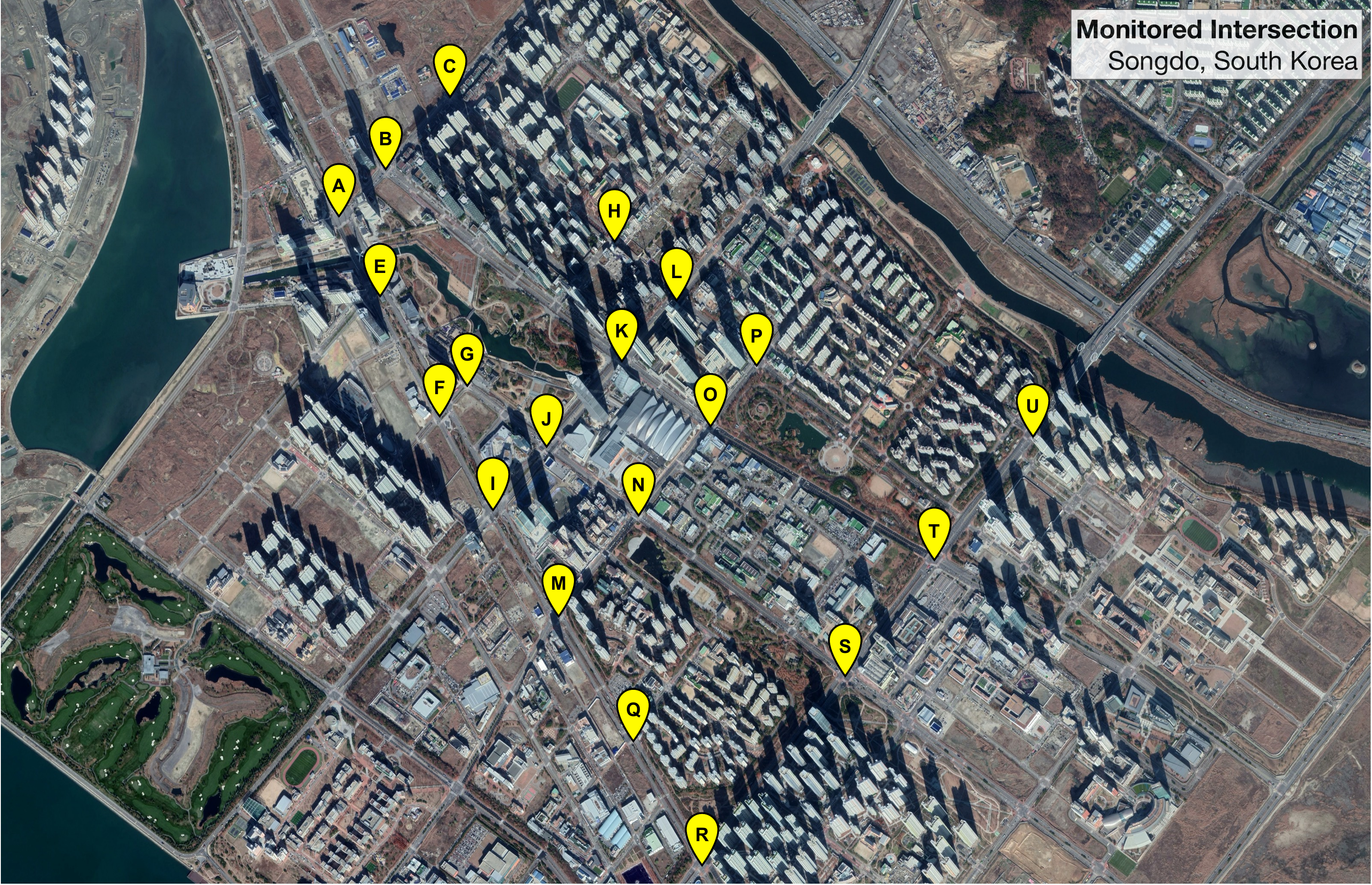}
  \includegraphics[height=0.34\columnwidth, valign=t]{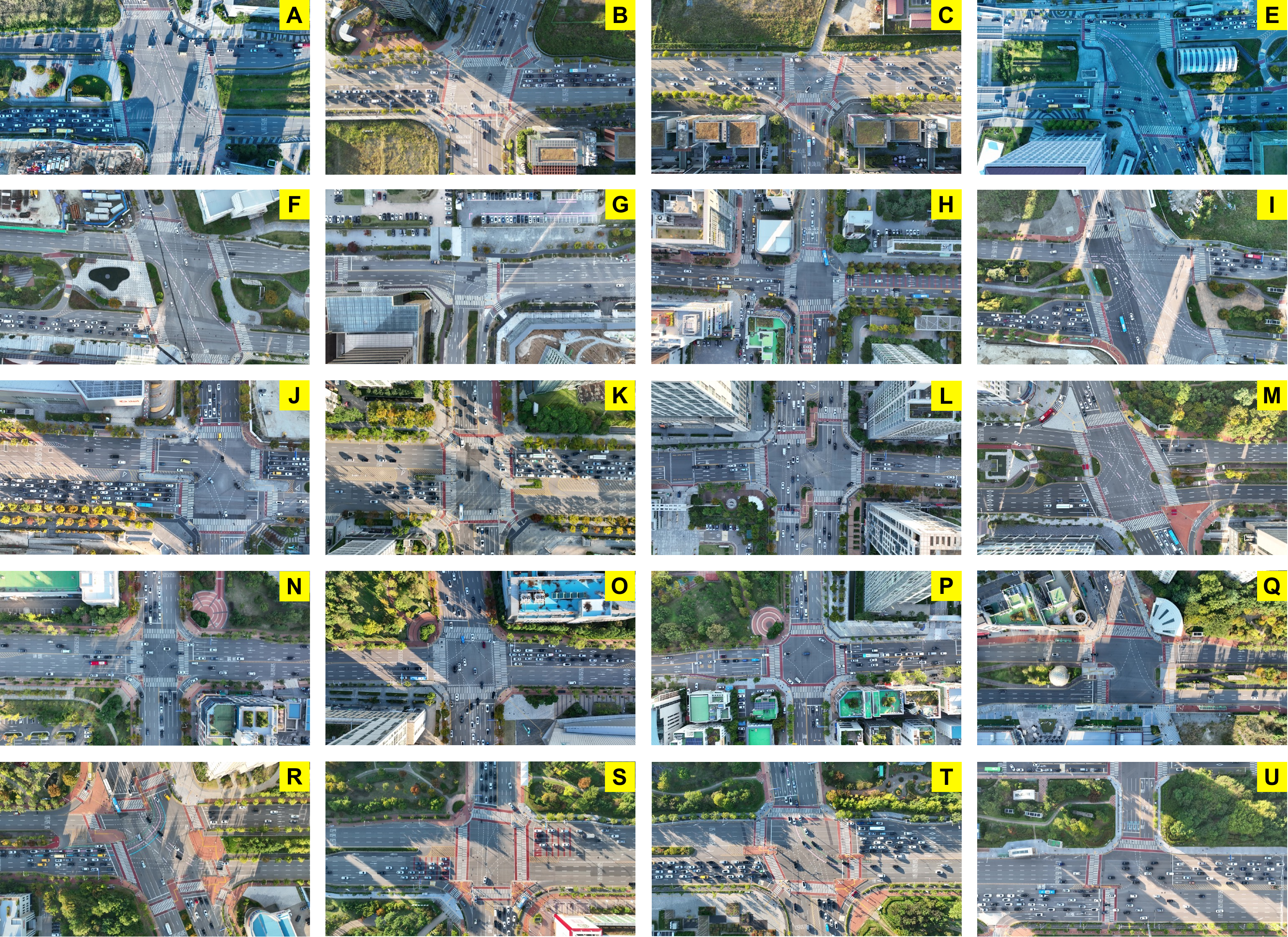}
  \caption{Left: Map of the Songdo International Business District, indicating the locations of the 20 monitored intersections labeled A to U (excluding~D, which denotes the drones). Right: Images of the 20 intersections as seen by our DJI Mavic 3 drones at 4K resolution.}
  \label{fig:intersections}
\end{figure}

In a collaborative effort between KAIST and EPFL, a unique multi-drone experiment was conducted in the Songdo International Business District, South Korea, from October 4 to 7, 2022. The first planned day, October 3, was excluded due to severe weather conditions, including strong winds exceeding the drones’ operational limits, rain, and fog, which disrupted flight operations. This adjustment ensured stable drone performance and consistent data quality. A fleet of 10 \acl{COTS} drones, operated by certified pilots, synchronized their operations to monitor 20 busy intersections, as depicted in \autoref{fig:intersections}. Advanced flight plans, which combine the hover and transition phases, were executed to optimize coverage. The drones maintained an altitude of 150 meters, some reduced to 140 meters to mitigate collision risks, ensuring compliance with South Korean UAV regulations. Adopting a \ac{BEV} perspective, the drones captured footage at 4K \ac{UHD} resolution ($3{,}840 \times 2{,}160$ pixels) at a smooth frame rate of 29.97 \ac{FPS}. Data collection was structured into 10 daily flight sessions, with five morning sessions (AM1–AM5) and five afternoon sessions (PM1–PM5), capturing traffic flow during peak morning and afternoon hours. Each synchronized flight session lasted approximately 30 minutes, followed by a brief battery replacement, bringing the total cycle time to exactly 40 minutes. Over the course of the experiment, 400 flights were conducted, generating 12TB of raw video data. Thirteen \ac{RTK}-calibrated \acp{GCP} were strategically placed throughout the study area as reference points for a dedicated drone, which captured overlapping images at an altitude of 75 meters. These images produced a high-resolution orthophoto that serves as a geometrically accurate reference for georeferencing the extracted trajectories and mapping.

This work introduces a comprehensive end-to-end trajectory extraction pipeline that employs advanced \ac{CV} algorithms to derive detailed traffic data from high-altitude drone footage. Although a detailed experiment design, regulatory considerations, and an in-depth analysis of the extracted data are outside this paper’s scope, our contributions significantly expand the resources available to the research community. Specifically, we introduce the \textit{Songdo Traffic} dataset~\cite{songdo_traffic_dataset}, comprising approximately 700,000 unique vehicle trajectories, making it one of the largest public traffic datasets collected from a smart city worldwide. This extensive collection provides unparalleled insights into urban mobility, serving as a valuable resource for traffic management and research. The dataset includes trajectory data at a high frequency of 29.97 points per second, complemented by various metadata. Songdo Traffic encompasses vehicle trajectories in multiple \acp{CS}, including global geographic, local Cartesian, and orthophoto \ac{CS}. Each trajectory is further annotated with metadata such as vehicle \ac{ID}, type, and, when feasible, vehicle dimension estimates. Additionally, each trajectory point is tagged with a precise ISO-formatted timestamp, along with instantaneous speed and acceleration estimates, road section and lane number information, and a visibility flag, where available.

Complementing Songdo Traffic, the \textit{Songdo Vision} dataset~\cite{songdo_vision_dataset} addresses the scarcity of detailed vehicle annotations necessary for training object detection models, particularly from high-altitude \ac{BEV} aerial perspectives, where smaller vehicles, like motorcycles, are often less discernible. This dataset contains over 5,000 human-annotated video frames, featuring nearly 300,000 vehicle instances across four distinct classes: cars (including vans and light-duty vehicles), buses, trucks, and motorcycles. To facilitate broader use, annotations are provided in multiple formats compatible with various \ac{DL} models, making the Songdo Vision dataset a valuable resource for improving and validating vehicle detection algorithms. In line with our commitment to open science, we have publicly released these datasets~\cite{songdo_traffic_dataset, songdo_vision_dataset} alongside the source codes~\cite{Fonod_Geo-trax_2025, Fonod_Stabilo_2024, Fonod_Stabilo_Optimize_2025} for our extraction and stabilization routines. These resources are expected to set new benchmarks for data quality and reproducibility in traffic research leveraging \ac{CV} and \ac{DL} techniques.

Our extraction methodology introduces several innovative techniques that significantly enhance the accuracy and reliability of vehicle trajectory data from drone imagery. We developed a tailored YOLOv8-based object detector optimized for detecting small vehicles at high-altitude \ac{BEV} perspectives. To address the challenges of drone movements, we proposed a novel track stabilization approach that applies homographic transformations directly to extracted tracks rather than raw video frames, allowing for the effective use of exclusion masks based on detected vehicle bounding boxes. This approach reduces both storage and computational overhead compared to full-frame video stabilization and is further optimized through a newly proposed metric designed to evaluate object-level stabilization accuracy. Additionally, we introduced a robust vehicle dimension estimation method that uses azimuth-based filtering and ratio-based criteria to ensure accurate measurements under varying conditions. Our methodology further includes speed and acceleration computations, incorporating visibility filtering and Gaussian smoothing to derive realistic traffic dynamics. To enhance consistency and minimize matching errors during georeferencing, we propose a novel master frame approach that serves as an intermediary \ac{CS}, ensuring uniform transformation accuracy across multiple drone viewpoints. Advanced georeferencing techniques, employing a high-resolution orthophoto enhanced with 13~\aclp{GCP}, further ensure precise mapping of trajectories into real-world coordinates. We optimized the stabilization routine and image registration through an extensive parameter search, resulting in highly accurate geospatial data for traffic analysis. 

During the Songdo experiment, various traffic events, including severe congestion, excessive speeds, and post-accident traffic dynamics, were captured, along with data from an instrumented \ac{AV} provided by \ac{SCIGC}, which was equipped with high-precision \ac{RTK-GNSS} sensors. Although the collision itself occurred during a battery swap and was not recorded, the prolonged presence of the involved vehicles enabled us to document the resulting traffic disruptions over time. The \ac{AV} data provided a valuable resource for comparison with our drone-derived trajectories and speed profiles, demonstrating very high consistency between the two measurement methods.  Additionally, our methodology was subjected to rigorous hyperparameter tuning for track stabilization and comprehensive benchmarks to evaluate image registration and georeferencing accuracy. For vehicle dimension estimation, the results showed close agreement with the known dimensions of specific vehicles, such as the \ac{AV} and certain bus types, confirming the method’s ability to produce realistic distributions.

By addressing key challenges in drone-based traffic monitoring through \ac{CV}, as discussed in \autoref{sec:related_work}, this research presents a comprehensive end-to-end framework for extracting georeferenced trajectories from high-altitude drone footage. We enhance detection and tracking algorithms to handle unique perspectives and dynamic conditions, surpassing traditional ground-based methods. Our innovative track stabilization, accurate vehicle dimension estimation, speed and acceleration computations, and precise georeferencing contribute to the extraction of high-quality trajectory data. Comparison with \ac{AV} data shows that our methods produce realistic speed, acceleration, and dimension distributions. By disseminating these methods through open-source datasets and code, we establish a new resource for data quality, facilitating more responsive, scalable, and cost-effective traffic management solutions in smart cities.

\section{Related work}\label{sec:related_work}

This section reviews the \ac{SOTA} in drone-based traffic data extraction, including object detection, tracking, video stabilization, georeferencing, and dataset availability. Traditional \ac{CV} methods for fixed cameras, such as background subtraction~\cite{mandellos2011background}, \ac{OF}~\cite{gokul2023lucas}, template matching~\cite{choi2006vehicle}, feature matching~\cite{coifman1998real}, and histogram-based methods~\cite{cao2011vehicle, wu2013relative}, face challenges in complex environments~\cite{khan2018unmanned} and under varying illumination conditions~\cite{kim2017illumination, zhang2019longitudinal, ghahremannezhad2023object}.

\subsection{Object detection}

The rise of \ac{DL} techniques for \ac{CV} has significantly improved vehicle detection. Leading architectures include \ac{SSD}~\cite{liu2016ssd}, \ac{R-CNN}~\cite{girshick2014rich}, and \ac{YOLO}~\cite{redmon2016you}. \ac{R-CNN} variants, such as Faster \ac{R-CNN}~\cite{ren2017faster}, have been used for vehicle detection in various studies~\cite{hsu2018vehicle, zhan2019interaction}. \Ac{YOLO} models (v3–v10)~\cite{redmon2018yolov3, bochkovskiy2020yolov4, jocher2022ultralytics, li2022yolov6, wang2023yolov7, yolov8_ultralytics, wang2024yolov9, wang2024yolov10} have also been adapted for accuracy in traffic surveillance~\cite{mao2020finding, zhang2022real, espadaler2023continuous}.

Despite their effectiveness, these models often rely on \ac{NMS}, a post-processing technique that removes redundant \acp{BB} by retaining only the one with the highest confidence score for each detected object. Recent \acp{DETR}\cite{carion2020end, zhu2020deformable, liu2021swin, zhao2023detrs} aim to eliminate \ac{NMS} but struggle with small object detection and require high computational resources. Specialized detectors, such as \ac{YOLO} for occluded vehicles\cite{he2024enhancing} and the Butterfly detector~\cite{adaimi2023traffic}, address challenges specific to aerial perspectives. To improve small vehicle detection in drone footage, Novikov et al.~\cite{novikov2024vehicle} incorporated the computationally expensive \ac{SAHI} technique \cite{akyon2022slicing} into their YOLOv8-based detector, enabling enhanced detection capabilities in high-resolution aerial imagery. Integration of multimodal data, like combining RGB with LiDAR \cite{wu2024fine} or thermal \cite{ha2017mfnet} imaging, has also shown promise in enhancing detection in complex urban environments.

\subsection{Object tracking}

Accurate tracking is essential for reliable movement data. Traditional methods~\cite{yilmaz2006object}, like Kalman filtering~\cite{kalman1960new} and the Hungarian algorithm~\cite{Kuhn1955Hungarian}, have been widely used in traffic analysis~\cite{hsieh2006automatic, kim2010evaluation, zhan2019interaction, chen2020conflict}. However, these approaches often struggle with drone-captured footage complexities, especially in dynamic urban environments where occlusions, varying object sizes, and diverse motion patterns pose significant challenges.

Current challenges in aerial tracking include persistent track loss in occluded areas, object similarity, high computational costs, and the additional complexity of camera-induced motion. These issues are particularly pronounced in dense urban environments. To address the challenge of camera-induced motion, \ac{EMC} emerges as an effective solution by isolating object motion, thereby enhancing tracking accuracy in dynamic drone-captured scenarios. While benchmarks like MOTChallenge~\cite{milan2016mot16, dendorfer2020mot20, dendorfer2021motchallenge} have advanced pedestrian tracking, vehicle tracking in \ac{BEV} contexts remains underrepresented.

Newer \ac{MOT} algorithms, such as \ac{SORT}~\cite{bewley2016simple}, \ac{OC-SORT}~\cite{Cao22}, ByteTrack~\cite{zhang2022bytetrack}, and \ac{BoT-SORT}~\cite{aharon2022bot}, follow the ``separate detection and tracking'' paradigm. Hybrid models, including DeepSORT~\cite{wojke2017simple}, StrongSORT~\cite{du2023strongsort}, and \ac{BoT-SORT}-ReID~\cite{aharon2022bot}, incorporate appearance information to reduce identity switches and handle occlusions more effectively. End-to-end tracking approaches like FairMOT~\cite{zhang2021fairmot} combine detection and tracking but face challenges in \ac{BEV} perspectives due to limited \acl{GT} data and high computational demands. Moreover, FairMOT’s joint detection-ReID architecture requires extensive domain-specific fine-tuning, which is impractical given the lack of annotated aerial tracking data.

\subsection{Stabilization}

Unlike larger drones equipped with high-performance gimbals, video stabilization is crucial for accurate traffic data extraction with smaller commercial drones that are more susceptible to environmental disturbances like wind~\cite{mingkhwan2017digital, barmpounakis2019accurate}. Stabilization, a form of video registration \cite{kumar1998registration}, involves aligning consecutive frames by matching keypoints to correct unwanted motion. Various keypoint detectors such as \ac{SIFT}~\cite{lowe2004distinctive}, \ac{RSIFT}~\cite{arandjelovic2012three}, \ac{BRISK}~\cite{leutenegger2011brisk}, KAZE~\cite{alcantarilla2012kaze}, \ac{A-KAZE}~\cite{alcantarilla2011fast}, and \ac{ORB}~\cite{rublee2011orb} are commonly used. \Ac{OF} methods, like the~\ac{KLT} tracker~\cite{shi1994good}, are faster but can be less reliable~\cite{ke2016real}. Recent \ac{DL}-based methods such as SuperPoint~\cite{detone2018superpoint} and SuperGlue~\cite{sarlin2020superglue} enhance keypoint detection and matching, although they often require fine-tuning with \acl{GT} data.

Keyframe correspondences established through these features allow for the computation of either perspective or affine transformations using the least squares method, optimized by variants of the \ac{RANSAC} algorithm~\cite{fischler1981random} for outlier rejection. Feature matching algorithms, such as \ac{BF} or \ac{FLANN}~\cite{muja2009fast}, compute similarity scores between extracted keypoints to form correspondences. Preprocessing steps, like grayscale conversion and \ac{CLAHE}, are implemented to enhance keypoint detection in varying lighting conditions.

Various methods can be combined to align video frames, allowing for smoother vehicle tracking. For instance,~\cite{mingkhwan2017digital} utilized speeded-up robust features (SURF) features for keypoint detection to ensure smooth traffic analysis. However, the main challenge in stabilization is achieving an optimal balance between speed and accuracy while mitigating distortions caused by moving objects.

\subsection{Georeferencing}\label{sec:related_work:georeferencing}

Georeferencing, in the context of our study, is the process of transforming vehicle trajectories from image coordinates into real-world geographic coordinates (e.g., latitude and longitude). This process is essential for integrating trajectory data with other geospatial datasets, enabling accurate traffic analysis and infrastructure assessment. Accurate georeferencing often requires a precise georeferenced image of the area~\cite{kumar1998registration, kumar2001aerial, barber2006vision}. An orthophoto is a geometrically corrected (orthorectified) image that eliminates distortions caused by camera tilt, lens distortion, and topographic relief, producing a uniform scale throughout the image. Unlike standard aerial photographs, orthophotos allow for precise measurements of distances, angles, and areas directly on the image, making them indispensable for traffic monitoring and urban planning~\cite{toutin2004review}. Nonetheless, previous studies often omit meticulous consideration of accurate georeferencing, relying on simplistic methods based on \ac{GSD} for estimating only a subset of traffic parameters, such as vehicle speed and location derived from drone’s location and \ac{GSD}-scaled image coordinates. However, these methods lack the accuracy required for traffic monitoring, particularly since \acl{COTS} drones often exhibit positional errors greater than claimed~\cite{ekaso2020accuracy}.

More sophisticated georeferencing involves the use of \ac{GIS} software~\cite{espadaler2023continuous} for manual alignment of imagery, which, while more accurate than basic \ac{GSD} methods, is time-consuming and demands significant expertise. Alternatively, \ac{RTK-GNSS} can significantly enhance accuracy by refining both the \acp{GCP} placement and the drone’s position during image capture. By providing centimeter-level precision through corrections from a nearby base station, \ac{RTK-GNSS} facilitates the generation of highly accurate orthophotos for the surveyed area. Moreover, integrating these orthophotos with automatic image matching not only reduces the need for manual interventions but also addresses the limitations of \ac{GSD} and manual \ac{GIS}-based methods, enabling more reliable mapping of vehicle movements in real-world coordinates.

\subsection{Availability of datasets}\label{sec:availability_of_datasets}

Recent datasets such as HighD~\cite{krajewski2018highd}, HARPY~\cite{makrigiorgis2020extracting}, pNEUMA~\cite{barmpounakis2020new}, HIGH-SIM~\cite{shi2021video}, and CitySim~\cite{zheng2022citysim} have primarily targeted specific traffic scenarios, such as single intersections or highway segments, often relying on relatively outdated \ac{CV} techniques. For example, HighD, HARPY, and HIGH-SIM employed U-Net, \ac{YOLO}v2, and \ac{YOLO}v3 for detection, respectively, but lacked robust mechanisms for trajectory validation. The pNEUMA dataset, generated by static drone swarms, provides insights into traffic flow but exhibits imprecise \acp{BB} and trajectory anomalies~\cite{kim2023visual, mahajan2023treating}, potentially due to visual limitations or suboptimal \ac{CV} tools. Meanwhile, CitySim has not fully disclosed its georeferencing methods, and the validation of its traffic data remains uncertain. 

In contrast, infrastructure-based approaches such as the I-24 MOTION~\cite{gloudemans202324}, while addressing similar challenges, capture comprehensive traffic dynamics over extended freeway segments using a dense network of 276 high-resolution, pole-mounted cameras spanning approximately 4.2 miles. The system utilizes a curvilinear coordinate system for accurate lane-level positioning and an automated post-processing pipeline to correct detection artifacts such as dropped frames, occlusions, and trajectory fragmentation~\cite{gloudemans202324}. More recently, efforts have been made to enhance the accuracy of I-24 MOTION trajectories by employing automated data reconciliation techniques~\cite{wang2024automatic}, leveraging vehicle dynamics and physical constraints to associate fragmented tracks, remove outliers, and impute missing data. However, such systems are inherently limited in adaptability and spatial coverage compared to \ac{UAV}-based systems, highlighting the distinctive advantage of \ac{UAV} flexibility in capturing detailed, dynamic urban environments.

Vehicle annotation datasets, such as CARPK~\cite{hsieh2017drone}, CyCAR~\cite{kouris2019informed}, HARPY~\cite{rafael_makrigiorgis_2022_7053442}, RAI4VD~\cite{puertas2022dataset}, UAVDT~\cite{yu2020unmanned}, UIT-ADrone~\cite{tran2023uit}, and VisDrone~\cite{zhu2021detection}, serve as valuable training resources for object detection models. Nonetheless, they frequently utilize basic annotation methods, cover a restricted or incoherent range of vehicle classes, and exhibit varying degrees of annotation accuracy, thereby potentially affecting the reliability and consistency of \ac{DL} models trained on these datasets. To mitigate these limitations, we incorporated these datasets in conjunction with our custom Songdo Vision dataset \cite{songdo_vision_dataset} and employed a multi-phase object detector training pipeline, which is described in \autoref{sec:object_detection}, where \autoref{tab:dataset_for_training} presents the relevant metadata for these datasets.

The datasets provided in this study, Songdo Traffic and Songdo Vision, are intended to significantly expand the scope and quality of existing traffic datasets by offering high-frequency georeferenced trajectories and extensive annotated data, thereby supporting advanced research in traffic monitoring and computer vision. For readers seeking detailed comparisons of existing datasets, we recommend referring to recent reviews such as \cite{gloudemans202324, butilua2022urban, wang2024automatic} for traffic datasets, and \cite{ding2021object, abdel2023advances, leng2024recent} for vehicle annotation-oriented datasets. These works provide extensive analyses, highlight existing limitations, and discuss future directions for dataset collection and methodological improvements.

\subsection{Summary}

In summary, while the potential of drones and \ac{CV} for urban traffic data extraction is evident, challenges such as varying object scales, occlusions, and the scarsity of comprehensive datasets persist~\cite{bouguettaya2021vehicle, ghahremannezhad2023object}. Most existing approaches are optimized for static cameras and lower altitudes, restricting their effectiveness in \ac{BEV} perspectives, where objects appear smaller and less distinctive. Video stabilization and georeferencing are critical for reliable trajectory extraction from high-altitude drone imagery. Common methods, including \ac{GSD}-based and manual \ac{GIS} georeferencing, often lack accuracy, while the incorporation of \ac{RTK-GNSS} with \acp{GCP} and advanced stabilization techniques shows promise in mitigating these limitations. Expanding the scope and quality of available datasets, as demonstrated in this study, can further support advanced traffic monitoring and urban planning.

\section{Methodology}\label{sec:methodology}

\begin{figure}[htbp]
  \centering
  \includegraphics[width=0.95\columnwidth]{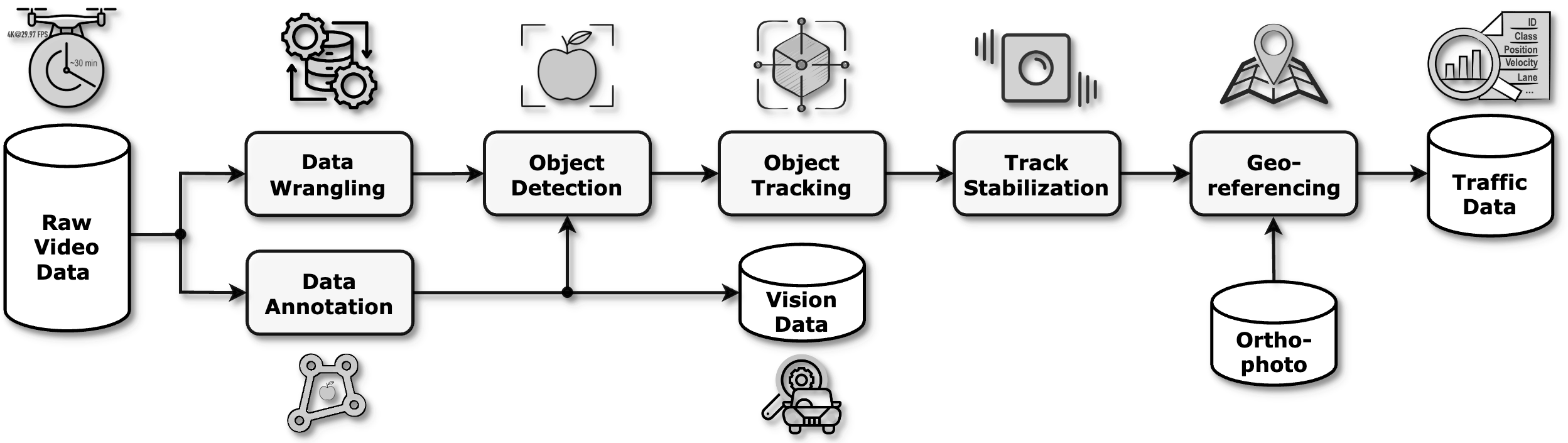}
  \caption{Diagram illustrating the vehicle trajectory extraction process from raw drone footage to traffic data, with auxiliary human annotation and vision data generation.}\label{fig:extraction_pipeline}
\end{figure}

Our methodology for transforming raw drone footage into detailed traffic data, which is compiled into the Songdo Traffic dataset, involves several key steps, as shown in \autoref{fig:extraction_pipeline}. First, in the data wrangling step, the raw videos are processed into meaningful segments. In parallel, selected frames are annotated to identify vehicle classes, which are used to train and validate an object detection algorithm, and are released as part of the Songdo Vision dataset. The detection algorithm processes the segmented videos to detect and locate vehicles. Next, a \ac{MOT} algorithm assigns unique \acp{ID} and tracks vehicles across frames. Image registration methods stabilize the footage and transform vehicle coordinates into a fixed reference frame. Georeferencing then converts these coordinates into real-world values using homographic and affine transformations, along with an orthophoto. Finally, the extracted trajectories are used to derive useful metadata, including vehicle dimensions, speed, acceleration, road segment, and lane number. By applying track stabilization after object tracking, we are able to bypass the limitations of traditional video stabilization and achieve more accurate results.

\subsection{Data wrangling}\label{sec:data_wrangling}

During the data wrangling (pre-processing) phase, we segmented 12TB of raw \ac{UHD} drone videos, saved in .mp4 format at 29.97 \ac{FPS}, along with their corresponding flight logs, into distinct segments. Each video segment corresponds to a stable hovering period of an individual drone over a monitored intersection (see \autoref{fig:intersections}). In addition to manual data wrangling, automated post-processing checks were performed to ensure data consistency and correct any remaining video or flight-log anomalies, as detailed in \ref{sec:appendix_validation}.

Due to the drone's built‐in file size limitation of 4GB per video file, the fragmented raw videos and corresponding flight logs were first merged into a single file per drone per flight session, resulting in approximately 30-minute segments on average. These merged files contained stationary footage at intersections and dynamic clips of drones transitioning between intersections or during take-offs and landings. Due to the lack of yaw measurements in the flight logs, manual segmentation by human annotators was necessary to identify the start and end frames of stable hoverings above intersections, as well as to identify any unintentional movements caused by technical glitches or human error, including significant drone movements during hoverings, recording issues, camera tilt, or zoom. This ensured that only stable and high-quality footage was retained.

After manual annotation of the intended cut locations, an automated process adjusted them to the nearest I-frame (intra-coded frame, a key reference point in video compression) to avoid re-encoding, significantly speeding up the editing process. The resulting video segments were then organized using an automated naming system, where each filename was generated based on the intersection label (see \autoref{fig:intersections}), determined by matching the average latitude and longitude from the flight logs with the nearest intersection center, and incorporated a unique sequence number appended to each hovering event, with the sequence resetting at the start of each flight session. Due to the optimized flight paths, video lengths varied, typically ranging from 2 to 4 minutes, depending on the drone and intersection.

\subsection{Data annotation}\label{sec:data_annotation}

The unique challenges of the Songdo study, particularly the \ac{BEV} aerial perspective at 140–150m altitudes, necessitated the creation of a specialized vehicle annotation dataset: \textit{Songdo Vision}~\cite{songdo_vision_dataset}. Existing public datasets often exhibit varying perspectives, resolutions, and object sizes incompatible with our study’s specific conditions. Additionally, annotations in these datasets are frequently inaccurate, containing poor-quality \acp{BB} that compromise detection performance. To overcome these limitations, we constructed a tailored dataset categorizing vehicles into four standardized classes: cars (including vans and light-duty vehicles), buses, trucks, and motorcycles. Annotation classes were consolidated to ensure consistency with public datasets used during training. Specifically, trucks were annotated as a single class without subdivision into subclasses (e.g., small vs. heavy trucks), while vans and light-duty vehicles, variously classified in external datasets, were uniformly grouped under the car class. This standardization prevented extensive re-annotation, which was beyond the scope of this study.

\begin{figure}[htbp]
  \centering
  \includegraphics[height=0.328\columnwidth, valign=t]{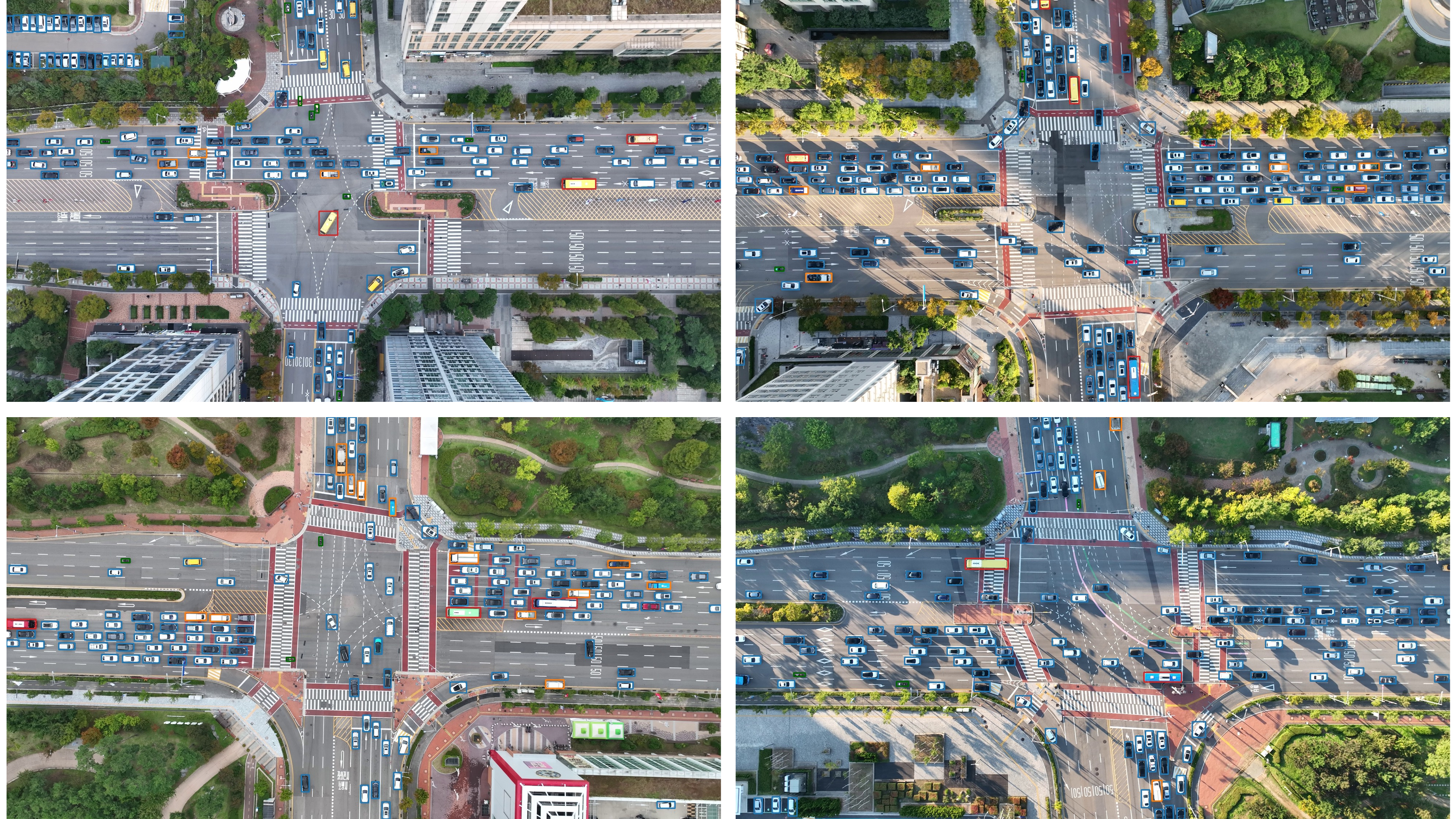}
  \hspace{0.002\columnwidth}
  \includegraphics[height=0.328\columnwidth, valign=t]{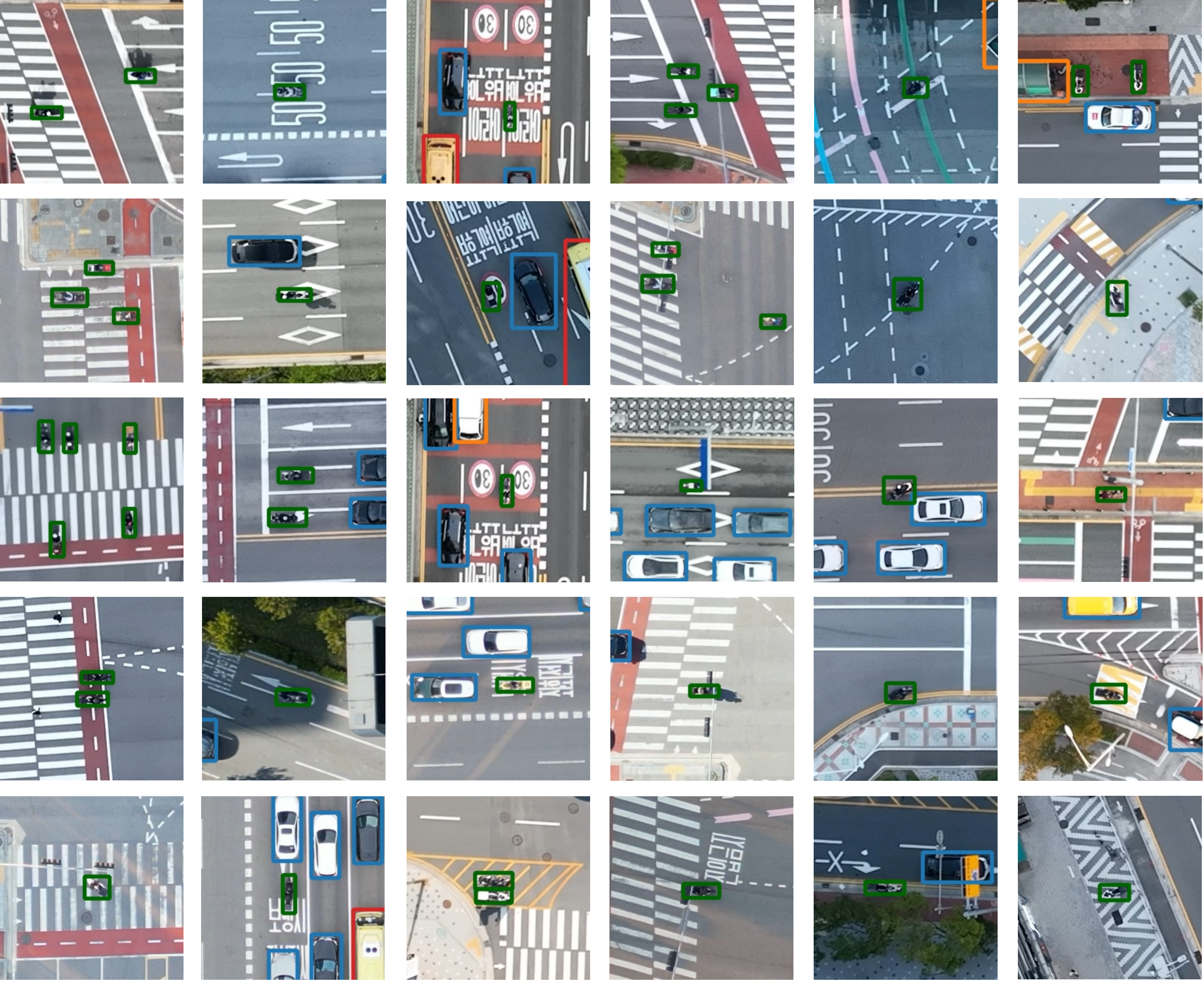}
  \caption{Left: Examples of human-annotated frames. Right: Examples of underrepresented motorcycle instances.}\label{fig:annotations}
\end{figure}

From the raw drone footage, we selected a representative subset of 5,419 video frames. Of these, 5,274 frames were randomly sampled from merged videos (described in \autoref{sec:data_wrangling}), covering stationary hovering and transitioning sequences. The remaining 145 frames were purposefully chosen to capture underrepresented yet critical scenarios, such as motorcycles at pedestrian crossings, bicycle lanes, traffic signals, and distinctive road markers, as illustrated in \autoref{fig:annotations} (right). Frames captured at unsuitable altitudes (e.g., drone take-off or landing sequences) or unfavorable camera angles were excluded, maintaining dataset integrity.

Annotations were conducted by trained annotators using Microsoft Azure ML Studio, requiring approximately 300 cumulative hours. Initially, bounding boxes were annotated manually. However, as annotations progressed, the platform’s automated labeling functionality was gradually activated, increasingly proposing candidate bounding boxes, thus accelerating the annotation process. To ensure high annotation accuracy and consistency, each frame underwent three iterative rounds of careful validation and refinement. Overall, nearly 300,000 vehicle instances were captured with tight, axis-aligned \acp{BB}, as illustrated in \autoref{fig:annotations} (left) and summarized in \autoref{tab:dataset_comparison}.

To facilitate reliable benchmarking, we established an 80 and 20 percent split of the Songdo Vision dataset for training and testing, respectively, ensuring an unbiased distribution of vehicle instances across all classes. By preserving this division in our public release, we enable other researchers to directly compare their custom-trained detectors against our results using the same test set. The detailed breakdown of this split is provided in \autoref{tab:dataset_comparison}, showing an almost perfect 80 to 20 percent ratio across all annotated class labels. The Songdo Vision dataset, enriched with human-verified annotations, is designed to significantly improve vehicle detection algorithms by providing a reliable \acl{GT} for model validation. The annotations are released in widely used formats such as COCO, \ac{YOLO}, and Pascal VOC, ensuring compatibility with a broad range of applications beyond traffic analysis, including autonomous drone delivery, surveillance, safety systems, and environmental studies.

\subsection{Object detection}\label{sec:object_detection}
Object detection is a pivotal component of our vehicle extraction framework. We employ a \ac{DL}-based object detector that predicts \acp{BB} encoded as  ($x$, $y$, $w$, $h$). Here, ($x$, $y$) represent the center coordinates of the \ac{BB}, while $w$ and $h$ denote its width and height, respectively. Each $k$-th image frame of video $v$, denoted $F_v^k$, is a $w_\mathcal{I} \times h_\mathcal{I}$ matrix of pixel intensities, where $w_\mathcal{I}$ and $h_\mathcal{I}$ represent the frame's width and height in pixels, respectively.

The object detector processes each frame $F_v^k$ to identify potential vehicles by predicting \acp{BB}. Following this initial detection step, our pipeline employs a class-agnostic \ac{NMS} algorithm with an \ac{IoU} threshold $\eta_{\text{IoU}}>0$. This crucial step effectively eliminates overlaps among the \acp{BB}, thereby refining the overall detection quality in densely populated urban settings. The refined set of detections, $\mathcal{D}_{v}[k]$, includes only the most reliable vehicle detections and is defined as follows:
\begin{equation}\label{eq:detection}
    \mathcal{D}_{v}[k] = 
    \left\{\left( 
    x^{(i)}[k], \; y^{(i)}[k], \; w^{(i)}[k], \; h^{(i)}[k], \; c^{(i)}[k], \; s^{(i)}[k] 
    \right) \; \big | \;  s^{(i)}[k] \geq \tau_\text{d} 
    \right\}_{i=1}^{N_{v}^d[k]},
\end{equation}
where $N_{v}^d[k]$ represents the number of vehicle detections in the $k$-th frame after processing through \ac{NMS}. For the $i$-th detection, $x^{(i)}$ and $y^{(i)}$ are the normalized center coordinates of the \ac{BB}, $w^{(i)}$ and $h^{(i)}$ are its normalized width and height, $c^{(i)}$ is the class label restricted to $c^{(i)} \in \mathcal{C} = \{0, 1, 2, \ldots, C-1\}$, with $C>0$ representing the total number of distinct object classes recognized by the detector, and $s^{(i)}$ is the confidence score constrained within $(0, 1]$. Detections are retained in $\mathcal{D}_{v}[k]$ only if their confidence scores exceed the detection threshold $\tau_\text{d} > 0$. All \ac{BB} coordinates and dimensions are normalized with respect to the input image size, ensuring that $0 \leq x^{(i)},\, y^{(i)},\, w^{(i)},\, h^{(i)} \leq 1$.

For our study, we employed the anchor-free \ac{YOLO}v8 detector architecture~\cite{yolov8_ultralytics}. Initially, we explored anchor-based \ac{YOLO}v5~\cite{jocher2022ultralytics} and transformer-based \ac{RT-DETR}~\cite{zhao2023detrs} as alternatives, but found that \ac{YOLO}v8 offered superior accuracy and faster inference speeds, particularly for smaller objects, making it more suitable for our operational requirements. Our choice of YOLOv8 was also influenced by the practical challenges associated with transformer-based architectures like DETR or \ac{RT-DETR}, which require extensive computational resources for training and inference due to their reliance on self-attention mechanisms. In addition, achieving satisfactory detection performance for smaller vehicles, such as motorcycles, necessitated increasing the input resolution beyond the default settings, which would further amplify DETR’s computational demands, as self-attention scales the computational complexity quadratically with the input resolution. While our 140–150m flight altitude ensured wide coverage, it inherently limited the detection of smaller objects such as pedestrians and bicycles. Lower altitudes and higher-resolution imaging improve small-object detection \cite{ren2018small}, but such configurations reduce spatial coverage and introduce operational trade-offs, such as the need for a larger drone fleet to maintain constant monitoring of the same geographical area.

After extensive trials, the ‘s’ model scale with standard output heads was selected for its optimal balance of accuracy and inference speed. While we experimented with an additional P2/4 output layer, as utilized in our previous \ac{YOLO}v5 setup\cite{espadaler2023continuous}, no significant performance gains were observed within the \ac{YOLO}v8 architecture. Our object detector was trained using a multi-stage approach. Initially, the model was trained on the large-scale ``BASE'' dataset, which consists of a diverse collection of images from various sources, offering broad coverage of vehicle types and scenarios. This training commenced with weights pre-trained on the COCO dataset~\cite{lin2014microsoft}, establishing a solid foundation for transfer learning. Subsequently, the model was fine-tuned on the ``FINE'' dataset, a curated, high-quality subset of BASE, specifically chosen for its accurate annotations and higher-resolution images, facilitating fine-grained model refinement. 

The datasets used for training, comprising images and annotations from eight public datasets and our Songdo Vision dataset (detailed in \autoref{sec:data_annotation}), are summarized in \autoref{tab:dataset_for_training}, which also specifies their usage across the BASE and FINE training stages. The public datasets include a variety of aerial images, characterized by varying altitudes, camera angles, perspectives, resolutions, illumination conditions, image quality, and train/test split ratios. For the public datasets, we adhered to the original train/test splits whenever available. In cases where no such split was provided, we applied an 80/20 division to maintain uniformity. Furthermore, datasets containing a high proportion of repetitive images were proportionally down-sampled to maintain a balanced representation across sources. This ensured that no single dataset dominated the training process, enhancing the model’s ability to generalize across diverse traffic conditions.

\begin{table}[htbp]
\small
  \centering
  \caption{Vision datasets used in multi-stage object detector training.}
  \label{tab:dataset_for_training}
  \begin{tabular}{lccll}
    \hline
    Dataset & BASE & FINE & Consolidated class labels & Present but unlabeled\\
    \hline
    Songdo Vision (ours)~\cite{songdo_vision_dataset} & \cmark &  \cmark & Car, bus, truck, motorcycle & Pedestrian, bicycle\\
    CARPK~\cite{hsieh2017drone} & \cmark &  \xmark & Car & Motorcycle, pedestrian\\
    PUCPR+~\cite{hsieh2017drone} & \cmark &  \xmark & Car & Motorcycle, pedestrian\\
    CyCAR~\cite{kouris2019informed} & \cmark &  \cmark & Car\\ 
    UAVDT~\cite{yu2020unmanned} & \cmark &  \xmark & Car, bus, truck & Motorcycle, pedestrian\\ 
    HARPY~\cite{rafael_makrigiorgis_2022_7053442} & \cmark &  \xmark & Car, bus, truck\\ 
    RAI4VD~\cite{puertas2022dataset} & \cmark &  \cmark & Car, bus, truck, motorcycle & Pedestrian\\
    UIT-ADrone~\cite{tran2023uit} & \cmark &  \cmark & Car, bus, truck, motorcycle, pedestrian, bicycle\\ 
    VisDrone~\cite{zhu2021detection}  & \cmark &  \xmark & Car, bus, truck, motorcycle, pedestrian, bicycle\\ 
    \hline
  \end{tabular}
\end{table}

To ensure consistency with our 140m–150m \ac{BEV} scenes in Songdo, we performed extensive preprocessing and filtering of these datasets. Although a comprehensive description of these steps is omitted here for brevity, we focused on consolidating the diverse class labels found in the public datasets to streamline our trajectory extraction process, which utilizes four primary vehicle categories and two auxiliary categories: $\mathcal{C}={0,, 1, , 2, , 3, , 4, , 5}$, where 0 corresponds to cars (including vans and light-duty vehicles), 1 to buses, 2 to trucks, 3 to motorcycles, 4 to pedestrians, and 5 to bicycles. While only the four primary vehicle categories were used during inference, the auxiliary categories (pedestrians and bicycles) were retained during training to improve overall classification performance. To ensure that our FINE dataset meets stringent training specifications, we excluded certain datasets, such as VisDrone and UAVDT. For example, VisDrone’s annotations primarily originate from non-\ac{BEV} environments at lower altitudes, resulting in inconsistent object size representations and varied camera settings that complicated integration. Additionally, this dataset exhibited mediocre annotation accuracy, further justifying its exclusion from our FINE dataset.

Although pedestrians and bicycles were excluded during inference due to their underrepresentation in the training data (see \autoref{tab:dataset_comparison}) and diminutive sizes, we included them in the training phase with the aim of reducing false positives from similarly sized objects like motorcycles and improving the model's ability to distinguish between closely related categories. To ensure consistency and enhance reliability, certain object classes in the public datasets were either disregarded or consolidated into the categories defined by $\mathcal{C}$. For instance, in the VisDrone dataset \cite{zhu2021detection}, where ``pedestrian'' and ``people'' are distinct labels, these categories were merged into a single ``pedestrian'' label, while classes like tricycles (including awning-tricycles) were excluded. 

\begin{table}[htbp]
\small
  \centering
  \caption{Annotation counts and class distributions across training and test sets for the BASE and FINE dataset groupings, with our Songdo Vision dataset also shown separately.}
  \label{tab:dataset_comparison}
  \begin{tabular}{lrrrrrrrrr}
    \hline
    \multirow{2}{*}{Vehicle class} & \multicolumn{3}{c}{BASE dataset} & \multicolumn{3}{c}{FINE dataset} & \multicolumn{3}{c}{Songdo Vision dataset} \\
    \cline{2-4} \cline{5-7} \cline{8-10}
     & Training & Test & Test ratio & Training & Test & Test ratio & Training & Test & Test ratio \\
    \hline
    Car & 561,666 & 148,375 & 20.9 \% & 266,745 & 52,260 & 16.4 \% & 195,539 & 49,508 & 20.2 \%\\
    Bus & 15,587 & 2,202 & 12.4 \%& 8,047 & 1,793 & 18.2 \%& 7,030 & 1,759 & 20.0 \% \\
    Truck & 28,830 & 4,869 & 14.4 \% & 14,305 & 3,452 & 19.4 \% & 11,779 & 3,052 & 20.6 \% \\
    Motorcycle & 44,512 & 11,478 & 20.5 \% & 30,925 & 10,309 & 25.0 \% & 2,963 & 805 & 21.4 \% \\
    Pedestrian & 24,239 & 1,983 & 7.6 \% & 1,260 & 243 & 16.2 \% & \multicolumn{3}{c}{} \\
    Bicycle & 4,472 & 276 & 5.8 \% & 86 & 45 & 34.4 \% & \multicolumn{3}{c}{} \\
    Total Vehicles & 679,306 & 169,183 & 19.9 \% & 321,368 & 68,102 & 17.5 \%& 217,311 & 55,124 & 20.2 \%\\    
    \hline
  \end{tabular}
\end{table}

To optimize our object detection for the complex and variable conditions of drone-based surveillance, we employed a comprehensive suite of data augmentation strategies. Specifically, the training set was subjected to random scaling (±50\%) and translations (±10\% of the image dimensions), as well as horizontal flips applied with a probability of 0.5. In addition, we introduced random perturbations to image hue, saturation, and brightness by scaling these channels with multiplicative factors of 0.015, 0.7, and 0.4, respectively. Mosaic augmentation was applied deterministically (probability = 1.0) by combining four distinct training images into a single composite image, thereby enhancing the model’s ability to generalize across varying spatial contexts and object scales~\cite{hao2020improved}. Furthermore, the Albumentations framework~\cite{buslaev2020albumentations} was employed to introduce additional perturbations, including Gaussian and median blur, grayscale conversion, and \ac{CLAHE}, each applied with a probability of 0.01, to improve robustness against variations in texture and illumination. All these strategies were essential for adapting the model to real-world variations in vehicle appearance caused by changing lighting conditions, weather, and diverse urban landscapes. 

Our training strategy employed \ac{SGD} optimization with \ac{YOLO}v8’s default hyperparameters, including an initial learning rate of 0.01, a final learning rate factor of 0.01, a momentum of 0.937, and a weight decay of 0.0005 to mitigate overfitting. Despite exploring various configurations, the default settings yielded the best performance. A batch size of 8 was used, with early stopping applied using a patience of 50 epochs, monitored on the test sets detailed in \autoref{tab:dataset_comparison}. \Acl{AMP} was enabled to improve efficiency without compromising accuracy. The loss function was calibrated with box, classification, and \acl{DFL} gains of 7.5, 0.5, and 1.5, respectively. Training and inference were conducted at a fixed resolution of $1{,}920 \times 1{,}920$ pixels using letterbox padding to preserve aspect ratio. This higher resolution was critical for detecting small objects like motorcycles, while a confidence threshold of $\tau_\text{d} = 0.25$ and an \ac{IoU} threshold of $\eta_\text{IoU} = 0.7$ ensured a balanced trade-off between precision and recall.

Songdo Vision, comprising 5,419 annotated images, was split into training and test sets, with 20\% (1,084 images) designated as the test set, allocated for both model validation during training and final testing of our top-performing object detector. \autoref{tab:dataset_comparison} shows the class distribution across the sets. Our top-performing model was selected based on its highest \ac{mAP} at an \ac{IoU} of 0.5 (mAP@50) across all classes. As shown in \autoref{tab:performance_metrics}, this model demonstrated remarkable performance across all vehicle types, achieving an overall mAP@50 of 0.951 and an exceptional mAP@50 of 0.888 for motorcycles, despite their smaller pixel size. Beyond conventional detection metrics, we also assessed the positional accuracy of \ac{BB} centers, revealing a mean Euclidean error of approximately 2.21 ± 1.99 pixels (12.2 ± 10.9 cm), further reinforcing the reliability of our detection step for trajectory extraction. These results highlight the model’s robustness in detecting small objects in \ac{BEV} drone videos. To briefly define the key metrics: Precision indicates the proportion of correctly identified vehicles out of the total detected, while Recall reflects the ratio of correct detections to the actual number of vehicles in a class. The mAP@50 measures detection accuracy using an \ac{IoU} threshold of 0.5, and mAP@50-95 evaluates performance over a range of \ac{IoU} thresholds (0.5 to 0.95), with a step size of 0.05, capturing variability in \ac{BB} overlaps.

\begin{table}[htbp]
  \centering
  \caption{Detection performance metrics on the Songdo Vision test set. Values before the slash (/) exclude the Songdo Vision train set from training, while \textbf{bold} values after the slash (/) include it.}
  \label{tab:performance_metrics}
  \begin{tabular}{lrrrrr}
    \hline
    Class & Precision & Recall & mAP@50 & mAP@50-95 \\
    \hline
    All & 0.766 / \textbf{0.911} & 0.690 / \textbf{0.935} & 0.742 / \textbf{0.951} & 0.503 / \textbf{0.711} \\
    Car & 0.934 / \textbf{0.979} & 0.971 / \textbf{0.981} & 0.970 / \textbf{0.992} & 0.702 / \textbf{0.835} \\
    Bus & 0.862 / \textbf{0.952} & 0.853 / \textbf{0.977} & 0.906 / \textbf{0.988} & 0.690 / \textbf{0.826} \\
    Truck & 0.737 / \textbf{0.887} & 0.410 / \textbf{0.916} & 0.593 / \textbf{0.935} & 0.421 / \textbf{0.722} \\
    Motorcycle & 0.532 / \textbf{0.827} & 0.527 / \textbf{0.866} & 0.499 / \textbf{0.888} & 0.197 / \textbf{0.463} \\
    \hline
  \end{tabular}
\end{table}

\begin{figure}[htbp]
  \centering
  \includegraphics[height=0.4\columnwidth, valign=t]{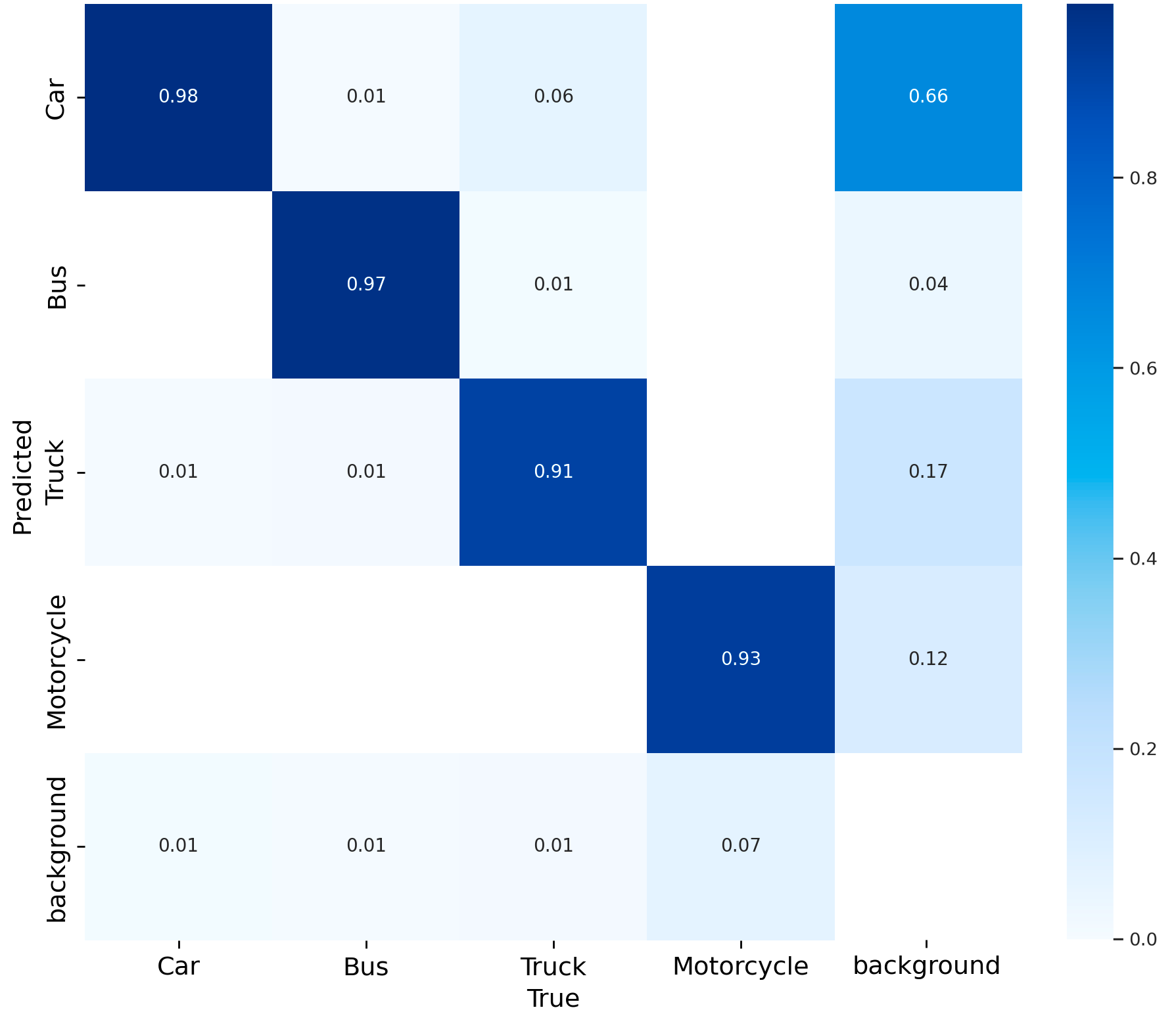}
  \hspace{6mm}
  \includegraphics[height=0.4\columnwidth, valign=t]{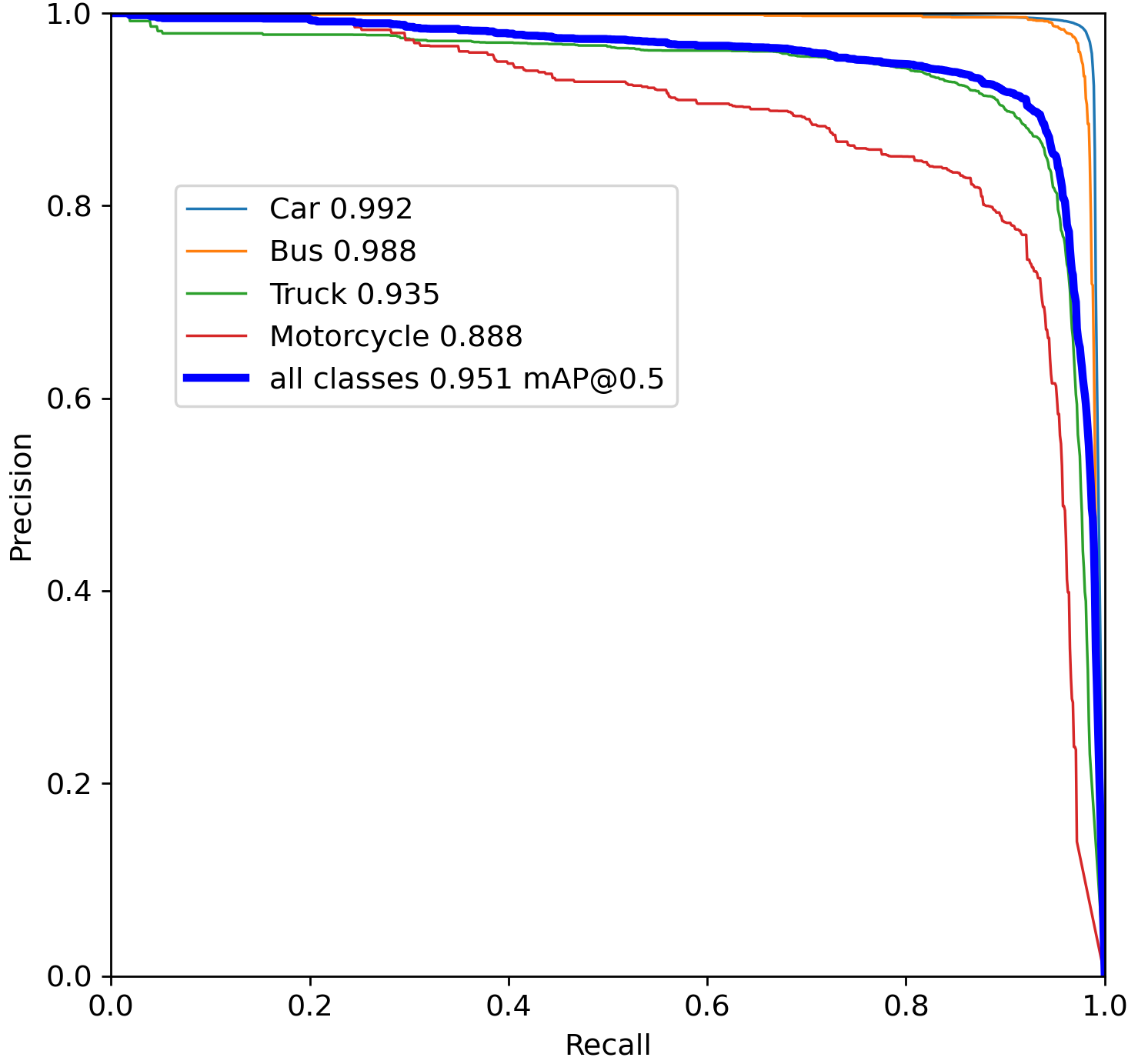}
  \caption{Left: Normalized confusion matrix. Right: Precision-recall curve.}\label{fig:confusion_matrix_and_PR_curve}
\end{figure}

The confusion matrix in \autoref{fig:confusion_matrix_and_PR_curve} (left) illustrates the model’s classification performance, highlighting accuracy and misclassification patterns across vehicle classes. The precision-recall curve in \autoref{fig:confusion_matrix_and_PR_curve} (right) further demonstrates strong detection performance, with low false positive and false negative rates. While smaller objects like pedestrians and bicycles are occasionally misclassified as motorcycles, particularly near pedestrian crossings or bike lanes, these instances were not filtered due to the inherent challenges in accurate differentiation. Despite this, the model's robust performance on motorcycles, though slightly lower than for larger vehicles, highlights the challenges of detecting smaller objects in drone footage. 

We also conducted an ablation study to assess the impact of our annotation efforts on detector performance. Specifically, we repeated the training procedure without the Songdo Vision dataset and observed substantial performance improvements across all metrics when it was included. As shown in \autoref{tab:performance_metrics}, the mAP@50 improved by +28.15\% and the mAP@50-95 by +41.35\%, highlighting the importance of high-quality annotations. Notably, the inclusion of Songdo Vision led to significant gains in detecting motorcycles, with a +78.16\% increase in mAP@50 and a +135.03\% increase in mAP@50-95, demonstrating the value of this dataset in enhancing detection performance for smaller objects.

In conclusion, our object detection system has demonstrated robust performance in identifying various vehicle types in drone-based surveillance. While our current model faces challenges in detecting smaller objects like pedestrians and cyclists, we are actively exploring strategies to address these limitations. By incorporating data augmentation techniques, investigating advanced algorithms, and leveraging newly emerging datasets with more comprehensive annotations, we aim to further enhance the system's capabilities.

\subsection{Object tracking}\label{sec:object_tracking}
Object tracking, specifically \ac{MOT}, is a crucial component of our framework, assigning unique \acp{ID} to each vehicle and consistently tracking their movements across video frames. This enables continuous tracking, even during temporary occlusions, providing valuable insights into vehicle behavior and interactions over time.

For the $k$-th frame of video $v$, the \ac{MOT} algorithm processes the detection set $\mathcal{D}_{v}[k]$ to generate a set $\mathcal{T}_{v}[k]$ of $N_{v}^t[k]$ tracks, each represented as:
\begin{equation}\label{eq:tracking}
\mathcal{T}_{v}[k] = 
\left\{\left(
id^{(i)}[k], \; x^{(i)}[k], \; y^{(i)}[k], \; w^{(i)}[k], \; h^{(i)}[k], \; c^{(i)}[k], \; s^{(i)}[k]
\right) \in \mathcal{I}_v[k] \times \mathcal{D}_v[k] \right\}_{i=1}^{N_{v}^t[k]},
\end{equation}
where '$\times$' denotes the Cartesian product, $\mathcal{I}_v[k]$ is the set of unique vehicle \acp{ID} in $v$ up to frame $k$, and $id^{(i)}[k]\in\{1,\,2,\,\ldots, V_v[k]\}$ is the unique \ac{ID} for each tracked vehicle, consistent across all frames. The number of active tracks in $F_v^k$ is $N_{v}^t[k]$, while $V_v[k]$ is the total number of unique vehicles observed up to frame $k$. The remaining symbols in \autoref{eq:tracking} have the same meanings as in \autoref{eq:detection}.

After applying tracking to all frames $F_v^k$, $k\in\{1,\ldots, N_v\}$, where $N_v$ is the total number of frames in $v$, we implement a class label refinement step for $c^{(i)}[k]$ in $\mathcal{T}{v}[k]$ to maintain classification consistency across frames for each tracked vehicle. Although our observations indicate such inconsistencies are very rare, we formalize this refinement through the following aggregation function:
\begin{equation}\label{eq:class_aggregation}
f_c(\text{id}) = \arg\max_{\text{c} \in \mathcal{C}} \left\{ \sum_{k=1}^{N_v}
\sum_{i=1}^{N_v^t[k]} s^{(i)}[k] \cdot
\mathbf{1}\left(c^{(i)}[k]=\text{c}\right) \cdot
\mathbf{1}\left(id^{(i)}[k]=\text{id}\right)
\right\}, \quad \text{id} = 1, \ldots, V_v[N_v],
\end{equation}
where $\mathbf{1}(\cdot)$ is the indicator function that is 1 if the condition is true and 0 otherwise. In this process, for each vehicle \ac{ID}, the ultimate class label is assigned by selecting the class $c \in \mathcal{C}$ that maximizes the sum of confidence scores across all frames where that \ac{ID} appears, that is, we refine the class labels to $c^{(i)}[k] = f_c(id^{(i)}[k])$.

The \ac{MOT} algorithm efficiently manages the assignment of new \acp{ID} and the removal of vehicles that no longer appear in the frame, ensuring that $N_v^t[k] \leq V_v[k]$ is maintained for all frames. While intermittent tracking failures due to prolonged occlusions or misdetections can result in multiple IDs for a single vehicle, such cases are rare and primarily occur with motorcycles navigating complex road infrastructure, as illustrated in \autoref{fig:annotations}. These instances have minimal impact on our primary goal of extracting precise trajectories.

We adopted the \ac{BoT-SORT} algorithm~\cite{aharon2022bot} for its enhanced Kalman filter state vector and efficient \ac{EMC}, which improve tracking accuracy for fast-moving and smaller vehicles like motorcycles. As an enhancement over the original ByteTrack~\cite{zhang2022bytetrack}, BoT-SORT’s optimizations make it effective in handling sporadic misdetections and suitable for dynamic environments. Given the minimal occlusions in our aerial footage, integrating a dedicated Re-ID module would yield limited improvement, making \ac{BoT-SORT} preferable due to its balance of performance and computational efficiency.

To manage occasional missed detections of smaller objects, particularly motorcycles, caused by the detector’s challenges in complex scenes, we employ a track buffer of $T_\text{buff}=30$ frames. This buffer retains information about tracks that may temporarily disappear due to occlusion or missed detections, helping prevent premature \ac{ID} deletion. However, in our Songdo experiment, its importance is limited since obstructions are rare and primarily confined to a few small bridges located away from main road arteries and intersections, typically at exit roads from underground garages (see \autoref{fig:intersections}). The \ac{BoT-SORT} algorithm manages track initiation and termination using several thresholds: a new track is created if the detection confidence score exceeds $\tau_\text{n} = 0.6$, high-confidence matches are made above $\tau_\text{h} = 0.5$, and low-confidence matches are allowed above $\tau_\text{l} = 0.1$. Associations between detections and tracks are ensured with a matching threshold of $\tau_\text{m} = 0.8$. Additionally, BoT-SORT’s efficient \ac{EMC} method based on sparse \ac{OF} compensates for camera movements by adjusting tracking to mitigate global background motion errors. This allows us to perform track stabilization as a subsequent step to object tracking, rather than relying on traditional video stabilization earlier in the pipeline (see \autoref{fig:extraction_pipeline}). The benefits of this approach are further discussed in \autoref{sec:stabilization_methodology}.

Although \ac{BoT-SORT} was originally validated on pedestrian tracking in street-level perspectives through the \ac{MOT} Challenge dataset \cite{aharon2022bot}, several factors support its reliability for our aerial vehicle tracking application. Unlike object detection, for which we have a representative sample of ground truth annotations, a comparable dataset for vehicle trajectories is unavailable in our context. Nevertheless, BoT-SORT inherently integrates appearance-based re-identification principles through its advanced Kalman filtering and efficient \ac{EMC}, mitigating global camera movements and significantly reducing identity-switching risks. Moreover, our setup benefits from favorable tracking conditions, including high-resolution \ac{BEV} imagery captured at 29.97 \ac{FPS} and vehicle speeds typically below 80 km/h, which ensure sufficient frame-to-frame overlap. Complete occlusions are exceedingly rare, minimizing the need for external visual-based ReID modules. Additionally, qualitative validation through manual inspection showed consistent tracking across sampled trajectories, while the comparison with trajectories recorded by the instrumented probe vehicle provided further confidence in the tracking accuracy.

To further assess the suitability of \ac{BoT-SORT}, we conducted a quantitative comparison with ByteTrack by evaluating vehicle trajectory length distributions across four representative drone videos from our dataset, see \autoref{sec:traffic_data_extraction_results}. Both algorithms produced virtually identical results, even in highly dense urban conditions and high-speed vehicle scenarios exceeding 150 km/h, as indicated by negligible KL divergence (0.0005) and matching trajectory counts. These findings align with previous studies showing that BoT-SORT achieves similar or higher accuracy than ByteTrack, despite ByteTrack’s faster processing speed~\cite{petersson2024object}. Combined with qualitative assessments, this analysis supports our choice of BoT-SORT. While ByteTrack offers comparable performance, BoT-SORT’s superior built-in \ac{EMC} and better handling of intermittent detections aligned more closely with our objectives. FairMOT was also considered; however, its joint detection-ReID architecture requires extensive fine-tuning and has high computational costs, which are impractical due to the lack of annotated aerial tracking data. Thus, FairMOT was deemed unsuitable for large-scale processing.

\subsection{Track stabilization}\label{sec:stabilization_methodology}

\begin{figure}[htbp]
\centering
\includegraphics[width=0.7\columnwidth]{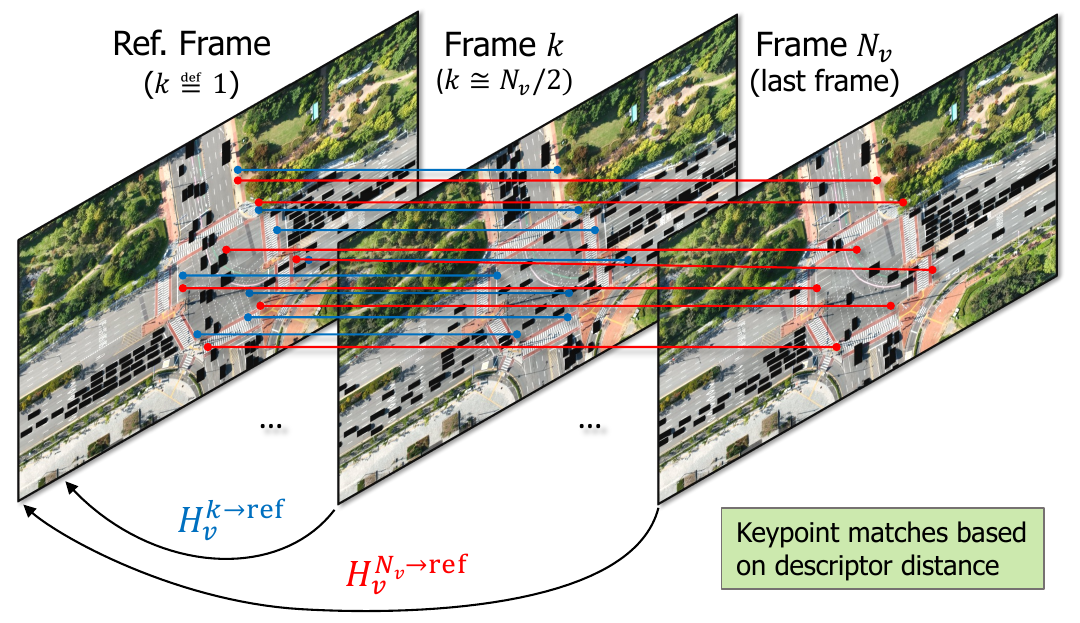}
\caption{Illustration of the image registration process. Keypoints from static background features are matched between frames and used to compute a homography matrix $H^{k \rightarrow \text{ref}}_v$ to align the frames. Exclusion masks prevent keypoint detection in dynamic vehicle regions, ensuring stable matching.} 
\label{fig:image_registration}
\end{figure}

To mitigate the effect of drone movements on the extracted trajectories, we ``stabilize'' each track $\mathcal{T}_v[k]$ by aligning the coordinates of each frame $F_v^k$ ($k = 2, \ldots, N_v$) with a reference frame $F_v^\text{ref}$, which is set as the first frame of the video $v$, i.e., $F_v^\text{ref} \triangleq F_v^1$. As illustrated in \autoref{fig:image_registration}, we achieve this alignment through image registration by identifying corresponding keypoints between pairs of grayscale frames ($F_v^\text{ref}$ and $F_v^k$), with optional \ac{CLAHE} applied to improve contrast. The matched keypoints are used to compute a $3 \times 3$  homography matrix $H_v^{k \rightarrow \text{ref}}$ using the least squares method, enhanced with a variant of the \ac{RANSAC} algorithm to reject outliers. Despite the marginally higher computational cost, we choose a projective transformation over an affine one to more effectively account for potential camera tilting caused by wind.

This approach ensures consistent alignment throughout the footage without accumulating errors over time, as each frame is independently aligned to the fixed reference frame rather than cumulatively re-using previous transformations. Consequently, for static scenes, the likelihood of error remains consistent for every frame, particularly considering the quasi-stationary nature of our drone setups. Additionally, this method significantly reduces storage and computational overhead compared to full-frame video stabilization techniques.

Recent studies~\cite{jin2021image, efe2021effect} suggest that classical image matching algorithms may perform comparably to the perceived \ac{SOTA} represented by learning-based approaches like SuperPoint or SuperGlue. These findings highlight the importance of careful hyperparameter tuning for classical methods and point out that \ac{DL}-based methods rely heavily on large training datasets and require fine-tuning for specific tasks. Given the absence of annotated training data for fine-tuning \ac{DL}-based methods and the computational cost associated with GPU-based inference, we opted for classical methods. This choice also aligns with the practical need for computational efficiency, considering our dataset size of approximately 12TB. Therefore, classical methods offer a robust solution without compromising accuracy, making them suitable for our experimental conditions.

\subsubsection{Implementation details}

We extract up to $K$ and $K_{\text{ref}}$ keypoints from each frame $F_v^k$ ($k = 2, \ldots, N_v$) and the reference frame $F_v^{\text{ref}}$, respectively, using a local feature detector. To ensure stable frame alignment, we focus keypoint detection on static background elements by leveraging the \acp{BB} from $\mathcal{D}_v[k]$ as ``exclusion masks''. These masks prevent keypoint detection within vehicle regions, which would otherwise introduce unstable matches due to vehicle movement. To account for potential inaccuracies in \acp{BB}, each box is enlarged by a factor $\epsilon > 0$, adjusting mask dimensions to $\tilde w = w(1 + \epsilon)$ and $\tilde h = h(1 + \epsilon)$.

Keypoint matches between each pair ($F_v^\text{ref}$, $F_v^k$) are identified based on a similarity measure calculated from keypoint descriptors, typically using Euclidean distance or cosine similarity, depending on the descriptor type. These potential matches, denoted as $m_v^{k \rightarrow \text{ref}}$, are refined using Lowe’s ratio test~\cite{lowe2004distinctive}, also known as \ac{SNN} test, which filters out non-discriminative matches using a threshold $\theta_\text{SNN} \in (0, 1]$, resulting in a refined match set $\tilde m_v^{k\rightarrow \text{ref}}$. This refined set is then used by a robust estimator to compute the homography matrix $H_v^{k \rightarrow \text{ref}}$. The confidence level $\tau$, the maximum number of iterations $\Gamma$, and the epipolar threshold $\eta$ are tuning parameters for most \ac{RANSAC} algorithms. To reduce computational load when determining homography matrices, we downscale the 4K input frames $F_v^k$ ($k = 1, \ldots, N_v$) using a scale factor $\rho$, resulting in adjusted frame dimensions $\tilde w_\mathcal{I} = \rho \times w_\mathcal{I}$ and $\tilde h_\mathcal{I} = \rho \times h_\mathcal{I}$.

The derived homography matrix $H_v^{k \rightarrow \text{ref}}$ can transform any pixel coordinates from frame $F_v^k$ to the reference frame. For a generic point $p_v^k = [x,\; y]^T$ in frame $F_v^k$, its corresponding location in the reference frame is:
\begin{equation}
p_v^\text{ref} = \begin{bmatrix}
x_\text{ref} & y_\text{ref}
\end{bmatrix}^T = \begin{bmatrix}
x'/z & y'/z
\end{bmatrix}^T,
\label{eq:homography_transform}
\end{equation}
where
\begin{equation}
\begin{bmatrix}
x' & y' & z
\end{bmatrix}^T = H_v^{k \rightarrow \text{ref}} \begin{bmatrix}
x & y & 1
\end{bmatrix}^T.
\label{eq:homogeneous_transform}
\end{equation}
Here, the point $p_v^k$ is converted to homogeneous coordinates by appending 1, the homography is applied, and the result is normalized by dividing by $z$. This transformation can be concisely represented as $p_v^\text{ref} = H_v^{k \rightarrow \text{ref}}(p_v^k)$, directly mapping each point from the original frame to its position in the reference frame.

In our track stabilization approach, we apply these transformations to all \ac{BB} coordinates in $\mathcal{T}_v[k]$, resulting in a new set of stabilized tracks, $\mathcal{T}_v^\text{ref}[k]$. We first convert the \acp{BB} from the \ac{YOLO} format ($x$, $y$, $w$, $h$) into a four-point format representing each box’s corners. Each corner is then transformed using $H_v^{k\rightarrow \text{ref}}$ to align with the reference frame. After transformation, the new positions are converted back into the \ac{YOLO} format by recalculating the center coordinates and dimensions, ensuring consistency. This direct application of homography to track data bypasses the need to stabilize the entire video, saving storage space and eliminating artifacts that can obscure parts of the scene, thereby enabling vehicle detection across the entire recorded area. This is a key feature of our work, emphasizing the novelty of stabilizing tracks rather than the video itself.

Furthermore, our masking strategy introduces another key innovation. Unlike conventional methods that warp each video frame to a reference frame before detection and tracking, our approach leverages the detected vehicles to enhance image registration. By applying exclusion masks based on detected \acp{BB}, we avoid keypoint matches on moving objects, which can undermine stabilization~\cite{zheng2022citysim}. This reordering improves accuracy and allows for less stringent hyperparameters without additional computational costs. Our stabilization process is available as the open-source library \textit{stabilo}~\cite{Fonod_Stabilo_2024}, which supports both video and track stabilization using user-provided masks, and can be integrated into diverse applications including autonomous surveillance, mapping, and environmental monitoring frameworks beyond vehicle tracking.

\subsubsection{Method and parameter search campaign}\label{sec:registration_campaign}

To systematically validate and refine our selection of classical algorithms and hyperparameters, we utilized 29 unique aerial scenes $\mathcal{S}$, each representing a randomly selected frame from our fleet of 10 drones monitoring 20 intersections. With multiple drones covering the same intersections, this compilation ensures a complete array of viewpoints, totaling $|\mathcal{S}|=29$ distinct scenes similar to those shown in \autoref{fig:intersections} (right).

For each method and hyperparameter combination, we performed a comprehensive evaluation using all scenes. Each scene was subject to $N_t=100$ trials involving both photometric and homographic distortions, amounting to 2,900 evaluations per combination. Photometric distortions include brightness variations of up to ±25\%, Gaussian blur with kernel sizes of up to 5 pixels, saturation adjustments within ±5\%, and fog effects with an intensity of up to 10\%. These distortions simulate various lighting conditions, weather scenarios, and camera focus variations to challenge the robustness of feature detection and matching algorithms. We then applied random homography transformations $H_s^{i\rightarrow i'}$ to each scene $s\in\mathcal{S}$. These transformations, designed to mimic drone movements and environmental effects like wind gust or altitude changes, included rotations up to ±15 degrees, translations up to ±10\% on both axes, scale changes up to ±5\%, and perspective shifts with coefficients up to $5 \times 10^{-5}$ to represent subtle changes in drone pitch or roll.

\begin{figure}[htbp]
  \centering
  \includegraphics[width=0.8\columnwidth]{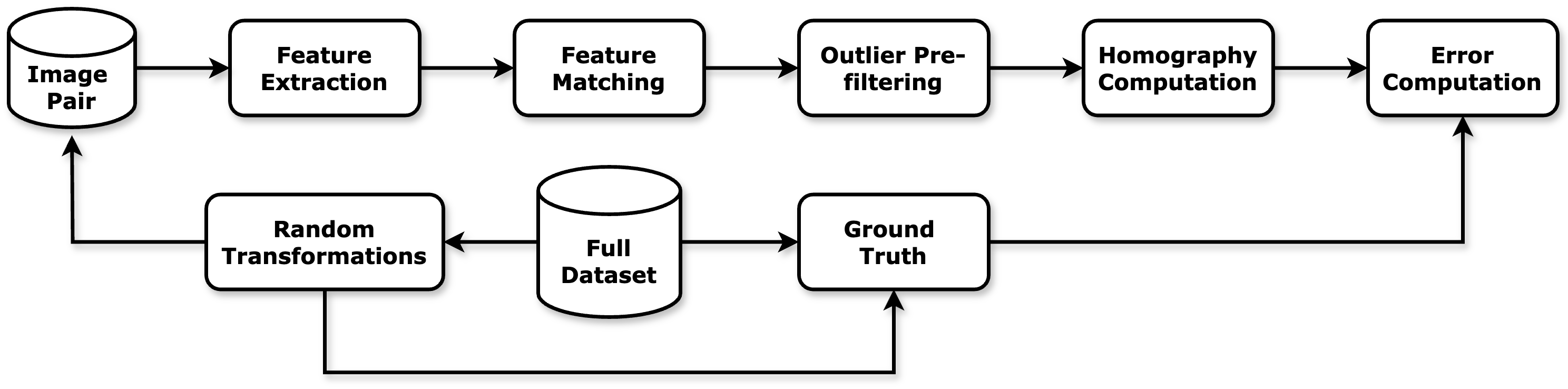}
  \caption{Overview of the image registration campaign.}
  \label{fig:registration_benchmark_pipeline}
\end{figure}

Using image pairs of original and distorted scenes, we applied the image registration techniques discussed earlier to estimate the homography matrix $\hat H_s^{i' \rightarrow i}$, using $H_s^{i \rightarrow i'}$ as the \acl{GT}. Comparing the estimated transformations against the actual ones ensures that our algorithms can robustly handle and correct the dynamic transformations typically encountered by drones in real-world scenarios. \autoref{fig:registration_benchmark_pipeline} provides an overview of our experimental setup, highlighting the rigor of our testing environment tailored for drone video analysis under varying and challenging conditions.

We implemented several prominent local feature detectors including \ac{SIFT}, \ac{RSIFT}, \ac{BRISK}, \ac{ORB}, KAZE, and \ac{A-KAZE}, along with various \ac{RANSAC} implementations such as the original \ac{RANSAC} algorithm, MAGSAC++, GC-RANSAC, DEGENSAC, etc. Both \ac{BF} and \ac{FLANN} matchers were evaluated, with the options to use exclusion masks and apply \ac{CLAHE}. We conducted extensive grid searches to optimize parameters such as the \ac{SNN} ratio threshold ($\theta_\text{SNN}$), the maximal number of extracted keypoints ($K$ and $K_\text{ref}$), the down-sample ratio ($\rho$), and the epipolar threshold ($\eta$), aiming to balance accuracy and computational efficiency. Following recommendations from recent studies~\cite{jin2021image}, we fixed $\tau$ at 0.999999 due to its minimal sensitivity to changes and set the maximum number of iterations $\Gamma$ at 5,000 for all \ac{RANSAC} algorithms to ensure sufficient convergence. Finally, we opted for unidirectional matching, avoiding bidirectional approaches to reduce computational demands. 

Traditionally, the \ac{mAA} metric is used to evaluate the effectiveness of image registration. However, this requires knowledge of the camera intrinsics and computation of the essential matrix. Alternatively, \ac{HEA} is a widely used metric~\cite{efe2021effect} for image registration evaluation, defined as:
\begin{equation}
\text{HEA} =
\frac{1}{N_t \times |\mathcal{S}|}
\sum\limits_{s\in\mathcal{S}}
\sum\limits_{i=1}^{N_t}
\left\{
\begin{aligned}
     1, & \quad \text{if} \quad  \left(\frac{1}{|\mathcal{P}|}\sum_{p \in \mathcal{P}} \left\|p - \hat H_s^{i'\rightarrow i} \left(H_s^{i\rightarrow i'}(p)\right) \right\|\right) \leq \varepsilon \\
     0, & \quad \text{otherwise}
\end{aligned}
    \right. .
\end{equation}
This equation calculates the mean accuracy across all transformations for each scene, where each corner’s displacement after a round-trip transformation through the true homography $H$ and its estimation $\hat{H}$ is computed. The set $\mathcal{P}$ contains the 2D coordinates of the four corner points of the original scene, and $H(\cdot)$ or $\hat H(\cdot)$ represent the homography transformation of a 2D point as described in \eqref{eq:homography_transform} and \eqref{eq:homogeneous_transform}. If the average displacement is within a threshold $\varepsilon$, it contributes a ‘1’ (accurate) to the mean; otherwise, it contributes a ‘0’.

Although \ac{HEA} effectively measures global scene alignment, it may overlook local inaccuracies critical for tracking vehicles. To overcome this limitation, we introduce a new metric, \ac{MIoU}, specifically designed to measure accuracy by comparing re-projected \acp{BB}, thus directly assessing object-level registration quality essential for reliable vehicle trajectory extraction. Formally, \ac{MIoU} is defined by:
\begin{equation}
    \text{MIoU}
    =
    \frac{1}{N_t \times |\mathcal{S}|}
    \sum\limits_{s\in\mathcal{S}}
    \sum\limits_{i=1}^{N_t}
    \left(
    \frac{1}{|\mathcal{B}_s|}
    \sum\limits_{B\in\mathcal{B}_s} 
    \frac{B \cap  \hat H_s^{i'\rightarrow i} \left(H_s^{i\rightarrow i'}(B) \right)}{B \cup \hat H_s^{i'\rightarrow i} \left(H_s^{i\rightarrow i'}(B) \right)}
    \right).
\end{equation}
This equation measures the mean \ac{IoU} for \acp{BB} across all scenes and transformations, where $\mathcal{B}_s$ denotes the set of \acp{BB} in scene $s\in\mathcal{S}$. It computes the \ac{IoU} between each original \ac{BB} $B$ and its transformed version after applying the true and estimated homographies. The advantage of this metric is that it does not require a pixel threshold and yields values ranging from 0 to 1, making it suitable for our application and valuable addition to \ac{HEA}.

In practical terms, \ac{HEA} provides a measure of the overall consistency of frame alignment, which is critical for maintaining accurate spatial relationships between frames during the trajectory extraction process. Meanwhile, \ac{MIoU} directly reflects the precision of object alignment by evaluating the overlap of re-projected vehicle bounding boxes, thereby serving as an indicator of the accuracy of detected vehicle positions after transformation. Together, these metrics ensure that both global frame alignment and local object positioning are accurately preserved, which is essential for reliable kinematic analysis and trajectory-based traffic studies.

\begin{figure}[htbp]
  \centering
  \includegraphics[width=0.495\columnwidth, valign=t]{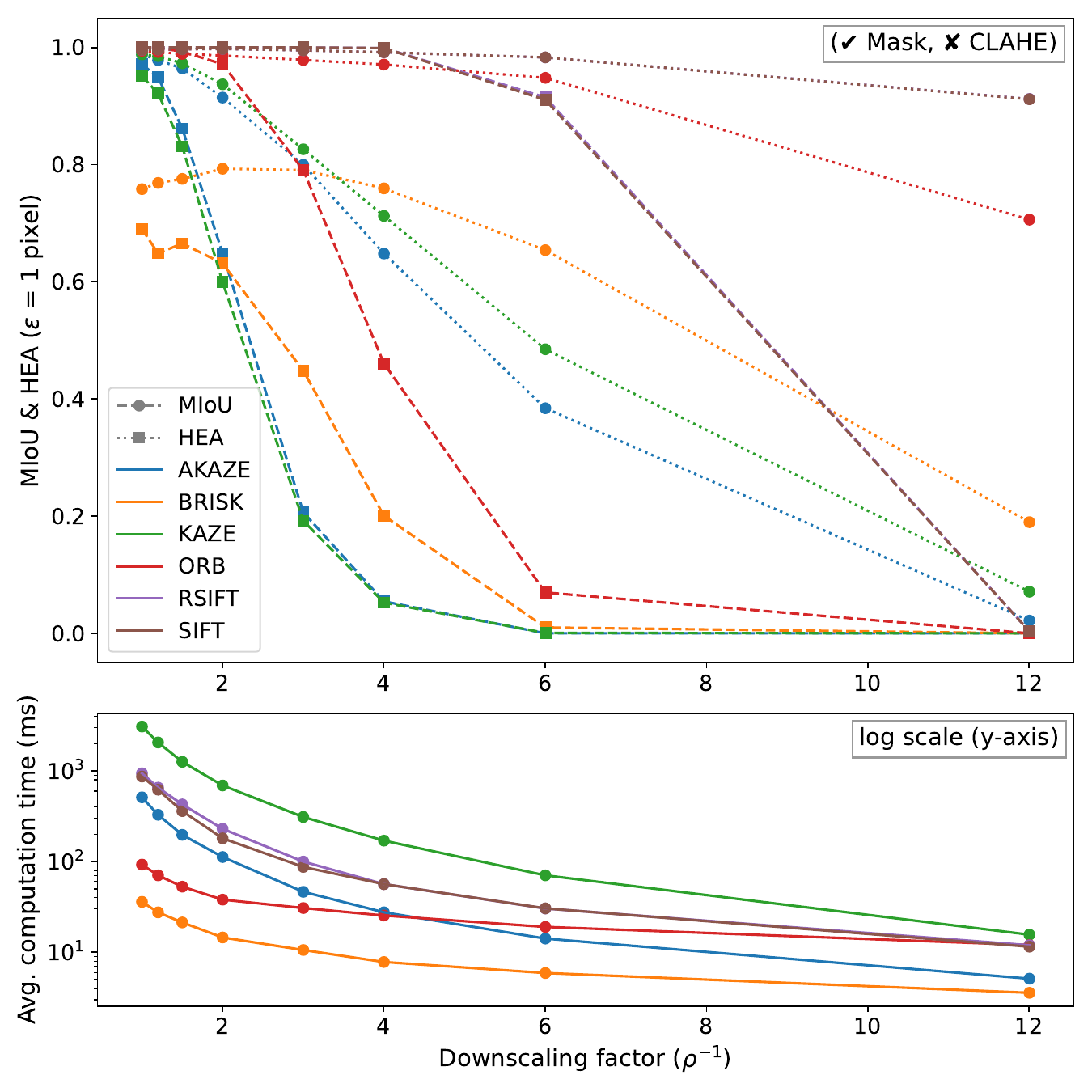}
  \includegraphics[width=0.495\columnwidth, valign=t]{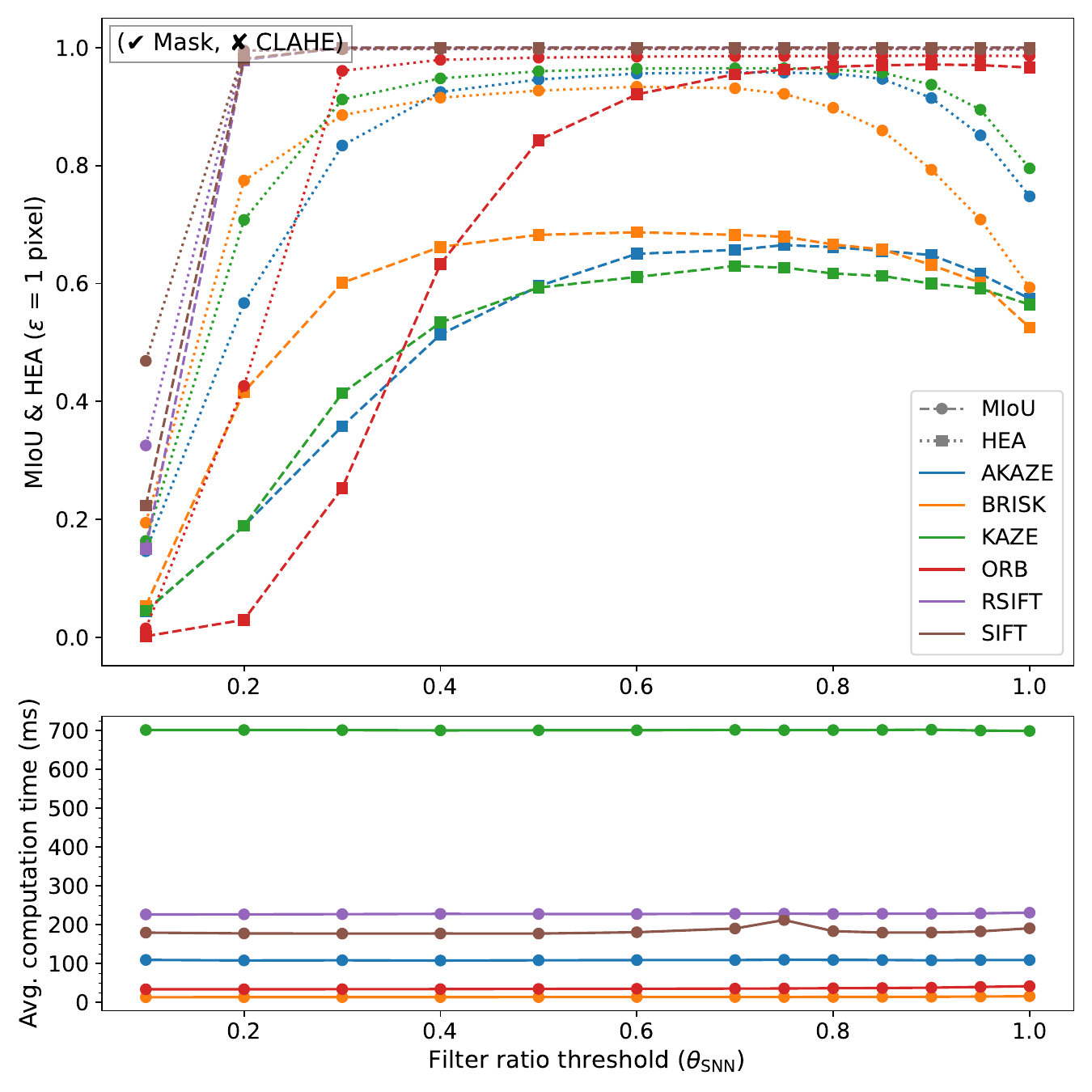}
  \caption[Effect of downscaling ratio and \ac{SNN} filter ratio threshold on the image registration performance.]{Effect of downscaling ratio and \ac{SNN} filter ratio threshold on the image registration performance\footnotemark.}
  \label{fig:stabilo_campaign}
\end{figure}
\footnotetext{All computational timings were obtained on a system equipped with an Intel Core i9-13900KF processor, NVIDIA GeForce RTX 3090 graphics card, and 64GB of RAM, operating on Ubuntu 22.04.3 LTS.}

While a detailed presentation of our campaign results is beyond the scope of this paper, we will make \textit{stabilo-optimize} \cite{Fonod_Stabilo_Optimize_2025}, an open-source optimization tool for stabilo, publicly available to enable replication and adaptation of our methods and hyperparameter tuning for similar image registration tasks. \autoref{fig:stabilo_campaign} offers insights into the optimization process, particularly highlighting the impact of downscaling and \ac{SNN} threshold on image registration performance. 

Ultimately, our findings suggest the following optimal set of methods and parameters: an \ac{ORB} detector with $K=2{,}000$ and $K_\text{ref}=4{,}000$, a \ac{BF} feature matcher with $\theta_\text{SNN}=0.9$, and MAGSAC++ with $\tau=0.999999$, $\Gamma=5{,}000$, and $\eta=2$, coupled with downscaling by a factor of $\rho=0.5$ and the use of \acp{BB} with $\epsilon=0.15$ as exclusion masks. The use of \ac{CLAHE} did not provide any significant benefits. Except for $\theta_\text{SNN}$ or $\rho$, these parameters were fixed for the considered trials. As shown in \autoref{fig:stabilo_campaign}, the selected combination of methods and parameters, particularly \ac{ORB} with a downscaling factor of $\rho=0.5$, achieves high \ac{MIoU} and \ac{HEA} values. This performance is on par with leading results reported in image matching studies~\cite{efe2021effect, jin2021image} and effectively balances superior accuracy with the computational efficiency required to process our extensive 12TB of video data.

\begin{figure}[htbp]
  \centering
  \includegraphics[width=0.95\columnwidth]{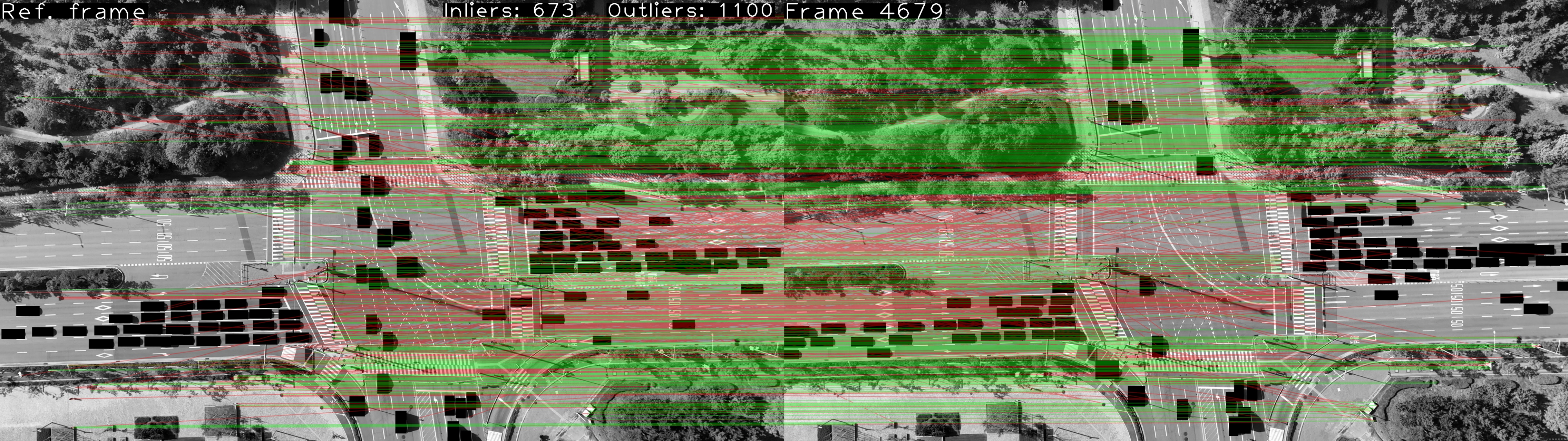}
  \caption{Image matching between the reference frame and the last frame of a video. Green lines indicate inliers used for homography computation, while red lines represent outliers rejected by the MAGSAC++ algorithm. Exclusion masks effectively prevent keypoint detection in vehicle regions.}
  \label{fig:image_registration_example}
\end{figure}

\autoref{fig:image_registration_example} illustrates image matching between the reference (first) frame and the last frame ($N_v=4{,}679$) of an example video $v$, spanning approximately 2 minutes and 36 seconds. Among the $1{,}773$ keypoints that passed the \ac{SNN} threshold, $673$ inliers were retained for computing $H_v^{N_v \rightarrow \text{ref}}$, while MAGSAC++ rejected the remaining $1{,}100$ as outliers. Notably, all matched keypoints lie outside the exclusion masks covering detected vehicles. Unlike the discarded matches, the inliers exhibit consistent directionality, indicating a robust subset for accurate homography estimation.

\subsection{Georeferencing}\label{sec:georeferencing}
\subsubsection{Orthophoto creation and road segmentation}\label{sec:ortho_creation}

\begin{figure}[htbp]
  \centering
  \includegraphics[height=0.4\columnwidth, valign=t]{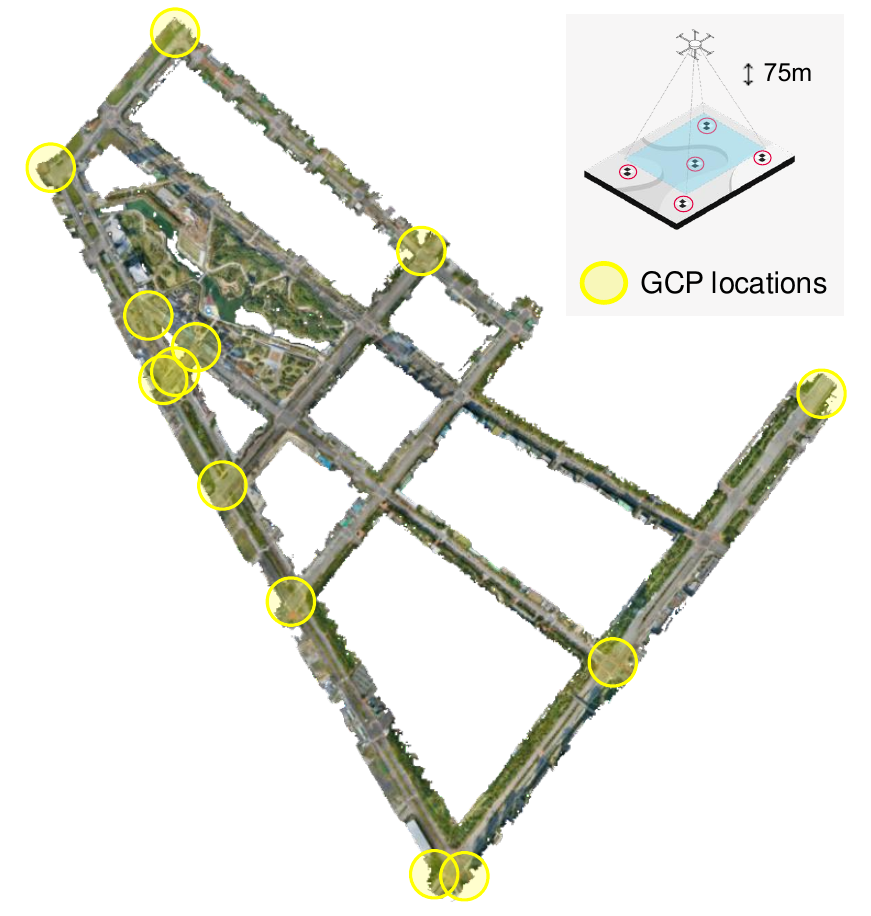}
  \hspace{10mm}
  \includegraphics[height=0.38\columnwidth, valign=t]{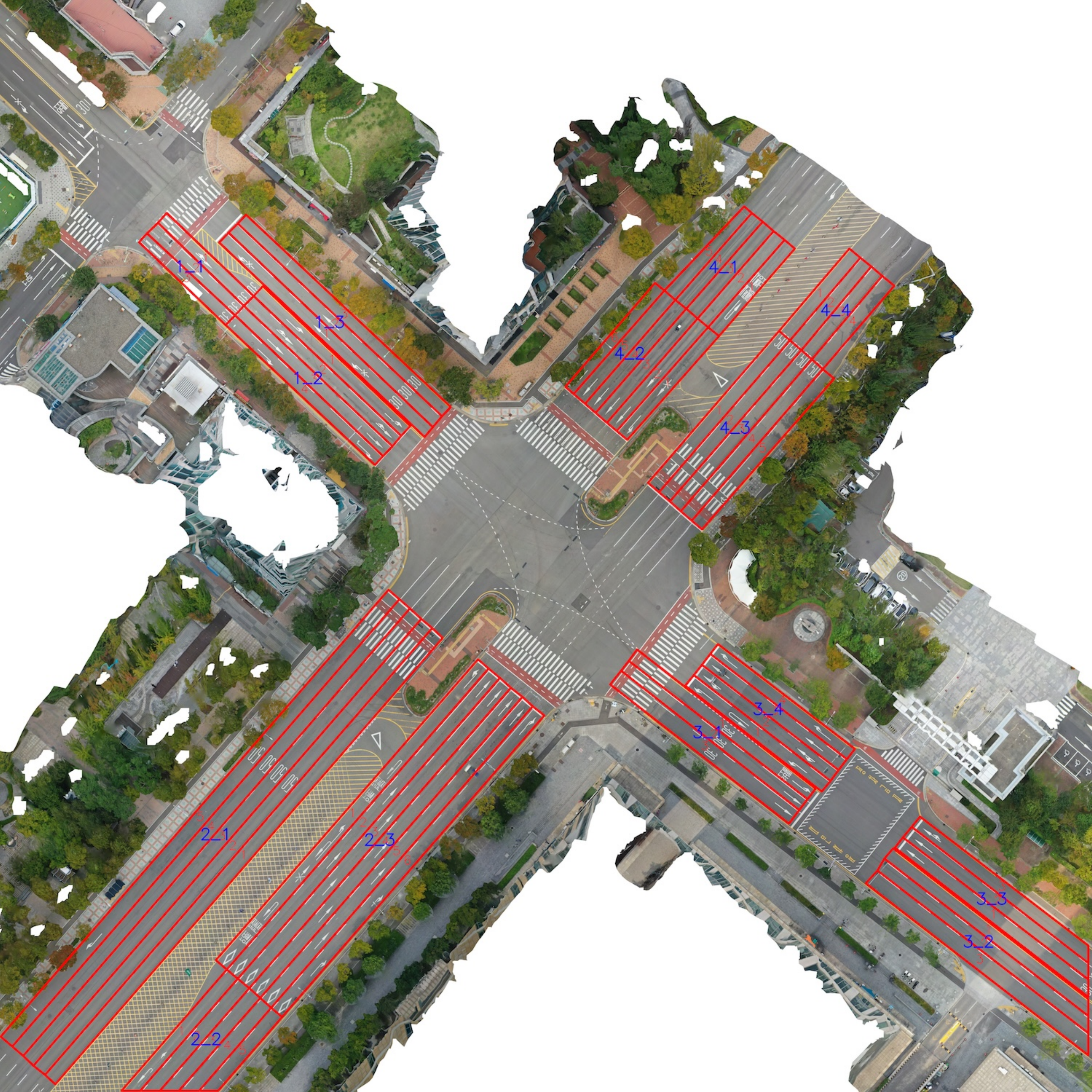}
  \caption{Left: The large orthophoto encompassing all 20 monitored intersections and showing the locations of the \acp{GCP} used during its creation. Right: Orthophoto cut-out example of an intersection, with overlaid road section and lane segmentations.}
  \label{fig:ortho_and_segmentations}
\end{figure}

In our study, a detailed orthophoto encompassing all 20 monitored intersections within the Songdo experimental area was produced, as shown in \autoref{fig:ortho_and_segmentations} (left). This was achieved using a dedicated DJI Mavic 2 Pro drone flying at an altitude of 75 meters, capturing overlapping high-resolution ($6{,}000 \times 4{,}000$ pixels) and distortion-free imagery that was then meticulously stitched using sophisticated photogrammetry methods facilitated by Dromii’s in-house \ac{DaaS} platform\footnote{
\url{http://new.dromii.com/}}. The drone captured images over the course of several days during the experiment and on additional dates to further enhance the final orthophoto. 

The geospatial accuracy of the orthophoto was significantly enhanced using 13 \acp{GCP}, each marked with a $1\times 1$ meter QR code and strategically placed across the experimental area, see \autoref{fig:ortho_and_segmentations} (left). These \acp{GCP} were calibrated using an EMLID RS2+ multi-band \ac{RTK-GNSS} receiver, featuring multi-feed antennas for multipath rejection and a fast \ac{RTK} convergence. According to the manufacturer, this rover can reliably ensure geolocation accuracy up to 8~km from the base station. The final orthophoto boasts a resolution of $209{,}181 \times 181{,}692$ pixels and a file size exceeding 30GB, providing an unprecedented level of detail. 

For focused traffic analysis, $15{,}000 \times 15{,}000$ pixel cut-outs centered on each intersection were first extracted from the large orthophoto, see \autoref{fig:ortho_and_segmentations} (right) of an illustration of a cut-out. These cut-outs sufficiently cover areas our drones monitor and are instrumental for in-depth traffic analysis and visualizations. \Ac{DL}-based techniques were integrated into the photogrammetry process to systematically exclude vehicles from the imagery, ensuring that the orthophoto is devoid of transient obstructions such as cars. Tall objects such as buildings were also removed to prevent parallax errors and ensure reliable planar homography for georeferencing. These exclusions do not affect the accuracy of georeferencing, as sufficient areas remain for extracting relevant feature points. Crucially, the primary objective for accurate trajectory georeferencing is to ensure feature correspondences at the road level.

Each of the 20 cut-outs underwent manual segmentation to delineate road sections and lanes, as illustrated in \autoref{fig:ortho_and_segmentations} (right). Sections refer to areas that group lanes moving in the same direction within defined parts of the intersection. Each section is labeled as ``N\_G’’, where N designates the specific intersection node and G represents the grouping of lanes by direction and number. Lanes within each section are numbered sequentially from the innermost lane outward. This manual segmentation required approximately 50 working hours, involving similar iterative validation procedures as described previously for vehicle bounding box annotations (cf. Section~\ref{sec:data_annotation}). This enabled linking each trajectory point directly to a specific road section and lanes, enhancing the dataset’s utility and providing invaluable metadata for intersection-level traffic analysis.

\subsubsection{Trajectory georeferencing}\label{sec:trajectory_georeferencing}

\begin{figure}[htbp]
  \centering
  \includegraphics[width=1\columnwidth]{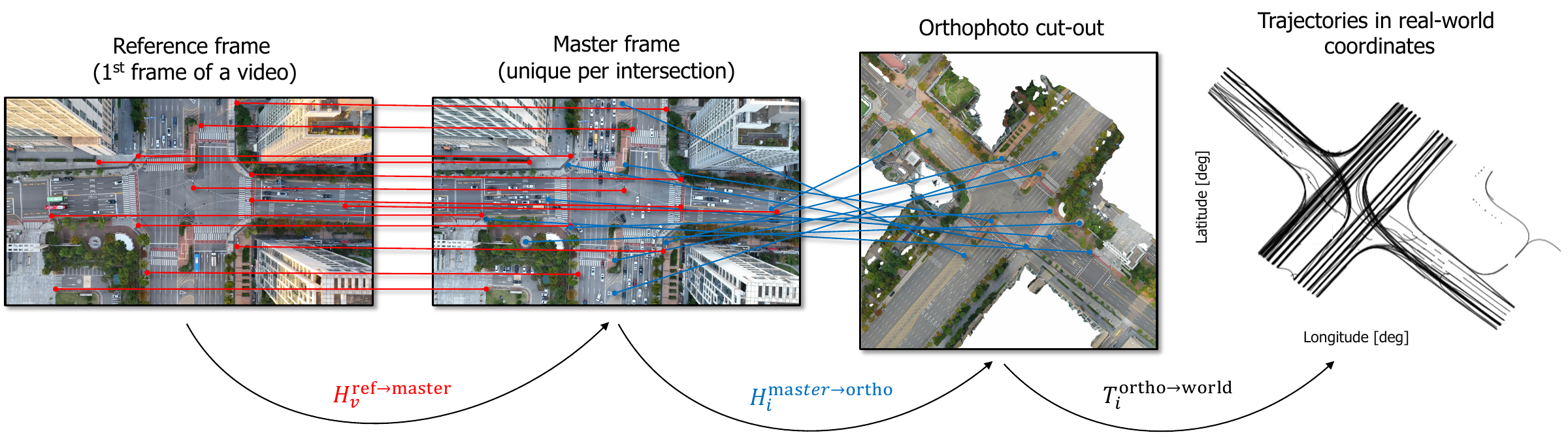}
  \caption{Illustration of the steps involved in the georeferencing process.}
  \label{fig:georeferencing_methodology}
\end{figure}

The objective of georeferencing is to convert vehicle trajectories in $\mathcal{T}^\text{ref}_v$ into real-world geographic coordinates, such as longitude and latitude. This transformation, illustrated in \autoref{fig:georeferencing_methodology}, is achieved through a three-step process, where a key novelty of our approach is the introduction of a ``master frame'' as an intermediary \ac{CS}. Using a unique master frame, selected based on optimal image quality and consistent reference features, for each intersection, we ensure that trajectories extracted from different drone viewpoints (with varying orientations, hovering locations, or altitudes) are consistently transformed into a unified coordinate system.

\textbf{Step 1:} For each video $v$, we calculate a homography matrix $H_v^{\text{ref} \rightarrow \text{master}}$, which maps the stabilized tracks from the reference frame $F_v^{\text{ref}}$ to a unique master frame $F_i^{\text{master}}$ designated for each intersection. The master frame serves as a unified basis to ensure consistent alignment across all videos capturing the same location.

\textbf{Step 2:} A homography matrix $H_i^{\text{master} \rightarrow \text{ortho}}$ is then computed to map coordinates from the master frame to the orthophoto coordinates $F_i^{\text{ortho}}$. Here, $F_i^{\text{ortho}}$ is the orthophoto cut-out corresponding to intersection $i$. The complete transformation is expressed as:
\begin{equation}\label{eq:ortho_transformation}
H_v^{\text{ref} \rightarrow \text{ortho}} = H_i^{\text{master} \rightarrow \text{ortho}} \cdot H_v^{\text{ref} \rightarrow \text{master}}.
\end{equation}
This composite homography $H_v^{\text{ref} \rightarrow \text{ortho}}$ allows us to map any point $p_v^{\text{ref}}$ in the reference frame $F_v^{\text{ref}}$ directly to its corresponding point $p_i^{\text{ortho}}$ in the orthophoto cut-out $F_i^{\text{ortho}}$, i.e., $p_i^{\text{ortho}} = H_v^{\text{ref} \rightarrow \text{ortho}}(p_v^{\text{ref}})$. Applying this transformation to all stabilized vehicle coordinates in $\mathcal{T}^\text{ref}_v[k]$, we obtain the vehicle trajectories in the orthophoto \ac{CS}.

\textbf{Step 3:} The transformation from orthophoto coordinates to real-world geographic coordinates is performed using a $2 \times 3$ affine transformation matrix $T_i^{\text{ortho} \rightarrow \text{world}}$ derived from the $i$-th orthophoto metadata. This matrix enables the accurate conversion of pixel coordinates into geographic coordinates as follows:
\begin{equation}
\begin{aligned}
\phi  &= a_i \cdot x_{\text{ortho}} + b_i \cdot y_{\text{ortho}} + t_{x_i}, \\
\lambda &= c_i \cdot x_{\text{ortho}} + d_i \cdot y_{\text{ortho}} + t_{y_i},
\end{aligned}
\end{equation}
where $x_{\text{ortho}}$ and $y_{\text{ortho}}$ are pixel coordinates in the orthophoto \ac{CS}, and $a_i$, $b_i$, $c_i$, $d_i$, $t_{x_i}$, and $t_{y_i}$ are the parameters of $T_i^{\text{ortho} \rightarrow \text{world}}$. These parameters account for scale, rotation, and translation adjustments required to map pixel coordinates to real-world coordinates $\phi$ (latitude) and $\lambda$ (longitude). By systematically applying these steps, we ensure precise conversion of all stabilized vehicle center coordinates ($x^{(i)}[k]$, $y^{(i)}[k]$) in $\mathcal{T}_v^\text{ref}[k]$ into both global geographic \ac{CS} (latitude and longitude, EPSG:4326, WGS84) and local Cartesian \ac{CS} (X and Y, EPSG:5186).

This three-step georeferencing approach addresses the challenges of directly matching drone imagery with orthophotos, which differ significantly due to orthorectification, resolution, and excluded areas. Direct matching for each video would be more time-consuming due to higher orthophoto resolution and would likely introduce varying errors, as the homography between each reference frame and the orthophoto needs to be computed independently. By introducing the master frame as an intermediary \ac{CS}, our method consolidates this process into a simple homography computation per video, converting potential varying errors into a consistent, systematic error for all trajectories at a given intersection. This approach simplifies the transformation process by relying on $H_v^{\text{ref} \rightarrow \text{master}}$ to match visually similar images, thereby improving reliability. To select the optimal master frame, we sorted reference frames based on their proximity to the average hovering location and chose the one with the least vehicle coverage among the top ten. This multi-stage approach contributes to the accuracy of trajectory mapping and enhances consistency across datasets obtained from multiple viewpoints, representing an important aspect of our methodology.

\subsubsection{Optimization of orthophoto cut-out resolution for accurate georeferencing}\label{sec:georeferencing_validation}

While the transformation $T_i^{\text{ortho} \rightarrow \text{world}}$ is directly obtained from the orthophoto creation process, the homographies involved in georeferencing must be computed using image registration techniques similar to those used in track stabilization. Since these homographies are computed only once per video ($H_v^{\text{ref}\rightarrow \text{master}}$) and once per intersection ($H_i^{\text{master} \rightarrow \text{ortho}}$), computational efficiency is not a limiting factor. This allows for the use of \ac{RSIFT}, which employs a square root (Hellinger) kernel to measure the similarity of \ac{SIFT} descriptors instead of the standard Euclidean distance. In our implementation, we set a small positive parameter of $10^{-8}$ for \ac{RSIFT} normalization to prevent division by zero. Optimization experiments confirmed that this approach improves performance across all stages of image registration~\cite{arandjelovic2012three}. We configured the implementation with a maximum of $K=250{,}000$ feature points, a \ac{BF} matcher with $\theta_\text{SNN}=0.55$, and MAGSAC++ with $\tau=0.999999$, $\Gamma=10{,}000$, and $\eta=3$. Furthermore, we did not apply \ac{CLAHE} or exclusion masks, as the vehicles in the reference and master frames differ, and the orthophoto does not contain vehicles.

\begin{figure}[htbp]
  \centering  
  \includegraphics[width=0.95\columnwidth]{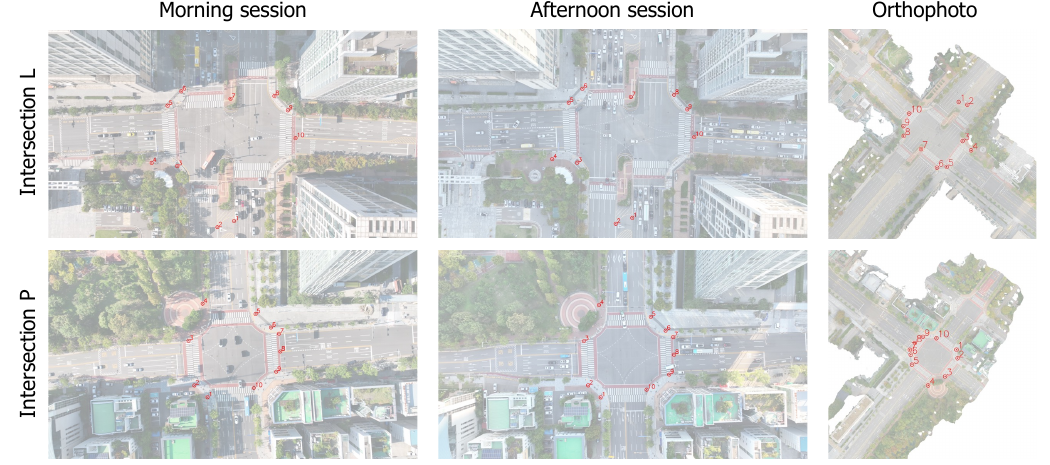}
  \caption{Locations of the \acp{GCP} for the selected intersections L and P. Each image has been slightly dimmed to enhance the visibility of the overlaid markers.}
  \label{fig:ortho_and_segmentations_georegistration}
\end{figure}

To validate the reliability of our georeferencing process, we conducted a comprehensive evaluation to determine the optimal resolution for orthophoto cut-outs. While the 4K reference and master frames were used at their full resolution without downscaling, we performed an ablation study to assess the impact of different resolutions for the orthophoto cut-outs. Specifically, we selected two intersections, L and P. For each intersection, we chose two reference frames, one from a morning and one from an afternoon flight session, to account for varying lighting conditions. We meticulously annotated these frames and the full-sized cut-outs with 10 corresponding \acp{GCP} for each frame and cut-out pair, aiming for pixel-level accuracy (see \autoref{fig:ortho_and_segmentations_georegistration} for illustration). This resulted in a total of 40 \ac{GCP} correspondences for validation.

For each combination of image frame and orthophoto cut-out, we computed the homography matrix using the specified image registration techniques and parameters. This process was repeated for various orthophoto cut-out resolutions. Mean and standard deviation of pixel reprojection errors were calculated by transforming points from the orthophoto cut-outs back to the image frame and comparing them with the original \acp{GCP}. \autoref{tab:ortho_error} summarizes these errors for each resolution, along with the mean computation time and the mean and minimum number of inliers after MAGSAC++ based outlier rejection. The results may include some human error due to manual \acp{GCP} assignment. All computations were performed on the same infrastructure reported in \autoref{sec:registration_campaign} to ensure consistency in timing measurements. 

\begin{table}[htbp]
  \centering
  \caption{Image matching performance between selected frames and orthophoto cut-outs at various resolutions.}\label{tab:ortho_error}
  \begin{tabular}{>{\raggedleft\arraybackslash}p{2.1cm}>{\raggedleft\arraybackslash}p{2.7cm}>{\raggedleft\arraybackslash}p{2.5cm}>{\raggedleft\arraybackslash}p{1.75cm}>{\raggedleft\arraybackslash}p{1.7cm}}
    \hline
    Cut-out resolution (px) & Mean computation time (s) & Reprojection error (px) & Mean \# of inliers & Minimum \# of inliers \\
    \hline
    2'000  &  2.584 &  3.225 $\pm$ 2.748 & 116.75 & 39 \\
    3'000  &  4.060 &  1.945 $\pm$ 1.357 & 189.25 & 62 \\
    4'000  &  8.632 &  1.671 $\pm$ 1.315 & 258.75 & 110 \\
    5'000  & 15.004 &  1.479 $\pm$ 1.014 & 235.50 & 96 \\
    6'000  & 23.337 &  1.424 $\pm$ 1.064 & 236.25 & 107 \\
    7'000  & 31.212 &  1.317 $\pm$ 0.888 & 234.50 & 95 \\
    8'000  & 41.978 &  1.323 $\pm$ 0.970 & 198.25 & 82 \\
    9'000  & 51.306 &  1.304 $\pm$ 0.928 & 208.00 & 104 \\
    10'000 & 53.432 &  1.421 $\pm$ 0.890 & 151.75 & 57 \\
    11'000 & 55.936 &  1.454 $\pm$ 0.942 & 145.75 & 71 \\
    12'000 & 58.552 &  1.378 $\pm$ 0.936 & 130.25 & 53 \\
    13'000 & 61.486 &  1.312 $\pm$ 0.886 & 115.25 & 38 \\
    14'000 & 64.622 &  1.419 $\pm$ 0.932 & 121.00 & 50 \\
    15'000 & 67.898 &  1.409 $\pm$ 0.908 & 96.50 & 37 \\    
    \hline
  \end{tabular}
\end{table}

From the results, we observe that increasing the orthophoto cut-out resolution improves matching accuracy, as indicated by a decrease in reprojection error. Although the lowest error is achieved at 9,000 pixels (1.304~px) and comparable results are seen at 7,000 pixels (1.317~px), the 8,000-pixel resolution offers a robust trade-off between accuracy and computational efficiency. Importantly, based on our observations, the minimum number of inliers must remain above approximately 30 to ensure reliable homography estimation. Additionally, we note that the number of inliers decreases at higher resolutions, likely due to the increased image detail causing more mismatches and the computational limits of feature matching algorithms at higher resolutions. 

Given these considerations, we selected an $8,000 \times 8,000$ pixel resolution for subsequent processing. This resolution not only facilitates the manual segmentation of road sections and lane delineations but also results in manageable orthophoto cut-out file sizes, enhancing dataset usability. With a mean reprojection error well below two pixels, our orthophoto registration method demonstrates high accuracy, validating the effectiveness of our georeferencing approach. Furthermore, the process of mapping from reference frames to master frames, an integral part of our method, is significantly faster and yields a notably higher average number of inliers, ranging from several hundreds to thousands, depending on the intersection.

\subsection{Traffic data}
\subsubsection{Robust vehicle dimension estimation}\label{sec:vehicle_dimensions}

Accurate estimation of vehicle dimensions from drone-based video imagery is vital for various traffic analysis applications, yet it remains an underrepresented challenge in scientific literature. To address this gap, we present a novel approach that leverages both un-stabilized and stabilized \acp{BB} from sequences $\mathcal{T}_v[k]$ and $\mathcal{T}^\text{ref}_v[k]$, respectively. Our unique methodology offers a significant advancement in the field, providing a level of accuracy and robustness previously unattained in similar studies.

Our approach employs a systematic process designed to enhance the robustness and accuracy of vehicle dimension estimates. The methodology focuses on fully visible \acp{BB} and vehicle motions parallel to the X or Y axes of the reference frame $F_v^\text{ref}$. The process encompasses visibility filtering, initial dimensions computation, azimuth-based filtering, final dimension computation, and real-world conversion. Additionally, we incorporate a solution for stationary vehicles, which is applicable only when these vehicles are aligned with the $F_v^\text{ref}$ axis. Detailed implementation steps and experiment-specific parameters are provided in \ref{sec:appendix_dimension_estimation}.

\begin{figure}[htbp]
  \centering
  \includegraphics[height=0.29\columnwidth]{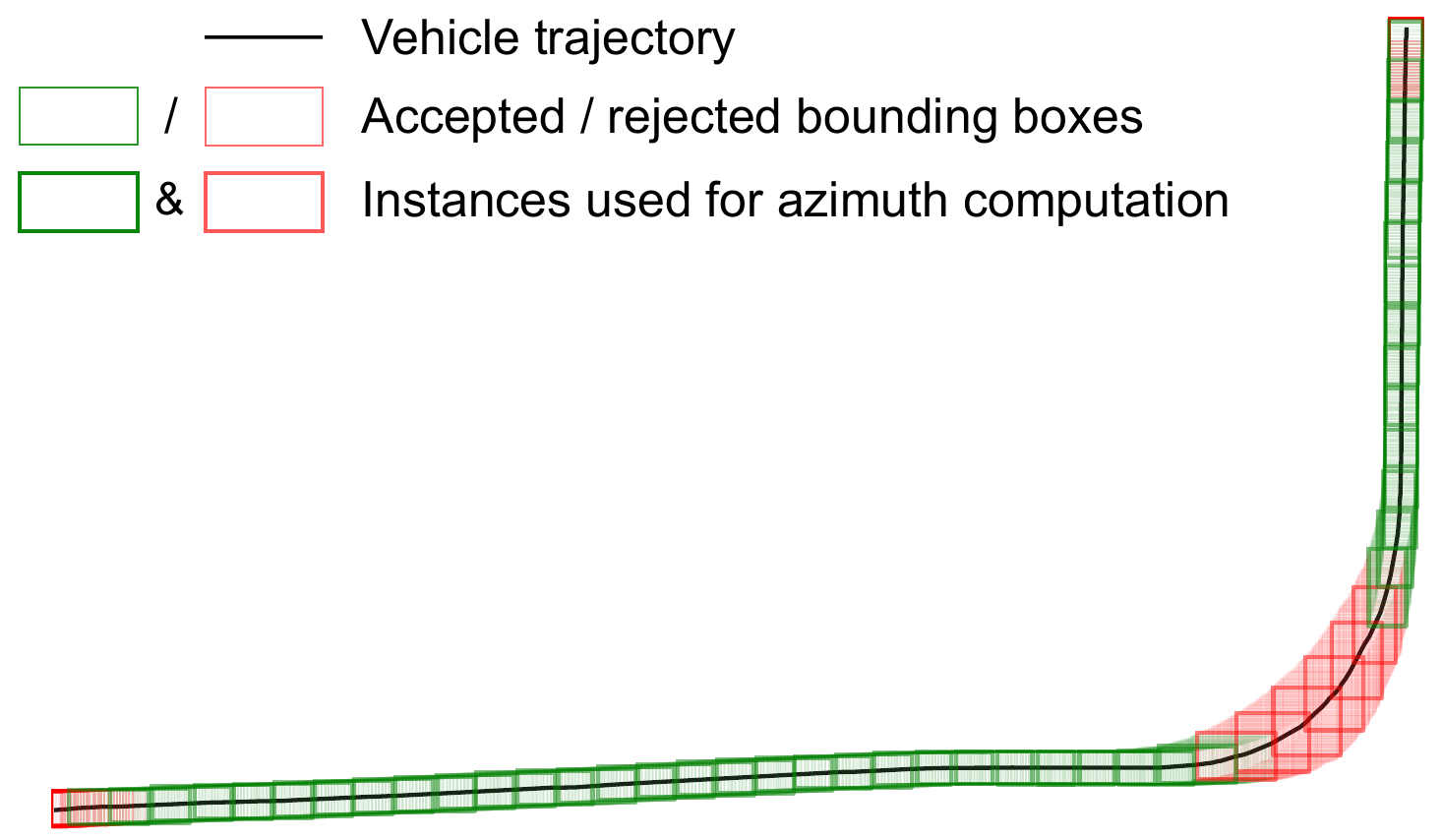}
  \hspace{0.01\columnwidth}
  \includegraphics[height=0.29\columnwidth]{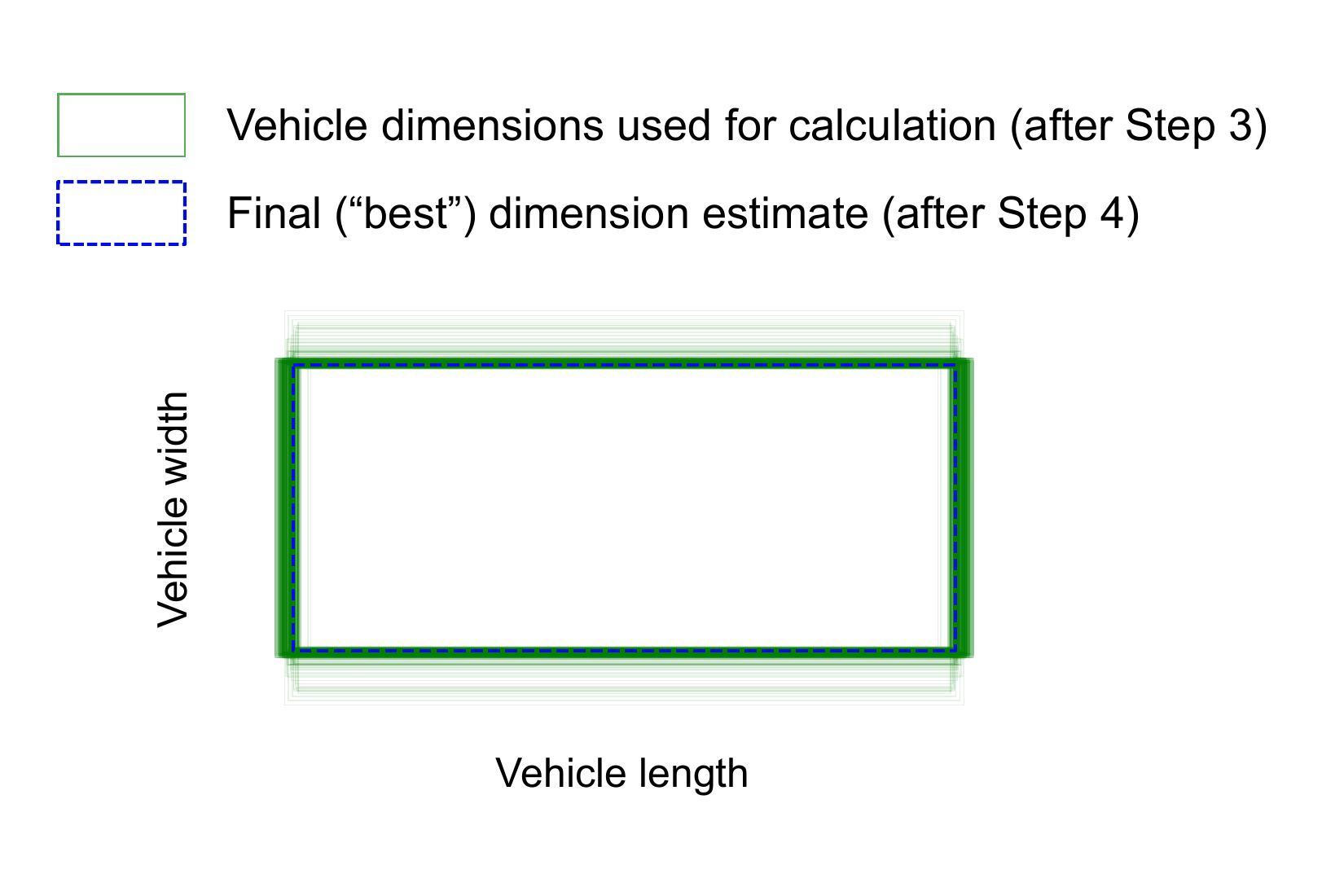}
  \caption{Illustration of the vehicle dimension estimation process for moving vehicles.}
  \label{fig:dimension_calculation}
\end{figure}

\hyperref[fig:dimension_calculation]{Figure~\ref*{fig:dimension_calculation}} illustrates the \ac{BB} filtering process, using actual \acp{BB} obtained from a sample trajectory. The figure also shows the final dimension estimate, computed as the first quartile of all ``accepted'' \acp{BB}. The right subfigure demonstrates the consistency of detected \acp{BB} along quasi-straight trajectories, with only minor inflations due to the permissible azimuth angle tolerance ($\bar\theta=15^\circ$) and deviations in road alignment relative to the reference drone's orientation.

It is important to acknowledge certain limitations of our approach. The dimensions derived from \acp{BB} in pixel coordinates may slightly overestimate the actual real-world vehicle sizes, often due to the \ac{BB} encapsulating protruding parts, such as rear-view mirrors, or strong shadows. The method also faces challenges when vehicles travel exclusively on non-parallel roads or remain parked or stationary out of alignment with the image axes. In these cases, as well as when persistent partial occlusions distort the \acp{BB}, no dimension estimates are calculated.

Despite these challenges, our approach provides a robust solution for vehicle dimension estimation from drone footage. By addressing the complexities of real-world scenarios and incorporating solutions for both moving and stationary vehicles, this method contributes to advancing the field of traffic analysis in drone-based applications.

\subsubsection{Speed and acceleration computation}\label{sec:kinematics}

To maintain the integrity of the extracted trajectories, we refrain from applying smoothing or interpolation techniques on the raw data, thereby preserving its original measurement characteristics. This approach allows users to select the most suitable post-processing methods for their specific needs. However, to facilitate efficient querying of trajectories based on predefined kinematics thresholds, we provide speed and acceleration estimates, which are computed after applying linear interpolation to fill in any potentially missing trajectory points, along with a simple smoothing technique to ensure reliable computation of the speed and acceleration profiles.

We apply a carefully tuned Gaussian filter to smooth the speed data, reducing noise while retaining critical dynamic patterns. The speed tuning process was guided by \ac{AV} data (see \autoref{sec:av}) to ensure accurate real-world vehicle motion representation. The details of the vehicle kinematics computation and tuning are provided in \ref{sec:appendix_kinematics}. The primary purpose of the smoothed speed and acceleration estimates is to facilitate efficient querying of trajectories based on kinematic thresholds, such as filtering out stationary vehicles. These estimates are not intended to represent definitive kinematic profiles for detailed traffic analysis. Our approach allows for future exploration of more advanced methods, such as those employed in~\cite{venthuruthiyil2020vehicle}. The availability of raw trajectory data in our dataset allows users to implement custom filtering techniques, such as the Kalman filter~\cite{kalman1960new}, or the \ac{RTS} smoother~\cite{rauch1965maximum}, tailored to their specific needs and applications.

\subsubsection{Construction of the Songdo Traffic dataset}\label{sec:songdo_dataset_creation}

The Songdo Traffic dataset \cite{songdo_traffic_dataset} is compiled from data collected by 10 drones monitoring 20 intersections over four days, encompassing 688,840 unique vehicle trajectories. These trajectories are preserved at the original camera capture rate of 29.97 \ac{FPS}, ensuring high temporal resolution without downsampling. The extraction and processing of trajectories were conducted first on a per-cut-video basis. As detailed in \autoref{sec:data_wrangling}, the video segmentation methodology systematically associates each video with its corresponding day, flight session, intersection, drone, and hovering sequence, resulting in over 3,000 individual \ac{CSV} files.

To streamline data management, trajectories were aggregated by intersection and flight session, with filenames encoding key metadata such as the collection date, intersection label, and session \ac{ID}. For example, \texttt{2022-10-04\_S\_AM1.csv} contains all extracted trajectories from the first morning session of the first day at the intersection 'S'. A detailed description of the dataset columns is provided in \autoref{tab:dataset_columns}. The aggregation process consolidated the dataset into 800 \ac{CSV} files, representing the complete Songdo Traffic dataset. These files were then packaged into 80 ZIP archives, one per intersection and experiment day, resulting in a total compressed trajectory dataset size of 12.21 GB. Each ZIP archive contains all 10 flight sessions for a given intersection and day (e.g., \texttt{2022-10-04\_S.zip}).

\begin{table}[htbp]
\footnotesize
\centering
\caption{Descriptions of columns in the Songdo Traffic dataset, detailing their format, data type, and interpretation. Floating-point values are reported with specified decimal places (d.p.).}
  \label{tab:dataset_columns}
  \begin{tabular}{llll}
    \hline
    Dataset Column Name & Format / Units & Data Type & Explanation\\
    \hline
    \verb|Vehicle_ID| & 1, 2, ... & Integer & Unique vehicle identifier within each CSV file \\
    \verb|Local_Time| & \texttt{hh:mm:ss.sss} & String & Local Korean time (GMT+9) in ISO 8601 format \\
    \verb|Drone_ID| & 1, 2, ..., 10 & Integer & Unique identifier for the drone capturing the data \\
    \verb|Ortho_X|, \verb|Ortho_Y| & px (1 d.p.) & Float & Vehicle center coordinates in the orthophoto cut-out image \\
    \verb|Local_X|, \verb|Local_Y| & m (2 d.p.) & Float & KGD2002 / Central Belt 2010 planar coordinates (EPSG:5186) \\
    \verb|Latitude|, \verb|Longitude| & ° DD (7 d.p.) & Float & WGS84 geographic coordinates in decimal degrees (EPSG:4326) \\
    \verb|Vehicle_Length|*, \verb|Vehicle_Width|* & m (2 d.p.) & Float & Estimated physical dimensions of the vehicle \\
    \verb|Vehicle_Class| & Categorical (0–3) & Integer & Vehicle type: 0 (car/van), 1 (bus), 2 (truck), 3 (motorcycle) \\
    \verb|Vehicle_Speed|* & km/h (1 d.p.) & Float & Estimated speed computed from trajectory data using~\eqref{eq:gaussian_smoothing} \\
    \verb|Vehicle_Acceleration|* & m/s$^2$ (2 d.p.) & Float & Estimated acceleration derived from smoothed speed values \\
    \verb|Road_Section|* & \texttt{N\_G} & String & Road section identifier (N = node, G = lane group) \\
    \verb|Lane_Number|* & 1, 2, ... & Integer & Lane position (1 = leftmost lane in the direction of travel) \\
    \verb|Visibility| & 0/1 & Boolean & 1 = fully visible, 0 = partially visible in the camera frame \\
    \hline
    \multicolumn{4}{l}{\small * These columns may be empty under certain conditions (see \autoref{sec:ortho_creation}, \autoref{sec:vehicle_dimensions}, \autoref{sec:kinematics}).}
  \end{tabular}
\end{table}

To minimize noise from detection anomalies, trajectories with 15 or fewer points are excluded. For uninterrupted tracking, this filtering corresponds to approximately 0.5 seconds of observation, ensuring that only meaningful motion patterns are retained. Additionally, a balance between file size and retained precision has been carefully considered: local and geographical coordinates are rounded to retain centimeter-level accuracy, while orthophoto coordinates are rounded to preserve sub-pixel granularity. For a detailed discussion of known dataset artifacts and limitations, see \ref{sec:appendix_artifacts}.

Beyond trajectory data, Songdo Traffic also includes the 20 high-resolution orthophoto cut-outs (8,000 $\times$ 8,000 pixels), complemented by corresponding road section and lane segmentations. These cut-outs provide essential spatial context, enabling seamless integration of vehicle pixel coordinates with roadway geometry for enhanced visualization. Additionally, the dataset features 29 video samples, each capturing the first 60 seconds of drone activity over its designated intersections during the final session (PM5) of the last day (2022-10-07) of the experiment. These supplementary assets enhance the dataset’s utility for detailed traffic analysis and visualization.

\section{Results}\label{sec:results}
\subsection{Traffic data extraction}\label{sec:traffic_data_extraction_results}

\begin{figure}[htbp]
  \centering
  \includegraphics[height=0.355\textwidth, valign=t]{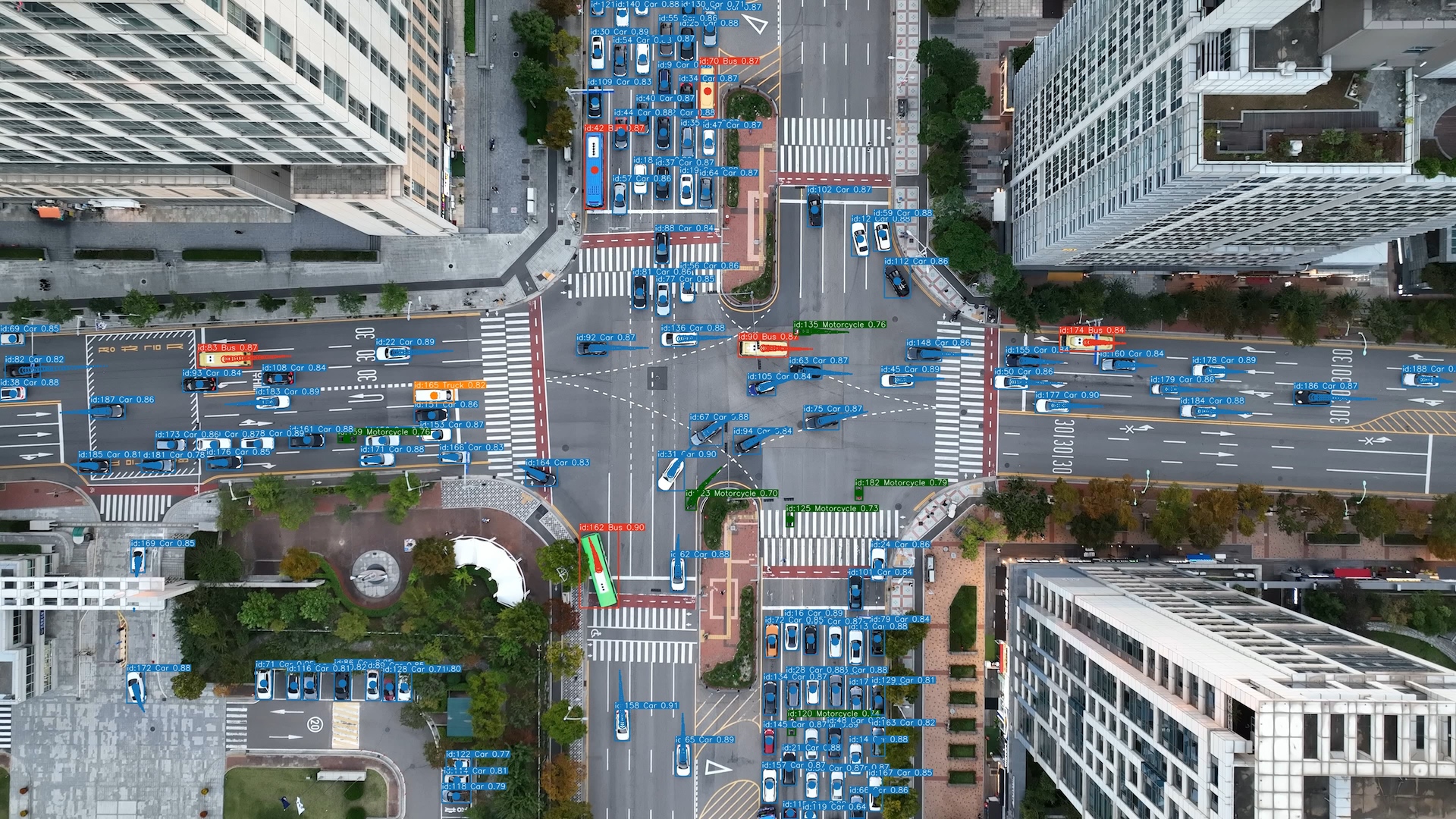}
  \hspace{0.5mm}
  \includegraphics[height=0.355\textwidth, valign=t]{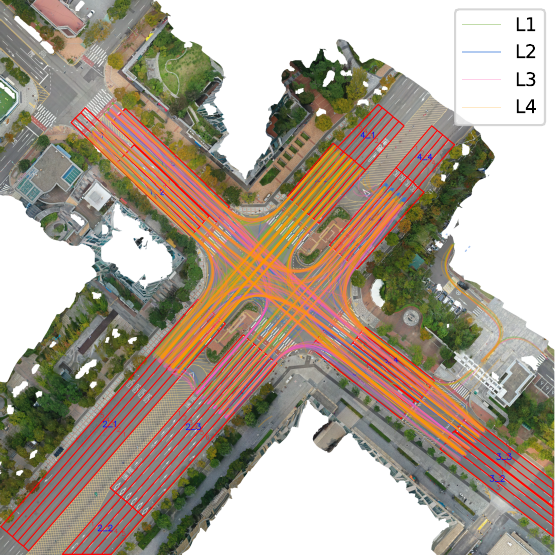}
  \caption{Left: Illustration of vehicle detection and tracking. Right: Nearly 1,200 vehicle trajectories extracted from four separate drone hovering sessions, overlaid on the segmented orthophoto cut-out.}
  \label{fig:tracking_and_trajectory_illustration}
\end{figure}

We present key characteristics of the georeferenced trajectories extracted using our complete methodology, focusing on the final day of the experiment (Friday) and the final afternoon session (PM5), which commenced at 17:40, representing one of the most congested periods. Specifically, we analyze intersection L, depicted in \autoref{fig:tracking_and_trajectory_illustration} (left), where the horizontal road has a speed limit of 30 km/h and the vertical road has a speed limit of 50 km/h. This intersection was monitored by both drones D6 and D7, however, we limit our analysis to D7, which maintained an altitude of 150~m.

During this period, D7 recorded a total of four video segments ($v \in \{\text{L1}, \text{L2}, \text{L3}, \text{L4}\}$), each corresponding to an individual hovering, with a combined duration of approximately 6 minutes and 55 seconds. From these videos, we extracted 1,194 unique vehicle trajectories of varying lengths. The effectiveness of our detection and tracking approach is demonstrated in \autoref{fig:tracking_and_trajectory_illustration} (left), where color-coded \acp{BB} indicate accurate vehicle identification and classification: blue for cars, red for buses, orange for trucks, and green for motorcycles. The trailing tails represent the last second of vehicle movement, providing insights into speed and stopping behavior.

The aggregated trajectories in the orthophoto cut-out \ac{CS}, shown in \autoref{fig:tracking_and_trajectory_illustration} (right), illustrate the ability of our framework to transform multiple separate video feeds into consistent georeferenced trajectories. This visualization highlights the robustness of our methodology for analyzing dynamic traffic flows in congested urban environments and emphasizes the contributions of the Songdo Traffic dataset in providing detailed insights into urban mobility.

\begin{figure}[htbp]
  \centering
  \includegraphics[width=0.503\textwidth, valign=t]{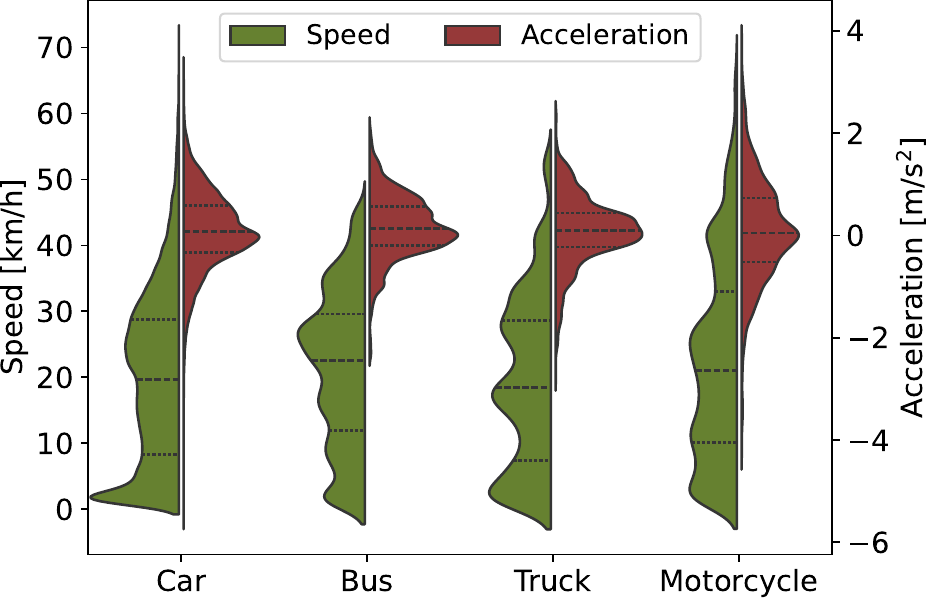}
  \hspace{3mm}
  \includegraphics[width=0.468\textwidth, valign=t]{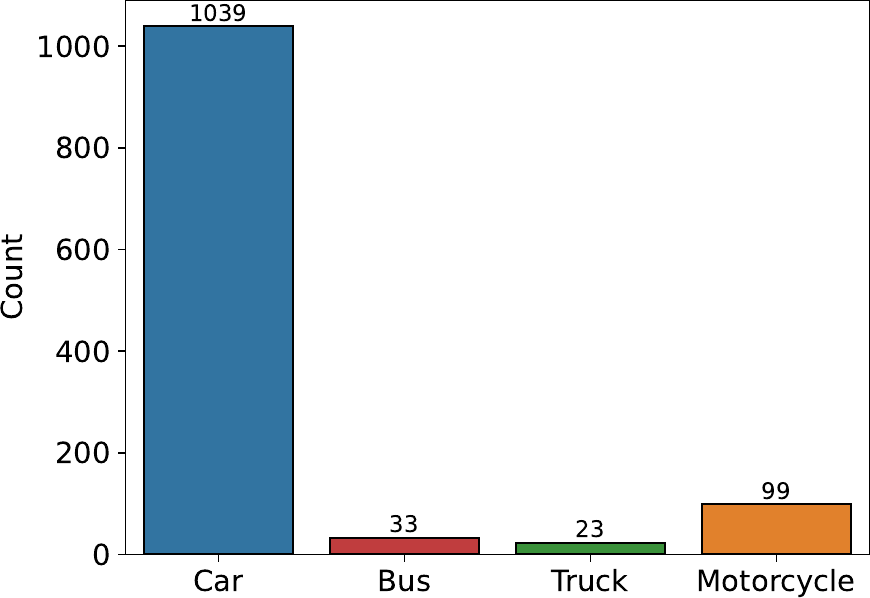}
\caption{Left: Speed and acceleration distributions for each vehicle class. Right: Distribution of vehicle classes.}
  \label{fig:speed_acc_distributions}
\end{figure}

Unlike previous UAV-based extraction methodologies, our approach yields realistic acceleration and speed distributions, as shown in \autoref{fig:speed_acc_distributions} (left). Speed values are filtered to exclude those below 1 km/h for better plot legibility. The corresponding vehicle class distribution is shown in \autoref{fig:speed_acc_distributions} (right).
Notably, the speed distributions resemble realistic patterns for an intersection with 30 km/h and 50 km/h road limits, however, the significant congestion during this period results in limited probability mass near the 50 km/h mark. Additionally, the acceleration distributions align closely with those reported in fixed camera setups~\cite{venthuruthiyil2020vehicle}, reflecting the effectiveness of our track stabilization. In contrast, previous \ac{UAV}-based trajectory extraction methods often exhibit large and unrealistic acceleration variances with heavy tails, often reaching up to ±300 m/s$^2$, mainly due to camera shakiness and inadequate video stabilization~\cite{venthuruthiyil2020vehicle}.

\begin{figure}[htbp]
  \centering
  \includegraphics[width=0.485\textwidth, valign=t]{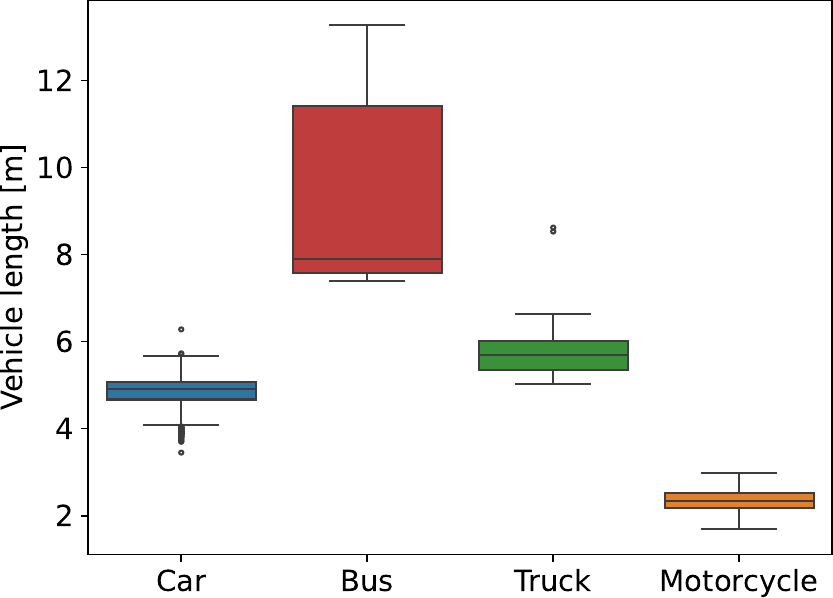}
  \hspace{3mm}
  \includegraphics[width=0.485\textwidth, valign=t]{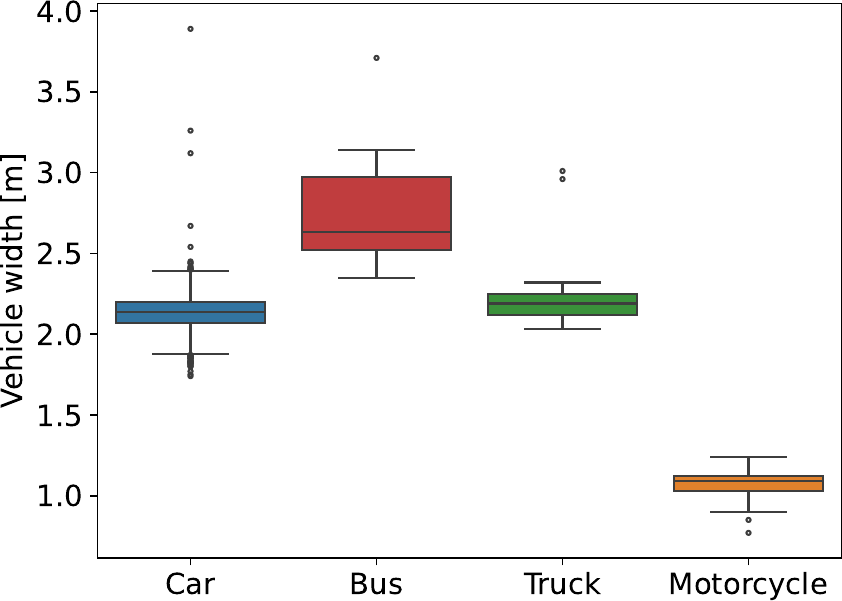}
  \caption{Estimated real-world vehicle length and width distributions.}
  \label{fig:dimension_distributions}
\end{figure}

The estimated vehicle dimensions are depicted in \autoref{fig:dimension_distributions}, which shows the length and width distributions for different vehicle classes. Notably, the median length and width for the car category are approximately 4.9 m and 2.1 m, respectively, which align closely with mean vehicle dimensions reported in previous studies. For example, a study on passenger car external dimensions found that the mean length of sedans is 5.1 m and the mean width is 1.89 m, while SUVs have an mean length of 4.95 m and an mean width of 1.94 m~\cite{zhang2022probability}. The significant differences in length and width between buses and motorcycles also demonstrate the effectiveness of our dimension estimation approach.

To further validate our dimension estimation approach, we leveraged the \ac{AV} experiment, which is detailed in \autoref{sec:av}. The \ac{AV} used in this experiment is a Hyundai IONIQ 5 compact crossover SUV, with known dimensions of 4.635 m in length and 1.890 m in width (2.116 m with side-view mirrors). Its distinctive Stanford logo made it easily identifiable in our drone footage. Our estimates, based on 15 video observations (see \autoref{fig:all_captured_AV_trajectories}), yielded a length distribution of $4.931\pm 0.225$ m and a width distribution of $2.15 \pm 0.072$ m. While the width estimate is relatively accurate, length estimates were often inflated due to strong shadows present in the majority of videos. This is confirmed by the lower quartile of the length distribution (4.685 m), which is closer to the \acl{GT} value (4.635 m).

We also conducted a similar analysis using a public transportation bus, specifically a Hyundai Super AeroCity model year 2018, which was easily identifiable in the drone footage due to its distinct color and shape. The bus, depicted in \autoref{fig:tracking_and_trajectory_illustration} (blue bus, top-center), has known dimensions of $10.955 \times 2.490$ m (excluding wing mirrors). From 12 videos, 6 from morning and 6 from afternoon sessions, captured at 2 intersections, we extracted 20 distinct instances of this bus and estimated their dimensions. Our results showed a length distribution of $11.540\pm 0.227$ m and a width distribution of $3.063\pm 0.179$ m, indicating that the estimates were slightly inflated due to the bus’s protruding mirrors, shadows, and height-related perspective distortions. The wing mirrors, each 21 cm wide, likely contribute to the width discrepancy, bringing the total width to approximately $2.910$ m, closer to our estimated width distribution.

To further assess our approach, we implemented a naive baseline that computes non-stationary vehicle dimensions by directly averaging the longer and shorter sides of all \acp{BB}, without any azimuth-based filtering. For the \ac{AV} (Hyundai IONIQ 5) cases, where trajectories are mostly parallel to the image axes (see \autoref{fig:all_captured_AV_trajectories}), the naive method estimated $5.005 \pm 0.246$ m (length) and $2.341 \pm 0.223$ m (width). This parallel alignment inherently favors the naive approach, yet the higher width variability indicates reduced reliability. A similar pattern emerged for the Hyundai Super AeroCity bus, where trajectories were also predominantly parallel, with the naive method estimating $11.585 \pm 0.218$ m (length) and $3.645 \pm 0.416$ m (width). The naive width estimate is significantly inflated compared to our results, reinforcing that width estimation is particularly sensitive without proper filtering. These findings suggest that while the naive approach may perform reasonably well for simple, axis-aligned trajectories, our method becomes crucial when dealing with more realistic scenarios involving frequent non-parallel vehicle motions.

\begin{figure}[htbp]
  \centering  
  \includegraphics[width=1\columnwidth]{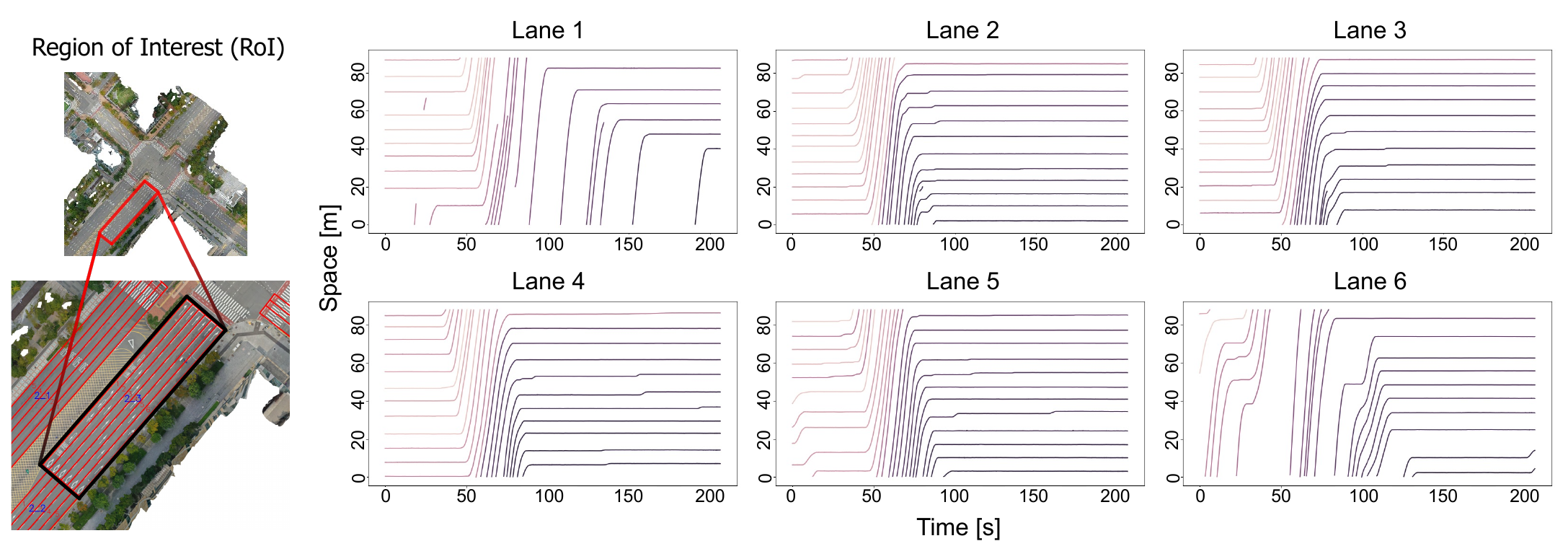}
  \caption{\acl{ST} diagram example for intersection L, road segment 2\_3 and lanes 1--6 within the depicted \ac{RoI}.}
  \label{fig:st_diagram}
\end{figure}

The \acl{ST} diagram in \autoref{fig:st_diagram} visualizes traffic flow dynamics at intersection L, highlighting the temporal progression of vehicle movements and queue formation and dissipation across six lanes. To maintain legibility, the diagram shows trajectories from a single video. Notable patterns include distinct queue buildup and clearance in the left-turn lane (Lane 1), consistent flow in straight-ahead lanes (Lanes 2-5), and dynamic stop-and-go behavior in the right-turn lane (Lane 6). Shorter trajectories, particularly in Lanes 1 and 2, correspond to motorcycles changing lanes or exiting the road section, causing intermittent traffic patterns. The inclusion of road section and lane-level information in our dataset enables users to create similar \acl{ST} analyses, facilitating granular studies of traffic behavior that are often limited by other datasets. A video demonstration of our extraction pipeline \cite{Fonod_Geo-trax_2025}, \textit{Geo-trax}, is available at \url{https://youtu.be/gOGivL9FFLk}.

\subsection{Comparison with RTK-GNSS data from the probe AV}\label{sec:av}

\begin{figure}[htbp]
  \centering
  \includegraphics[height=0.45\columnwidth, valign=t]{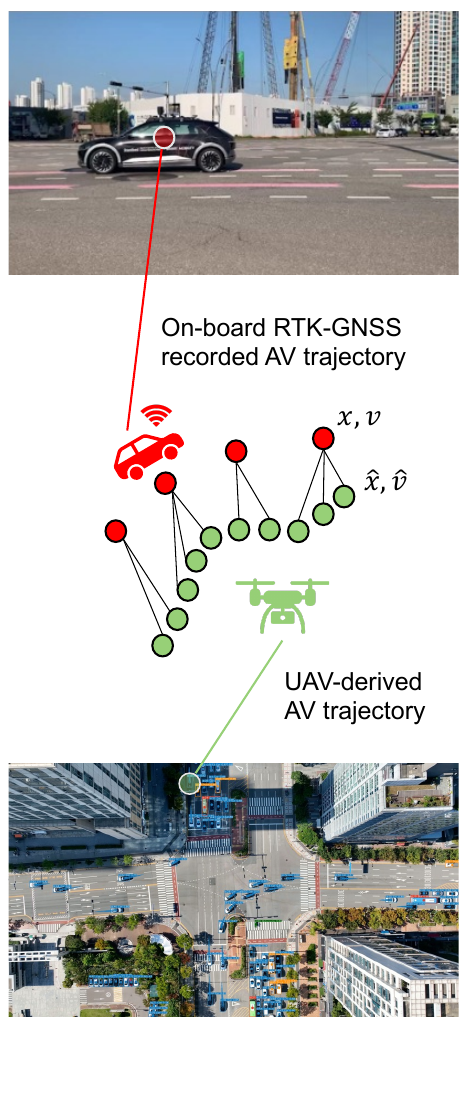}
  \hspace{8mm}
  \includegraphics[height=0.45\columnwidth, valign=t]{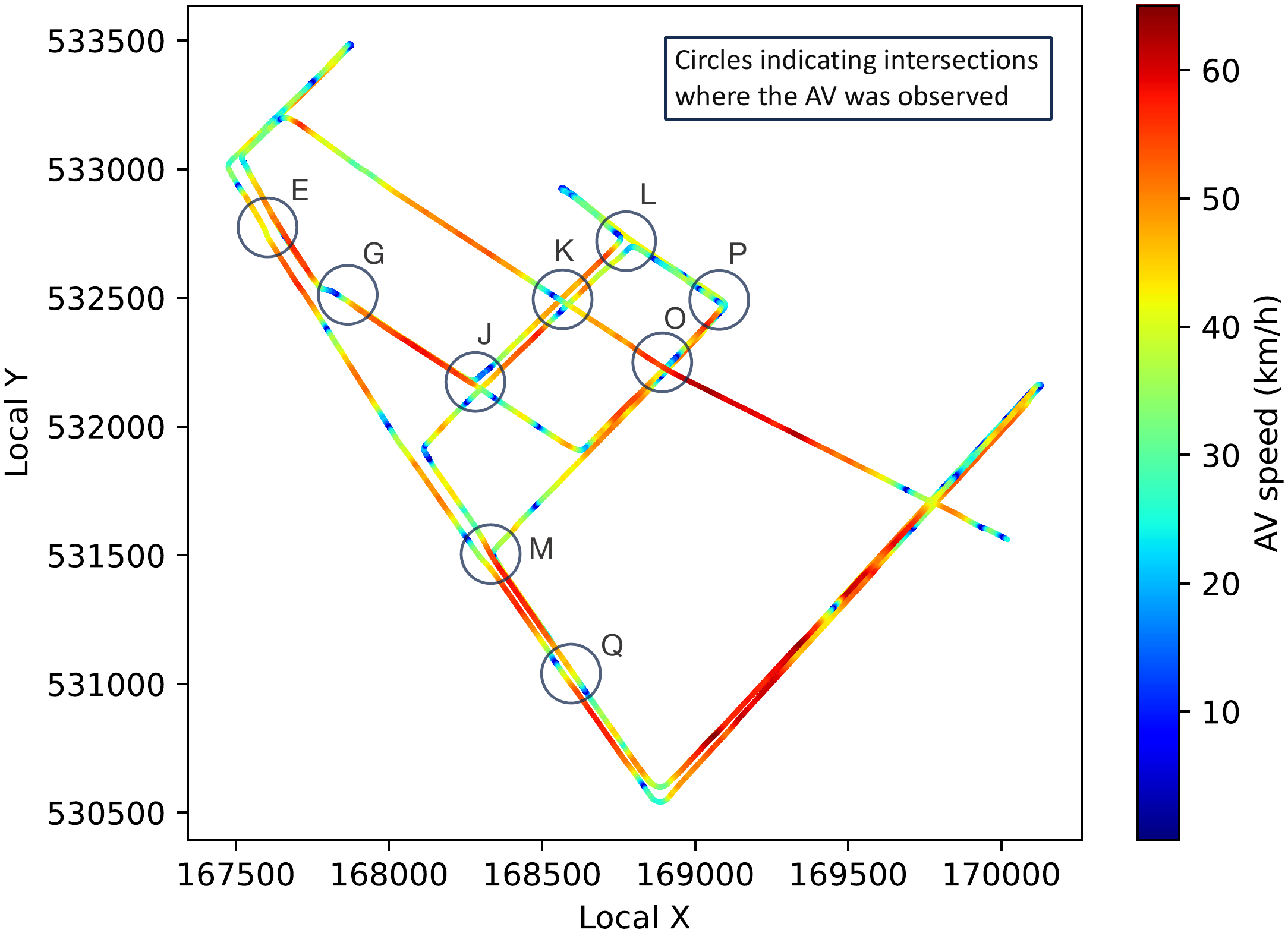}
  \caption{Left: Illustration of the two AV trajectory measuring principles. Right: On-board \ac{RTK-GNSS} recorded AV trajectory converted to local \ac{CS}, with marked intersections where our drones recorded the AV passing.}
  \label{fig:comparison_with_AV}
\end{figure}

To assess the reliability of our extraction framework, we compared the drone-derived trajectories and speed profiles with those recorded by the \ac{RTK-GNSS} sensor on the \ac{AV}. Conducted by \ac{SCIGC}, the \ac{AV} experiment coincided with our monitoring campaign, providing a unique validation opportunity. This comparison allowed us to evaluate the accuracy of our framework against an independent measurement source (see \autoref{fig:comparison_with_AV}, left).

\begin{figure}[htbp!]
  \centering
  \includegraphics[width=0.33\columnwidth]{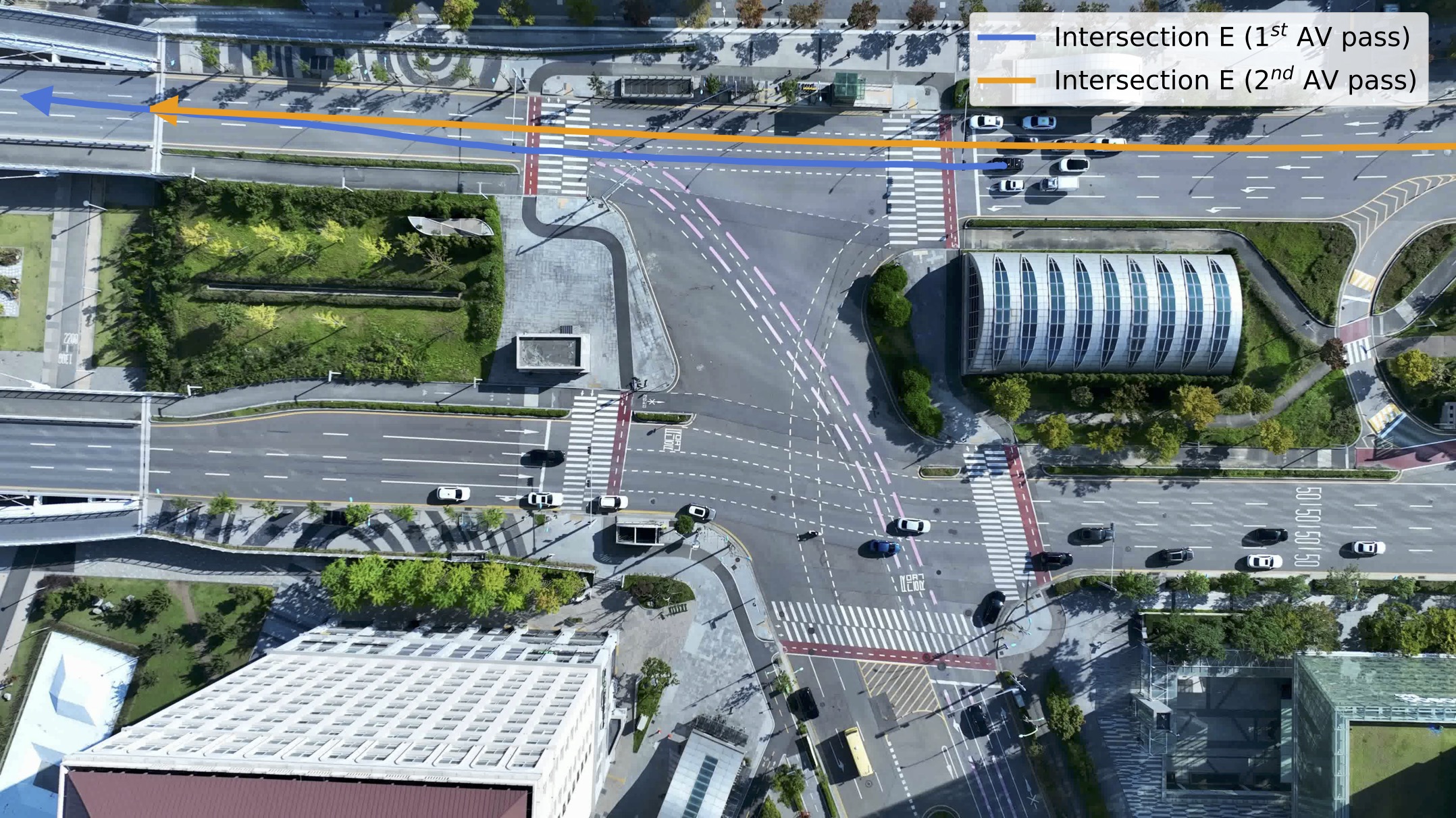}
  \includegraphics[width=0.33\columnwidth]{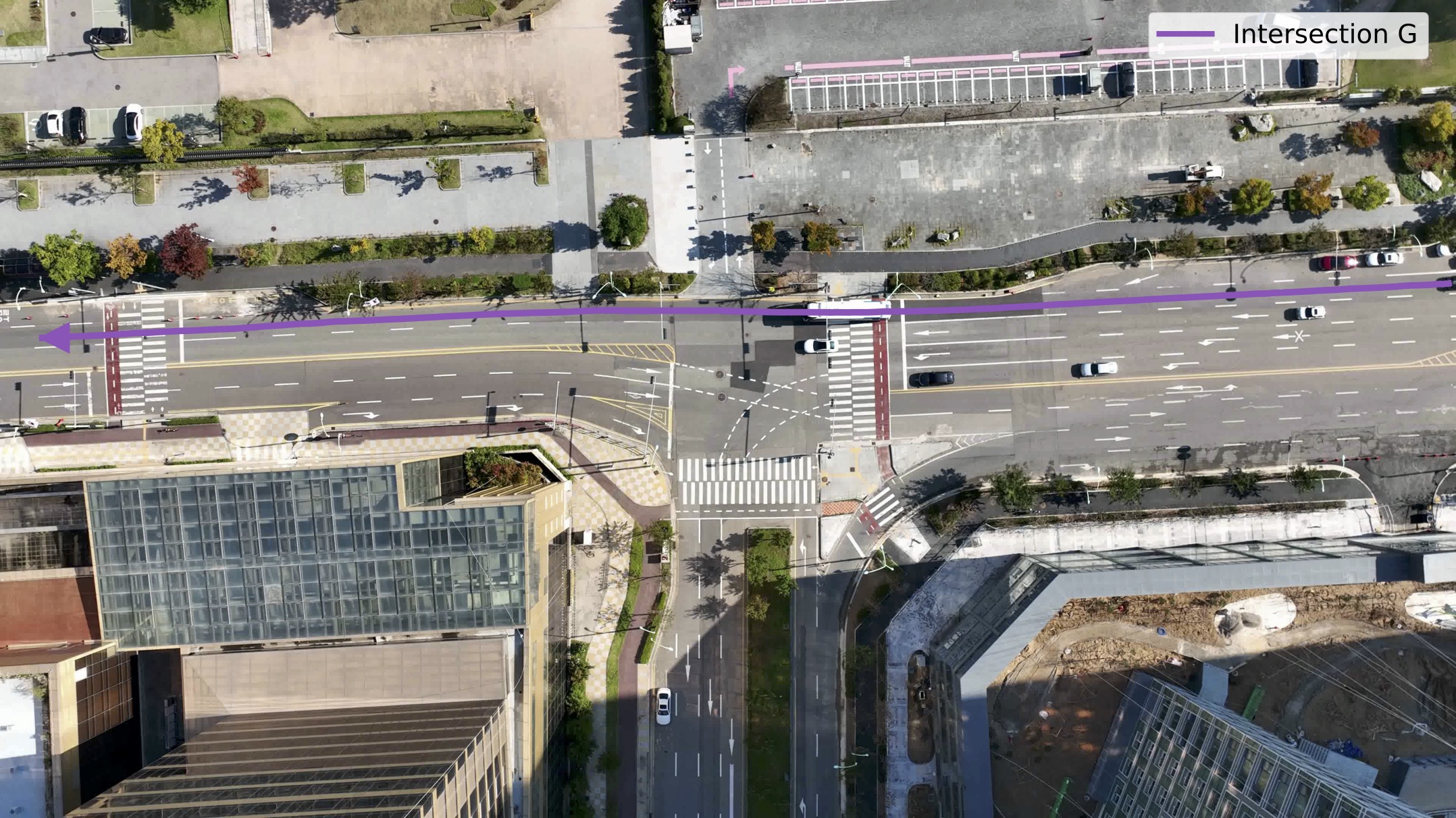}
  \includegraphics[width=0.33\columnwidth]{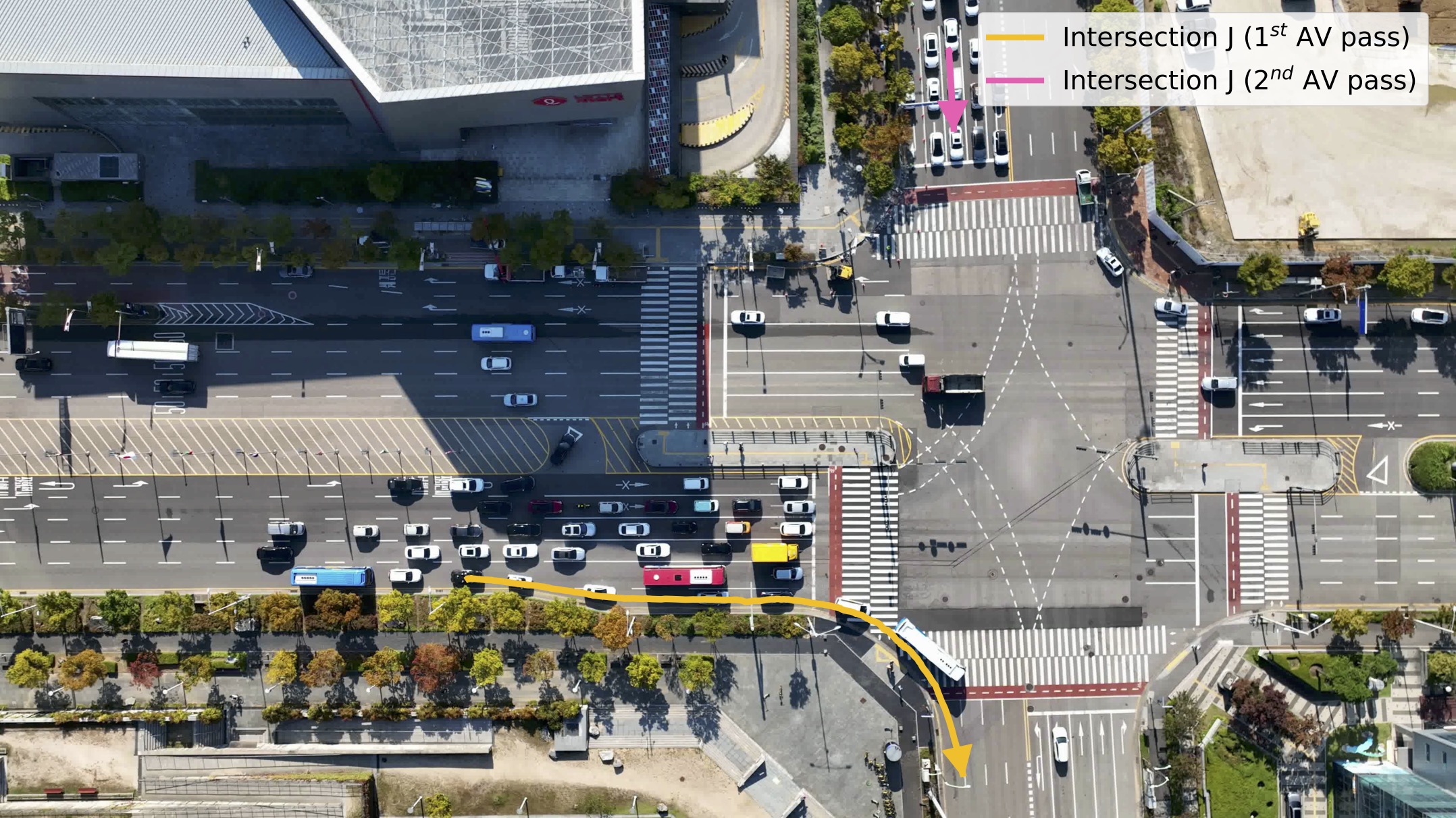}\\[0.4mm]
  \includegraphics[width=0.33\columnwidth]{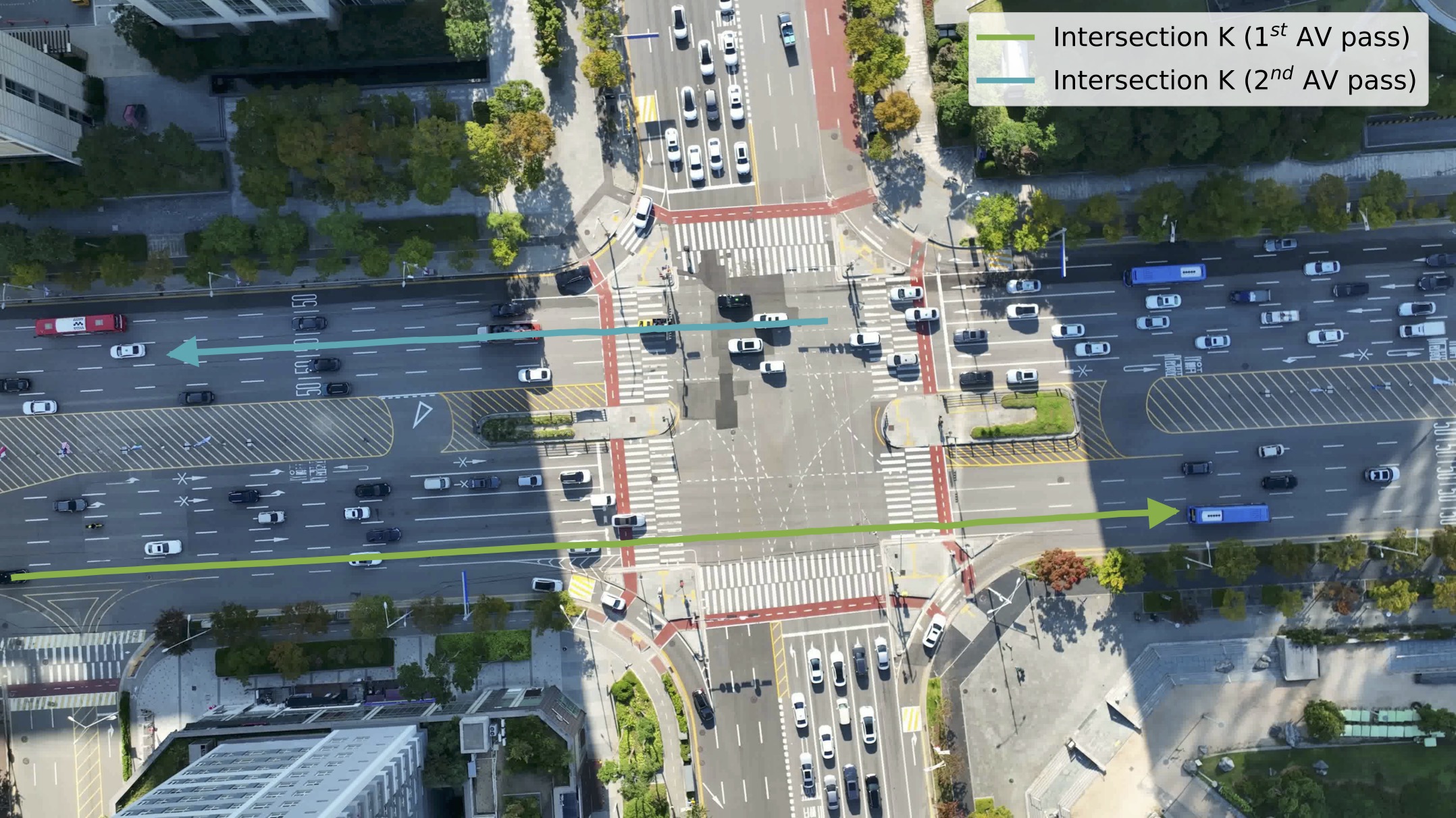}
  \includegraphics[width=0.33\columnwidth]{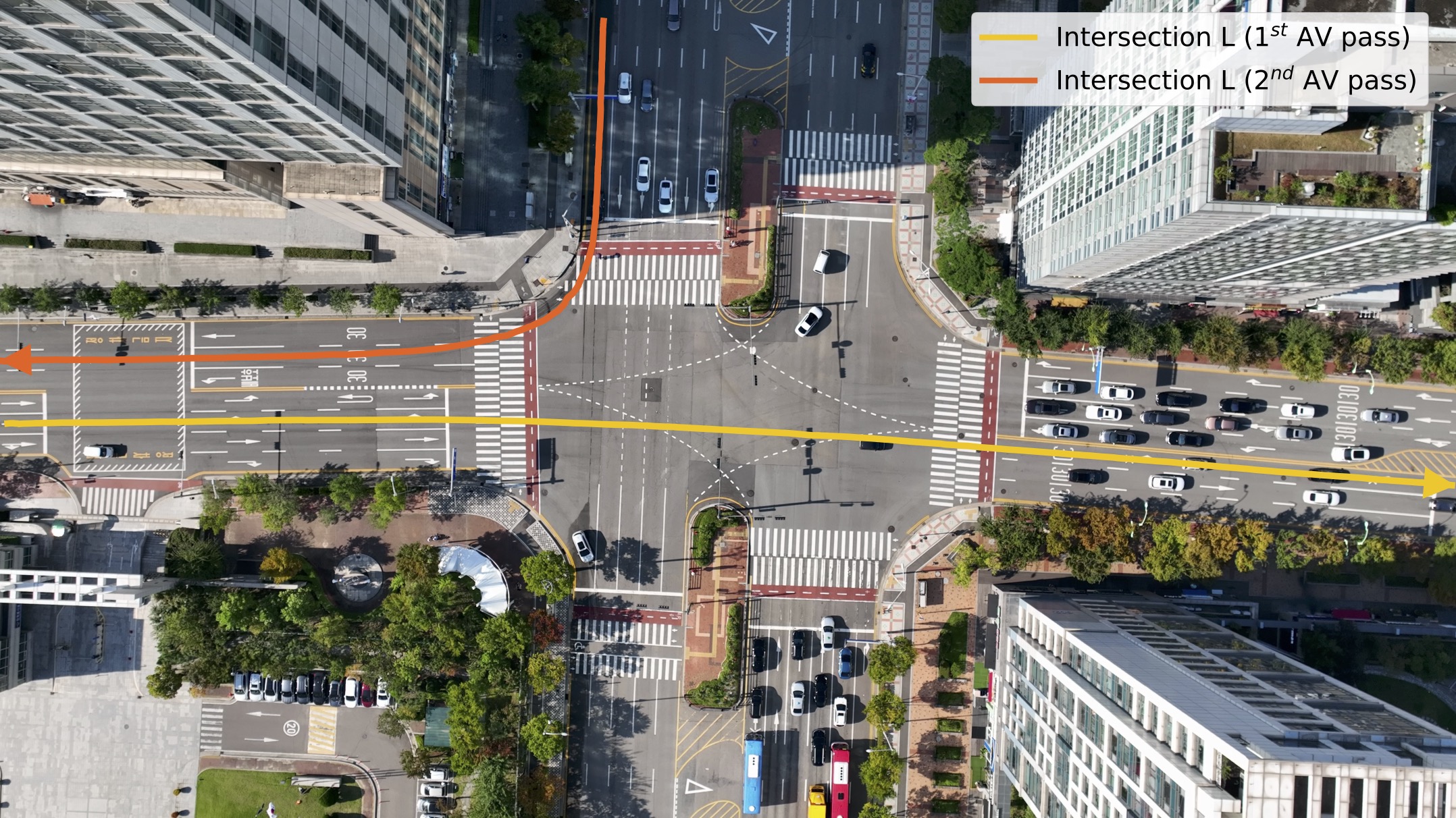}
  \includegraphics[width=0.33\columnwidth]{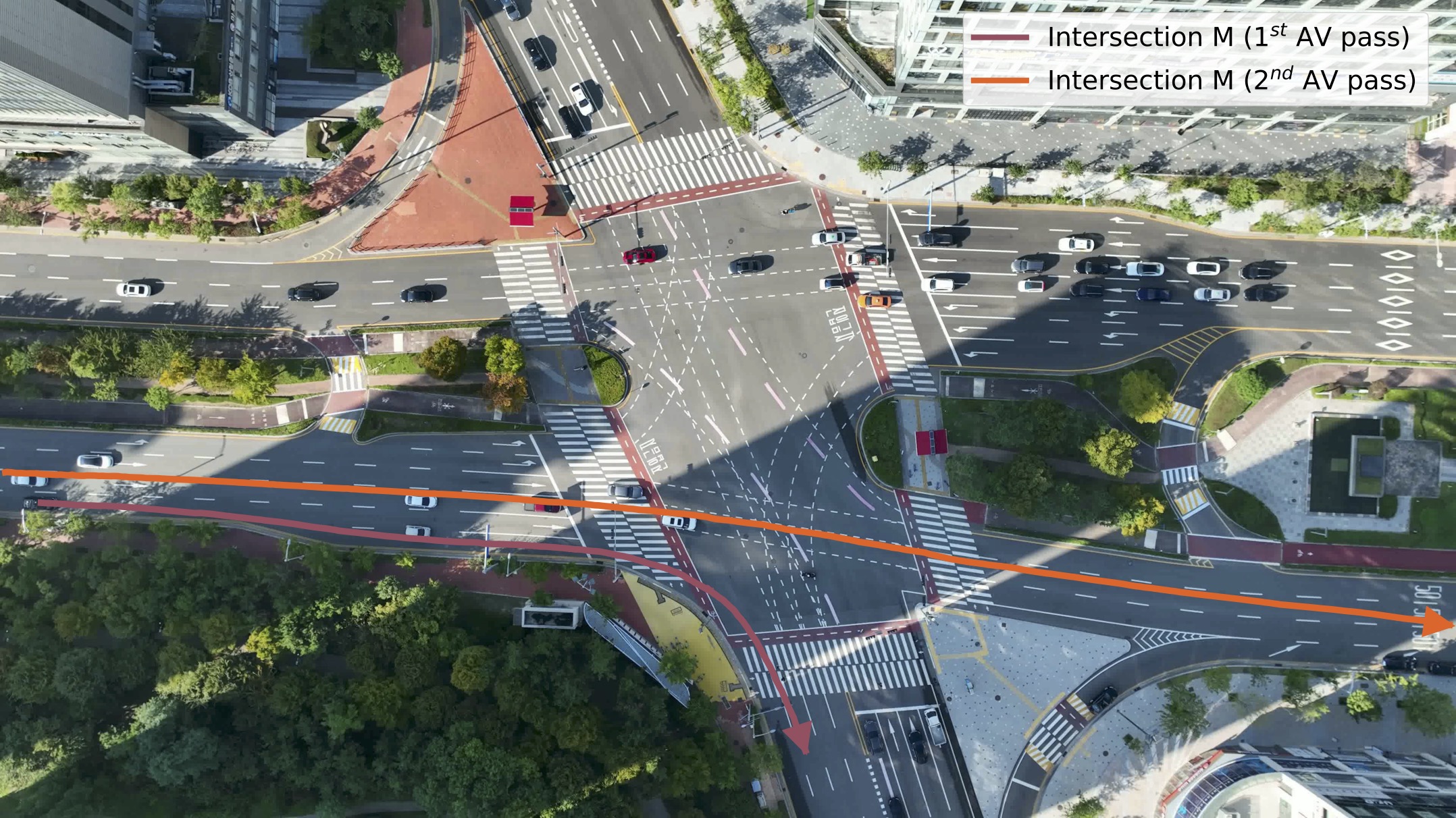}\\[0.4mm]
  \includegraphics[width=0.33\columnwidth]{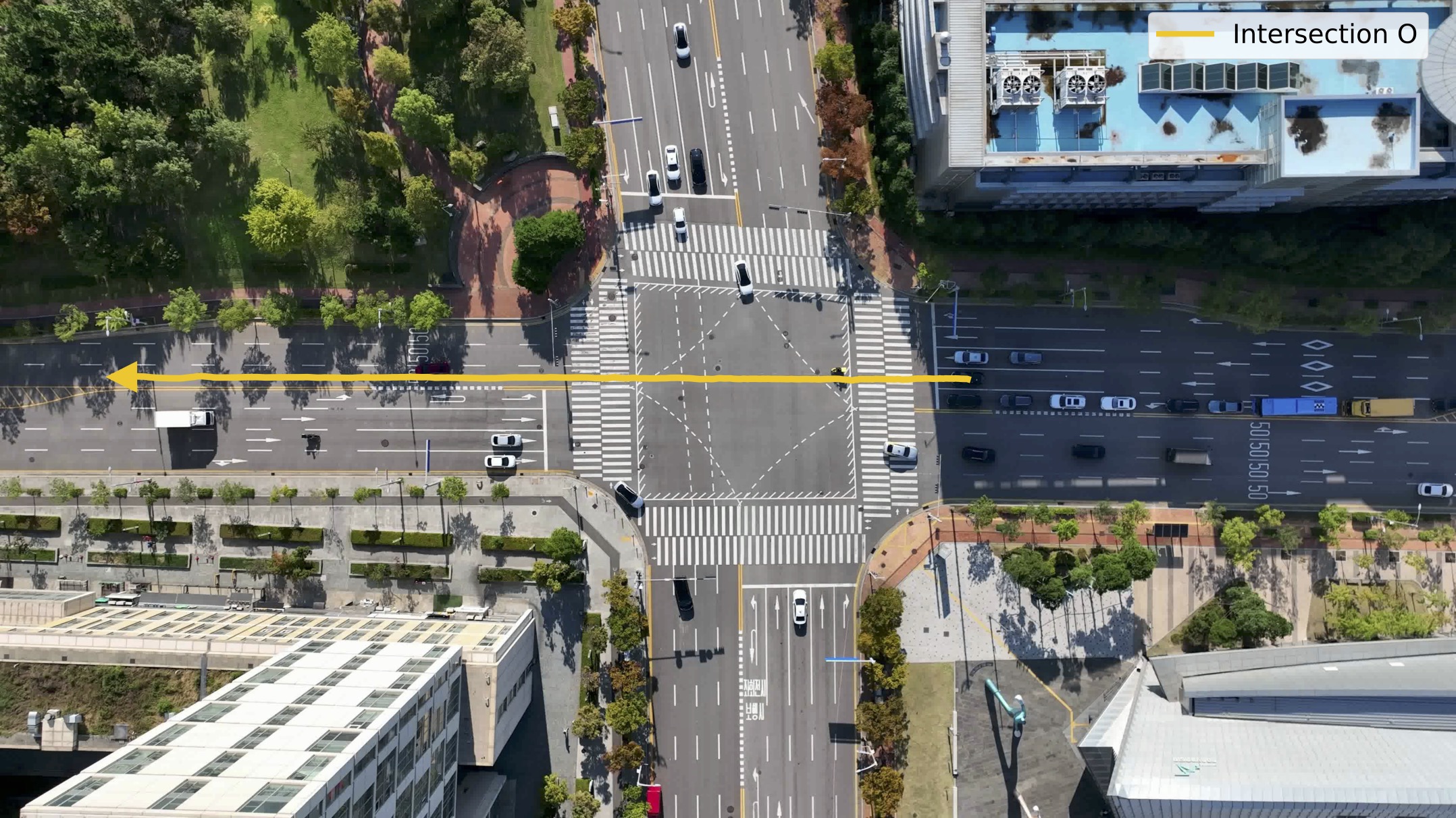}
  \includegraphics[width=0.33\columnwidth]{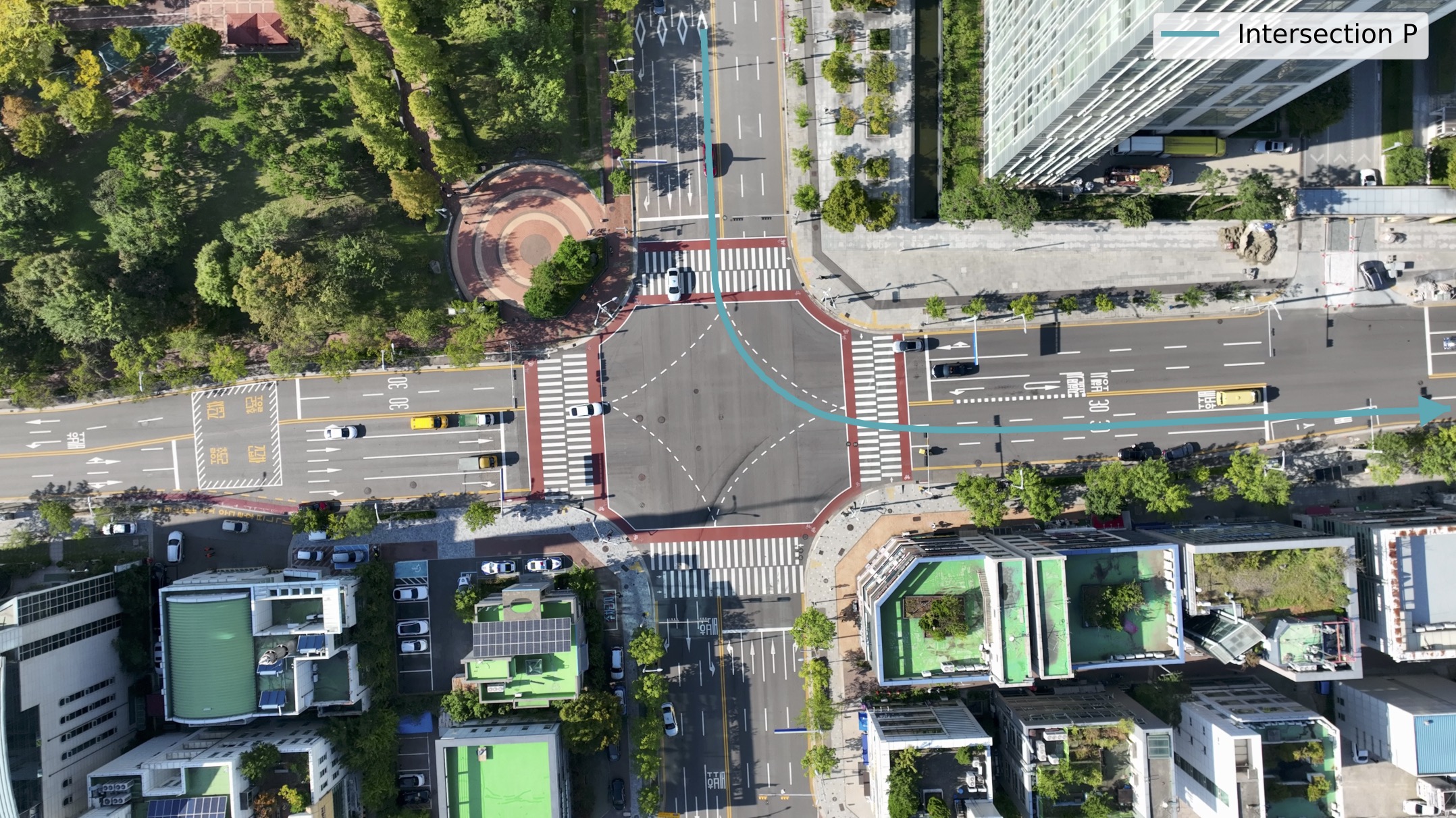}
  \includegraphics[width=0.33\columnwidth]{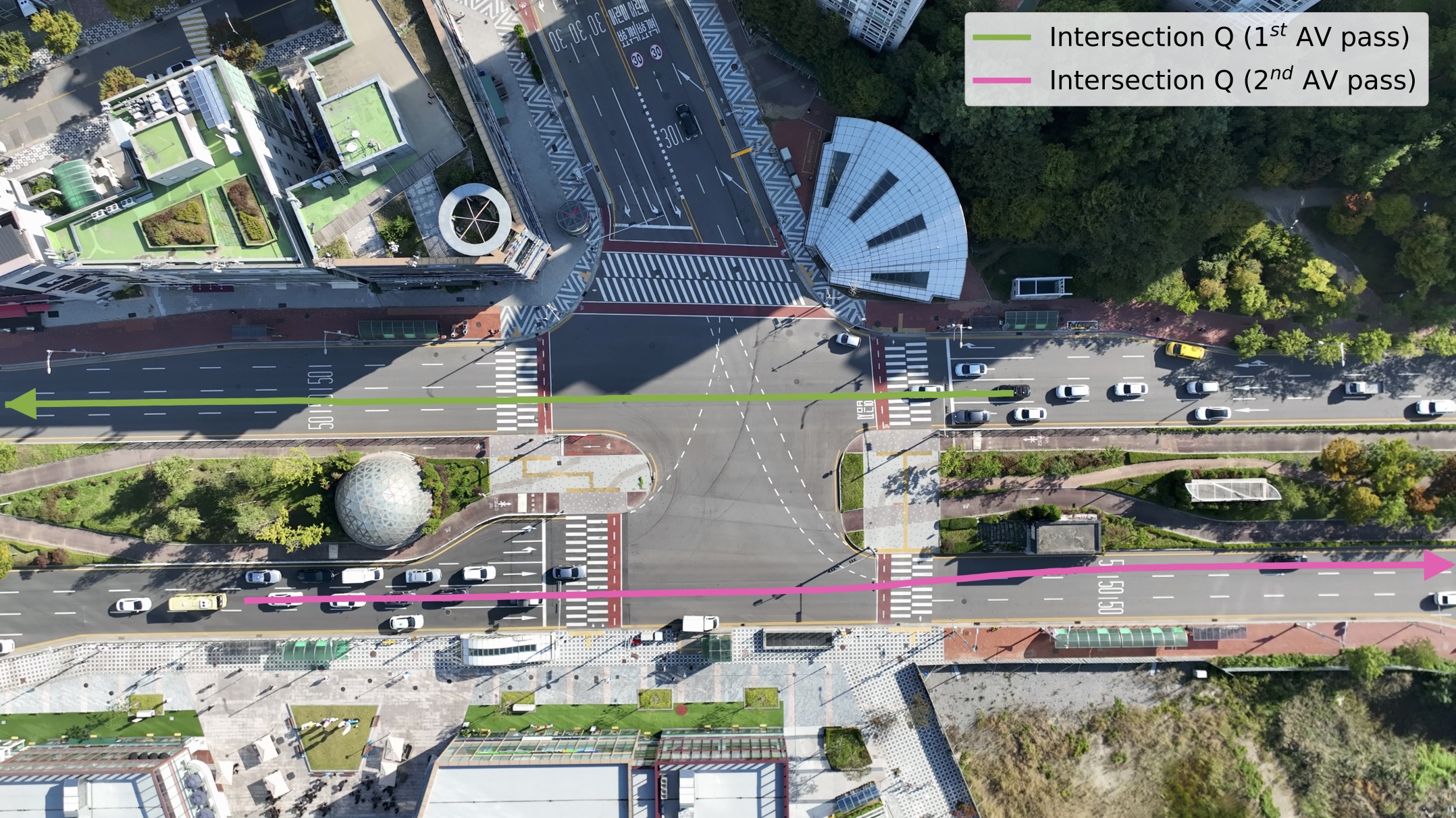}
  \caption{Trajectories of the AV observed by our drones.}
  \label{fig:all_captured_AV_trajectories}
\end{figure}

The \ac{AV} experiment commenced at 8:53:38 AM on the final day of our monitoring campaign and lasted approximately 1 hour and 10 minutes. \autoref{fig:comparison_with_AV} (right) shows the \ac{AV}’s full trajectory and speed profile. Given the dynamic nature of our drone operations, which alternated over multiple intersections during each flight session, our analysis here focuses exclusively on static hoverings above intersections. Consequently, we observed the \ac{AV} at nine intersections (see \autoref{fig:comparison_with_AV}), recording a total of fifteen distinct \ac{AV} passes, as illustrated in \autoref{fig:all_captured_AV_trajectories}.

Unfortunately, the timestamps of the \ac{AV} trajectory from the \ac{RTK-GNSS} sensor were not synchronized with our \acp{UAV}’ internal clocks. Additionally, the \ac{RTK-GNSS} data was asynchronously sampled and undersampled, averaging 10 Hz, while our trajectories were extracted at a fixed 29.97 \ac{FPS} rate. Due to these differences, we use separate time indices: $k_1$ for the \ac{RTK-GNSS} data and $k_2$ for our UAV-derived trajectory. This discrepancy makes direct Euclidean distance comparisons infeasible, necessitating a more suitable error measurement method~\cite{su2020survey}.

\begin{figure}[htbp!]
  \centering
  \includegraphics[width=0.5\columnwidth]{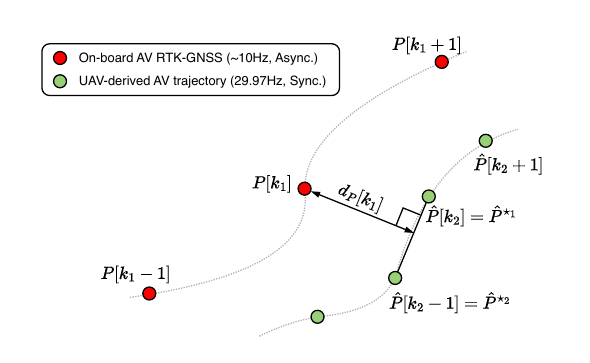}
  \caption{Illustration of positional deviation determination between \ac{RTK-GNSS} data point and UAV-derived trajectory.}\label{fig:distance_error_computation}
\end{figure}

In our methodology, we calculate the positional deviations between the two trajectories in the local \ac{CS} by measuring the distance $d_P[k_1]\geq 0$. This distance represents the perpendicular separation from the \ac{RTK-GNSS} trajectory point $P[k_1]=(x_L[k_1], \,y_L[k_1])$ to the line defined by the two closest trajectory points from our \ac{UAV}-derived data. For a visual illustration of this computation, see \autoref{fig:distance_error_computation}.

The absolute positional deviation $d_P[k_1]$ at discrete time $k_1$ is computed as follows:
\begin{equation}
    d_P[k_1] = 
    \left |
    \frac{(\hat P^{\star_2} - \hat P^{\star_1}) \times (\hat P^{\star_1} - P[k_1])}{\|\hat P^{\star_2} - \hat P^{\star_1} \|}
    \right |,
\end{equation}
where '$\times$` denotes the cross product. Here, $\hat{P}^{\star_1}$ is the point on the \ac{UAV}-derived trajectory that is closest to $P[k_1]$, and $\hat{P}^{\star_2}$ is the adjacent point to with the second shortest Euclidean distance to $P[k_1]$.

In a similar manner, we compute the speed difference $\Delta v[k_1]$ at $k_1$ as follows:
\begin{equation}\label{eq:speed_error}
    \Delta v[k_1] = v[k_1] - \left(w^{\star_1} \tilde v^{\star_1} + w^{\star_2} \tilde v^{\star_2}\right),
\end{equation}
where $\tilde v^{\star_i}$ is the smoothed speed estimate associated with $\hat P^{\star_i}$. The weights $0\leq w^{\star_i}\leq 1$ are given by: $w^{\star_1} =  d^{\star_2}/(d^{\star_1} + d^{\star_2})$ and $w^{\star_2} = d^{\star_1}/(d^{\star_1} + d^{\star_2})$, where $d^{\star_1} = \|\hat{P}^{\star_1} - P[k_1]\|$ and $d^{\star_2} = \|\hat{P}^{\star_2} - P[k_1]\|$ are the Euclidean distances between the on-board measurement $P[k_1]$ and the UAV-derived points $\hat{P}^{\star_1}$ and $\hat{P}^{\star_2}$, respectively.

We applied the above methodology to compute $d_P$ and $\Delta v$ for each of the captured \ac{AV} passes (illustrated in \autoref{fig:all_captured_AV_trajectories}). To ensure accurate calculations, we segmented the \ac{AV} videos to exclude any initial or final stationary periods, such that the start and end of each video segment capture the \ac{AV} in motion. Additionally, we filtered out speed difference values from $\Delta v[k_1]$ that corresponded to $v[k_1]\leq 1$ to prevent bias in our results. Finally, we aggregated the computed differences for each intersection to facilitate a clear and concise presentation of our findings.

The comparison results are summarized in \autoref{tab:av_error}. Overall, the \ac{UAV}-derived trajectories show strong agreement with \ac{RTK-GNSS} measurements, with mean positional deviations remaining below 1 meter at most intersections, except for L and P. Similarly, speed differences between the two methods are minimal, averaging within $\pm 1$ km/h, except at intersection M, where a larger discrepancy likely reflects the high speed variability observed during both \ac{AV} passes. The observed deviations represent differences rather than errors, as neither the \ac{UAV}-derived data nor the \ac{RTK-GNSS} measurements can be considered absolute ground truth. These differences may stem from variations in measurement methodologies, such as the location of the on-board \ac{RTK-GNSS} sensor relative to the vehicle center derived from the \acp{BB} in our \ac{UAV} footage. Additionally, potential inaccuracies in the \ac{RTK-GNSS} data could be caused by signal reflections or obstructions from surrounding structures. A more detailed analysis of intersections L and P, including possible sources of \ac{RTK-GNSS} inaccuracies, is provided later in this section. There could also be cumulative effects of small inaccuracies in \ac{BB} detection, object tracking, track stabilization, and georeferencing. Despite these potential sources of error, the high consistency between the two measurement methods strengthens confidence in the reliability of the Songdo Traffic dataset’s extracted trajectories and speed estimates.

\begin{table}[htbp]
  \centering
  \caption{Comparison of positional and speed differences (mean $\pm$ standard deviation) between drone-extracted trajectories and on-board \ac{RTK-GNSS} measurements of the \ac{AV}, along with aggregated trajectory statistics (length and duration).}\label{tab:av_error}
  \begin{tabular}{>{\raggedleft\arraybackslash}p{1.9cm}>{\raggedleft\arraybackslash}p{2.45cm}>{\raggedleft\arraybackslash}p{2.7cm}>{\raggedleft\arraybackslash}p{1.8cm}>{\raggedleft\arraybackslash}p{1.7cm}}
    \hline
    Intersection \linebreak label & Positional \linebreak deviation (m) & Speed \linebreak difference (km/h) & Trajectory \linebreak length (m) & Trajectory \linebreak duration (s) \\
    \hline
    E & $0.459 \pm 0.334$ & $-0.555 \pm 1.551$ & 336.13 & 26.63\\ 
    G & $0.241 \pm 0.073$ & $-0.606 \pm 0.646$ & 190.41 & 15.70\\ 
    J & $0.494 \pm 0.252$ & $0.867 \pm 1.314$ & 99.78 & 20.20\\ 
    K & $0.326 \pm 0.220$ & $-0.127 \pm 0.538$ & 266.16 & 21.81\\ 
    L & $1.879 \pm 0.330$ & $-0.103 \pm 0.928$ & 331.22 & 35.27\\ 
    M & $0.558 \pm 0.616$ & $-1.083 \pm 2.872$ & 334.37 & 49.88\\ 
    O & $0.647 \pm 0.138$ & $-0.816 \pm 1.182$ & 123.47 & 14.00\\ 
    P & $1.134 \pm 1.200$ & $-0.545 \pm 2.343$ & 151.97 & 38.71\\ 
    Q & $0.382 \pm 0.259$ & $-0.198 \pm 1.578$ & 322.18 & 33.20\\ 
    \hline
  \end{tabular}
\end{table}

To investigate the elevated positional discrepancies at intersections L and P, we generated locally rectified experimental orthophotos using enhanced georeferencing techniques. These orthophotos were intended to examine and potentially rule out any inaccuracies in the original orthophoto. A DJI MAVIC 3 drone was employed to capture high-resolution, overlapping images from an altitude of 75 m. To ensure higher geospatial accuracy, four \acp{GCP} were strategically placed at each corner of intersections L and P, each marked with a $3 \times 3$ meter QR code. These \acp{GCP} were precisely positioned using dual EMLID RS2+ \ac{RTK-GNSS} receivers operating in FIX mode, providing centimeter-level accuracy, in contrast to the less precise SINGLE mode. The experimental orthophotos were then generated using Agisoft Metashape Pro software\footnote{https://www.agisoft.com/}, leveraging the improved accuracy and increased number of \acp{GCP}.

The newly generated orthophotos, intended to enhance georeferencing accuracy, resulted in only marginal reductions in positional discrepancies. The recalculated positional deviations were $1.176 \pm 0.327$ m for intersection L and $0.816 \pm 0.539$ m for intersection P, with corresponding speed differences of $-0.359 \pm 0.966$ km/h and $-0.475 \pm 2.192$ km/h, respectively. Considering the superior track stabilization performance (see \autoref{fig:stabilo_campaign}) and orthophoto matching accuracy (see \autoref{tab:ortho_error}), these elevated discrepancies are unlikely to originate from errors in the original orthophoto. Instead, a closer examination of the urban environments surrounding intersections L and P suggests that signal multipath effects and occlusions are more plausible contributors. The presence of tall buildings around these intersections (see \autoref{fig:all_captured_AV_trajectories}) is typical of densely built urban areas like Songdo and is well-documented to cause the ``urban canyon effect'', where \ac{GNSS} accuracy is degraded due to signal reflections and obstructions from surrounding structures~\cite{groves2011shadow, kaplan2017understanding}.

\section{Conclusion}\label{sec:conclusion}

In this study, we introduced a comprehensive framework for extracting georeferenced vehicle trajectories from high-altitude drone imagery, addressing key challenges in urban traffic monitoring. Our approach integrates several novel contributions, including a tailored object detector optimized for high-altitude \ac{BEV} perspectives, a unique track stabilization method utilizing detected vehicle bounding boxes as exclusion masks during image registration, and a georeferencing strategy based on orthophoto and master frame that enhances consistent alignment across multiple drone viewpoints. Additionally, we proposed a robust vehicle dimension estimation method leveraging bounding box dimensions and azimuth information derived from vehicle dynamics. A novel traffic-oriented evaluation metric was introduced to complement existing metrics, directly assessing the accuracy of vehicle alignment after transformation. These contributions collectively advance the accuracy, efficiency, and scalability of drone-based traffic monitoring systems. Our methodology demonstrated robust detection performance despite challenges posed by high-altitude perspectives. The innovative use of track stabilization after object tracking effectively circumvented limitations inherent in traditional video stabilization methods, enabling reliable extraction of vehicle trajectories and kinematics. The incorporation of \acp{GCP} within the orthophoto further enhanced georeferencing accuracy, critical for detailed traffic dynamics analysis.

The application of our methodology in a multi-drone experiment over 20 intersections in the Songdo International Business District resulted in two high-quality datasets: the Songdo Traffic dataset, comprising nearly 690,000 unique vehicle trajectories, and the Songdo Vision dataset, featuring over 5,000 human-annotated frames for object detection. Notably, the Songdo Vision dataset fills a critical gap in existing datasets by providing detailed annotations for smaller vehicles, such as motorcycles, which are often challenging to detect from elevated \ac{BEV} perspectives. Publicly releasing these datasets along with the extraction pipeline's complete source code sets new standards for data quality, reproducibility, and scalability, promoting collaboration and accelerating research innovation. Comparisons between drone-derived trajectories and high-precision sensor data from an instrumented probe vehicle further validate the consistency and accuracy of our extraction methods. Although the dataset may exhibit minor systematic positional biases, these discrepancies do not significantly hinder its applicability for a broad range of traffic analyses, including traffic flow analysis, congestion detection, speed profiling, and lane-level behavior studies. The provided road and lane segmentations, derived directly from visual data, further enhance the usability of the data set.

While our study makes significant contributions, we acknowledge certain limitations, including the exclusion of smaller, non-motorized road users and the manual effort required for video and road segmentations. Additionally, the vehicle classification granularity was constrained by practical annotation constraints, grouping certain vehicle types together. Future work will focus on refining classification schemes by differentiating trucks into subclasses and including smaller, non-motorized road users, automating segmentation tasks, and improving trajectory accuracy through rotated bounding boxes. Enhancements in video stabilization, such as deep learning-based feature matching, could further refine alignment accuracy. Additionally, exploring transformer-based detection models and integrating motion forecasting modules may enhance tracking robustness in complex urban settings. By combining drone-based monitoring with advanced computer vision, our approach provides a scalable and cost-effective solution for urban traffic analysis, contributing to data-driven transportation planning and intelligent mobility systems.

\section*{CRediT authorship contribution statement}
\textbf{Robert Fonod}: Conceptualization, Methodology, Software, Validation, Formal analysis, Investigation, Data curation, Visualization, Writing – original draft, Writing – review \& editing, Visualization, Supervision, Project administration. \textbf{Haechan Cho}: Methodology, Software, Validation, Investigation, Data curation, Visualization. \textbf{Hwasoo Yeo}: Conceptualization, Resources, Funding acquisition, Supervision, Project administration, Funding acquisition. \textbf{Nikolas Geroliminis}: Conceptualization, Investigation, Resources, Writing – review \& editing, Supervision, Project administration, Funding acquisition.

\section*{Acknowledgments}

This work was supported by research grants from the Swiss National Science Foundation under NCCR Automation (No.~51NF40\_180545, N.G. and R.F.), Innosuisse (No.~101.645 IP-ENG, N.G. and R.F.), the National Research Foundation of Korea (NRF) grant funded by the Korean government (MSIT) (No.~2022R1A2C1012380, H.Y. and H.C.), and the Open Research Data (ORD) Program of the ETH Board (R.F.). The authors express their gratitude to Sohyeong Kim, Yura Tak, and Weijiang Xiong from EPFL for their invaluable assistance in the data-wrangling process, Artem Vasilev for his dedicated efforts in data annotation, Jasso Espadaler Clapés for his insightful contributions to vehicle dimension estimation, and Muhammad Ahmed for his support in creating the experimental orthophoto for validation purposes and his contributions to road segmentation. We also thank DroMii company for producing the primary orthophoto and \acf{SCIGC} for providing the on-board recorded trajectories of their autonomous vehicle. We also thank the research teams of Prof. Simon Oh (Korea University) and Prof. Minju Park (Hannam University) for their assistance during the data collection campaign, including the provision of drone equipment and student support.

\section*{Data availability}

The Songdo Traffic and Songdo Vision datasets are publicly available for download from~\cite{songdo_traffic_dataset} and~\cite{songdo_vision_dataset}, respectively, under a CC-BY-4.0 license. These datasets will also be listed on \href{https://open-traffic.epfl.ch}{https://open-traffic.epfl.ch}. The complete source code for the trajectory extraction framework, including the stabilization library and the associated optimization tool, is archived in~\cite{Fonod_Geo-trax_2025}, \cite{Fonod_Stabilo_2024}, and \cite{Fonod_Stabilo_Optimize_2025}, respectively, and is maintained on GitHub under the MIT license.

\appendix
\section{Vehicle dimension estimation steps and implementation details}\label{sec:appendix_dimension_estimation}

For ease of notation, the sequentially ordered \acp{BB} for a vehicle with \acl{ID} ``$\text{id}$'' in video $v$ are denoted as 
($x^\text{id}[k]$, $y^\text{id}[k]$, $w^\text{id}[k]$, $h^\text{id}[k]$), where $k$ ranges from $1$ to $N_v^\text{id}$. These values, un-normalized by the image dimensions $w_\mathcal{I}$ and $h_\mathcal{I}$, are used to calculate the vehicle’s length ($L^\text{id}$) and width ($W^\text{id}$) in meters through a structured five-step process:

\textbf{Step 1 - Visibility filtering:} 
This initial step filters out \acp{BB} that are not fully visible within the frame, focusing on fully observable vehicles. We define a visibility set $\mathcal{V}^\text{id}$ containing indices $k$ of \acp{BB} satisfying:
\begin{equation}\label{eq:visibility_filter}
    \begin{array}{cccc}
    x^\text{id}[k] - \dfrac{w^\text{id}[k]}{2} > \varepsilon, &  
    x^\text{id}[k] + \dfrac{w^\text{id}[k]}{2} < w_\mathcal{I} - (\varepsilon + 1), &
    y^\text{id}[k] - \dfrac{h^\text{id}[k]}{2} > \varepsilon, &  
    y^\text{id}[k] + \dfrac{h^\text{id}[k]}{2} < h_\mathcal{I} - (\varepsilon + 1),
    \end{array}
\end{equation}
where $\varepsilon > 0$ is a margin to keep \acp{BB} well within the frame's boundaries. The detection system ensures that \acp{BB} dimensions do not exceed the frame limits, with pixel coordinates ranging from $0$ to $w_\mathcal{I} - 1$ horizontally and $h_\mathcal{I} - 1$ vertically. The margin $\varepsilon$ helps filter out transient \acp{BB} generated as vehicles enter or exit the frame, as the detector can identify partially visible vehicles, leading to variable \ac{BB} sizes due to vehicles' gradual appearance or disappearance at the frame boundaries.

\textbf{Step 2 - Initial dimensions computation:}
Using the remaining \acp{BB} after Step 1, we curate the set of instantaneous vehicle lengths $\mathcal{L}^\text{id}$ and widths $\mathcal{W}^\text{id}$ according to the following simple rule:
\begin{equation}
\begin{aligned}
\mathcal{L}^\text{id} & 
\triangleq \left\{L^\text{id}_{\inf\mathcal{V}^\text{id}}, \ldots, L^\text{id}_{\sup\mathcal{V}^\text{id}} \right\}
= \left\{\max\left(w^\text{id}[k],\; h^\text{id}[k]\right) \,\mid\,  k \in \mathcal{V}^\text{id}\right\}, \\
\mathcal{W}^\text{id} & 
\triangleq \left\{W^\text{id}_{\inf\mathcal{V}^\text{id}}, \ldots, W^\text{id}_{\sup\mathcal{V}^\text{id}} \right\}
= \left\{\min\left(w^\text{id}[k],\; h^\text{id}[k]\right) \,\mid\, k \in \mathcal{V}^\text{id}\right\}.
\end{aligned}    
\end{equation}

\textbf{Step 3 - Azimuth-based filtering:}
We refine the sets $\mathcal{L}^\text{id}$ and $\mathcal{W}^\text{id}$ using the vehicle's azimuth estimate $\theta_k^\text{id} \in [0,\;2\pi)$, calculated asynchronously from trajectory points in the \ac{CS} of the reference frame $F_v^\text{ref}$, as follows:
\begin{equation}
    \theta^\text{id}_k = \text{atan2}\left(y^\text{id}[\delta^\text{id}_{k-1}] - y^\text{id}[\delta^\text{id}_{k}], \; x^\text{id}[\delta^\text{id}_{k}] - x^\text{id}[\delta^\text{id}_{k-1}]\right), \quad k = 1,2,\ldots,
\end{equation}
where $\delta_k^\text{id}>0$ is a discrete-time window growing recursively:
\begin{equation}
    \delta^\text{id}_k = \delta_{k-1}^\text{id} + 
    \inf_{\delta\in\mathbb{N}} \left\{\delta \; \Bigg | 
    \sqrt{\left(\Delta_x^\text{id}(\delta_{k-1}^\text{id},\delta)\right)^2 + 
    \left(\Delta_y^\text{id}(\delta_{k-1}^\text{id},\delta)\right)^2} 
    \geq r_\text{px}, \;  \delta_{k-1}^\text{id}+\delta \leq \sup(\mathcal{V}^\text{id}) \right\}.
\end{equation}
Here, $\Delta_x^\text{id}(\delta_{k-1}^\text{id},\delta)=x^\text{id}[\delta^\text{id}_{k-1}+\delta] - x^\text{id}[\delta_{k-1}^\text{id}]$ and $\Delta_y^\text{id}(\delta_{k-1}^\text{id},\delta)=y^\text{id}[\delta^\text{id}_{k-1}+\delta] - y^\text{id}[\delta_{k-1}^\text{id}]$ represent the distances traveled in the $x$ and $y$ directions, respectively, for a step $\delta$. The initial index
$\delta_0^\text{id}=\inf (\mathcal{V}^\text{id})$ corresponds to the time index of the 1st visible \ac{BB}, and $r_\text{px}>0$ denotes the radius in pixels regulates the frequency of azimuth computations. 

To provide a physical interpretation of $r_\text{px}$, we define $r_\text{px} = r_\text{m} / \text{GSD}$, where \ac{GSD} is the ground distance per pixel (meters/pixel) and $r_{\text{m}}>0$ is the minimum distance (meters) required for azimuth computation. This ensures azimuth calculations rely on significant vehicle movements, avoiding inaccuracies during stationary periods. \ac{GSD} can be calculated using camera specifications and flight altitude or derived from the orthophoto.

If at least one azimuth is available $\theta_k^\text{id}$ ($k\geq1)$, we further refine $\mathcal{L}^\text{id}$ and $\mathcal{W}^\text{id}$ using all available $\theta_k^\text{id}$ as follows:
\begin{equation}\label{eq:azimuth_filter}
\begin{aligned}
    \mathcal{L}^\text{id} &= \left\{L^\text{id}_i \, \mid \, L_i^\text{id}\in\mathcal{L}^\text{id}, \;  i\in [\delta_{k-1}^\text{id}, \, \delta_{k}^\text{id}) \; \text{if} \; |\theta_k - \phi_j| \leq \bar\theta, \; \text{for any} \; \phi_j \in \Phi\right\},\\
    \mathcal{W}^\text{id} &= \left\{W^\text{id}_i \, \mid \, W_i^\text{id}\in\mathcal{W}^\text{id}, \; i\in [\delta_{k-1}^\text{id}, \, \delta_{k}^\text{id}) \; \text{if} \; |\theta_k - \phi_j| \leq \bar\theta, \; \text{for any} \; \phi_j \in \Phi\right\},\\
\end{aligned}
\end{equation}
where $\Phi=\left\{0,\, \pi/2,\, \pi,\, 3\pi/2, \; 2\pi\right\}$ is a set of cardial directions and $\bar\theta>0$ is the maximum azimuth deviation allowed. The parameter $\bar{\theta}$ accounts for vehicle orientations that are non-parallel to the image axes, with a smaller $\bar{\theta}$  indicating more stringent filtering, thus ensuring only representative vehicle dimensions are considered.

When no azimuth data is available (e.g., stationary or parked vehicles), we filter \acp{BB} exhibiting square-like proportions, as these often reflect non-parallel parking or partial occlusions that compromise accuracy. This is achieved by a ratio-based criterion:
\begin{equation}\label{eq:ratio_filter}
    \mathcal{L}^\text{id} = \left\{L_i^\text{id} \, \mid \, L_i^\text{id}\in\mathcal{L}^\text{id}, \; \tfrac{L_i^\text{id}}{W_i^\text{id}} \geq  \kappa_c\right\}, \quad 
    \mathcal{W}^\text{id} = \left\{W_i^\text{id} \, \mid \, W_i^\text{id}\in\mathcal{W}^\text{id}, \; \tfrac{L_i^\text{id}}{W_i^\text{id}} \geq  \kappa_c \right\}.
\end{equation}
Here, $\kappa_c>1$ is a class-specific threshold for the degree of non-squareness, with ratios $L^\text{id}_i/W^\text{id}_i$ approaching 1 indicating square-like dimensions, which are atypical in the Songdo footage and are therefore flagged as unreliable.

\textbf{Step 4 - Final dimension computation:}
After implementing Steps 1 to 3, if the sets  $\mathcal{L}^\text{id}$ and  $\mathcal{W}^\text{id}$ are not empty, the unique vehicle dimensions in pixels are determined as follows:
\begin{equation}
     L^\text{id}_{\text{px}} = \text{Q}_\text{lower}\left(\mathcal{L}^\text{id}\right), \quad 
     W^\text{id}_{\text{px}} = \text{Q}_\text{lower}\left(\mathcal{W}^\text{id}\right).
\end{equation}
Here, $\text{Q}_\text{lower}(\cdot)$ computes the first quartile, indicating the value below which 25\% of observations fall. This measure is preferred due to the leniency of $\bar\theta$, resulting in dimensions that meet or exceed the \acl{GT}, while mean or median methods often yield overly conservative estimates. Minimum values are avoided to prevent errors from partial occlusions.

\textbf{Step 5 - Real-world conversion:}
In this final step, the vehicle dimensions $L^\text{id}_\text{px}$ and $W^\text{id}_\text{px}$ are converted from pixels to meters by placing a hypothetical \ac{BB} at the center of the video frame. Centering the \ac{BB} leverages the higher precision of the orthophoto, enhanced by strategically placed \acp{GCP} at intersection edges. The conversion uses a three-point \ac{BB} decomposition defined as:
\begin{equation}
p^\text{id}_{\text{px}1} = \tfrac{1}{2}(w_\mathcal{I},\, h_\mathcal{I}), \quad
p^\text{id}_{\text{px}2} = \tfrac{1}{2}(w_\mathcal{I},\, h_\mathcal{I}+W_\text{px}^\text{id}), \quad
p^\text{id}_{\text{px}3} = \tfrac{1}{2}(w_\mathcal{I}+L_\text{px}^\text{id},\, h_\mathcal{I}).
\end{equation}
These points are transformed to local Cartesian coordinates using $H_v^{\text{ref}\rightarrow\text{ortho}}$ and $T_i^{\text{ortho}\rightarrow\text{world}}$ (see \autoref{sec:georeferencing}) yielding real-world coordinates $p^\text{id}_{1}$, $p^\text{id}_{2}$, and $p^\text{id}_{3}$. Finally, the real-world vehicle dimensions $L^\text{id}$ and $W^\text{id}$ are determined using appropriate geometric relations:
\begin{equation}
L^\text{id} = 2 \|p^\text{id}_{3} - p^\text{id}_{1}\|, \quad 
W^\text{id} = 2 \|p^\text{id}_{2} - p^\text{id}_{1}\|,
\end{equation}
where $\|\cdot\|$ is the Euclidean norm of a vector, effectively measuring the physical distances between the points.

All steps outlined above should employ the unstabilized \acp{BB} from $\mathcal{T}_v$, with the sole exception of the azimuth angle calculation, $\theta_k^\text{id}$, which should be based on the stabilized vehicle trajectories contained in $\mathcal{T}^\text{ref}_v$.

\subsubsection*{Implementation details}

In our experiment, we set $\varepsilon=4$, $\bar\theta = 15^\circ$, $r_\text{m} = 1.25$, and derived $\text{GSD} = 0.02725$ based on camera specifications and an average altitude of 145 m, resulting in $r_\text{px} \approx 45.87$. The calibration of $r_\text{m}$ is essential for effective azimuth-based filtering and is primarily influenced by the quality of the track stabilization process. $r_\text{m}$ must be sufficiently large to avoid triggering new azimuth computations from drone movement artifacts while remaining small enough to capture azimuth updates during brief intervals of vehicle alignment with the image axes, particularly during road turns. This calibration is essential for vehicles on roads misaligned with the image axes due to infrastructure layout or drone orientation. Accurate dimension estimation can be achieved if, during brief moments of alignment, an azimuth satisfying $\bar{\theta}$ is computed.

The class-dependent threshold parameters $\kappa_c$, used in \eqref{eq:ratio_filter} to filter dimensions of stationary vehicles, were derived by applying $Q_\text{lower}$ to the empirical distributions of class-filtered $L/W$ ratios from moving vehicles. Extensive trajectory data analysis yielded specific thresholds for each vehicle category: $\kappa_{\text{car}} = 1.83$, $\kappa_{\text{bus}} = 2.85$, $\kappa_{\text{truck}} = 1.7$, and $\kappa_{\text{motorcycle}} = 1.8$. To obtain reliable estimates of $L$ and $W$, we enforced stricter azimuth angle constraints ($\bar\theta = 5^\circ$) and set $\kappa_c$ to $\infty$ to retain \acp{BB} only from vehicles perfectly aligned with the reference frame $F_v^\text{ref}$.

\section{Vehicle kinematics computation and tuning}\label{sec:appendix_kinematics}

The raw speed of a vehicle with identifier $\text{id}$ at time $k$ is computed as the Euclidean distance between two consecutive interpolated trajectory points $(x_L^\text{id}[k], \, y_L^\text{id}[k])$ and $(x_L^\text{id}[k-1], \, y_L^\text{id}[k-1])$ in the local coordinate system, divided by the time step $\Delta t = t_k - t_{k-1}$, which is equal to $\text{FPS}^{-1}$ for the processed video $v$.

To smooth the speed data $v^\text{id}[k]$, we apply a Gaussian filter through a discrete convolution with a 1D Gaussian kernel:
\begin{equation}    
\label{eq:gaussian_smoothing}
\tilde{v}^\text{id}[k] = \sum_{i=-M}^{M} v^\text{id}[r(k+i)] \cdot g_\sigma[i], \qquad k\in\{2,\ldots,N_v^\text{id}\},
\end{equation}
where $g_\sigma[i] = \frac{1}{\sqrt{2\pi}\sigma} \exp\left(-\frac{i^2}{2\sigma^2}\right)$, $\sigma>0$ is the standard deviation, and $M$ is set as $3\sigma$ to encompass the significant range of the Gaussian function $g_\sigma[i]$. The reflect function $r(j)$ handles boundary conditions by mirroring the data at the edges:
\begin{equation}
r(j) =
j + 2(N_v^\text{id} - j) \cdot \mathbf{1}(j > N_v^\text{id}) - 2(j - 1) \cdot \mathbf{1}(j < 1),
\end{equation}
where the indicator function $\mathbf{1}(\cdot)$ effectively reflects indices when $j$ lies outside the range $[1, \, N_v^\text{id}]$.

The acceleration profile $a^\text{id}[k]$ is computed by taking the backward difference of consecutive smoothed speed values: $a^\text{id}[k] = (\tilde{v}^\text{id}[k] - \tilde{v}^\text{id}[k-1])/\Delta t$, for $k\in{3,\ldots,N_v^\text{id}}$. Only indices that satisfy $k\in\mathcal{V}^\text{id}$ retain $\tilde v^\text{id}[k]$ and $a^\text{id}[k]$ in the dataset, where $\mathcal{V}^\text{id}$ is the visibility set defined in \eqref{eq:visibility_filter}. This ensures that kinematic computations are unaffected by changes in \acp{BB} as vehicles enter or exit the camera's \ac{FoV}.

The standard deviation $\sigma$ of the Gaussian kernel is the only tuning parameter for smoothing speed data. To determine its optimal value, we used \ac{AV} data to assess the relationship between different $\sigma$ values and the speed estimation errors $\Delta v[k_1]$ from \eqref{eq:speed_error}. Specifically, we computed $\Delta v[k_1]$, applied the absolute value, and averaged these absolute errors. This analysis showed that the mean optimal $\sigma$ across all intersections is 14.5, while the weighted average, accounting for the lengths of observed \ac{AV} trajectories, is 14.49.  Additionally, the optimal $\sigma$ that minimizes the error variance was found to be 14.17. Therefore, we set $\sigma$ to 14 frames, equivalent to 0.467 seconds at 29.97 \ac{FPS}, to balance smoothing and detail retention. Notably, the Savitzky-Golay filter~\cite{savitzky1964smoothing} produced comparable results to our Gaussian filter approach.

\section{Post-processing validation, corrections, and potential systematic biases}\label{sec:appendix_validation}

In addition to the manual data wrangling steps detailed in \autoref{sec:data_wrangling}, automated post-processing checks were conducted to ensure data consistency and correct any remaining video or flight-log anomalies in the Songdo Traffic dataset. These analyses examined flight logs and extracted trajectories to assess drone hovering stability (location and altitude), abrupt camera parameter variations (focal length, ISO, shutter speed, f-number, and color temperature), timestamp continuity, and outlier kinematic values.

Approximately 2\% of segmented videos exhibited anomalies such as significant drone movements or zoom changes undetected during data wrangling, timestamp irregularities, or data recording gaps. To correct these, affected videos and flight logs were re-cut by trimming the beginning or end, or in some cases, splitting the video into two segments to remove anomalies that occurred mid-sequence. The latter approach, applied to approximately 1\% of segmented videos, inevitably resulted in trajectory fragmentation and vehicle \ac{ID} resets. These corrections ensured that the final dataset is free from major recording inconsistencies, preserving the integrity of extracted trajectories and associated metadata.

As discussed in \autoref{sec:trajectory_georeferencing}, the use of the master frame as an intermediary coordinate system implies that any potential inaccuracies in georeferencing would manifest as systematic biases consistently affecting all trajectories at a given intersection. Such systematic biases can potentially be corrected post hoc, for instance, through map-matching techniques or recalibration of the orthophoto mapping parameters. The retention of vehicle trajectories in orthophoto cut-out coordinates, along with the release of these cut-outs, facilitates such corrections if required. Additionally, our object detector exhibited high localization accuracy, with \ac{BB} center deviations averaging 2.21 pixels from human annotations. This reinforces the reliability of the initial detection step, which serves as the foundation for trajectory extraction, independent of any downstream georeferencing or stabilization effects.

\section{Known dataset artifacts and limitations}\label{sec:appendix_artifacts}

In line with best practices for dataset transparency and reproducibility, this appendix summarizes known artifacts, limitations, and potential sources of error identified in the Songdo Traffic dataset \cite{songdo_traffic_dataset}.

\textbf{1) Trajectory fragmentation:} As discussed in \autoref{sec:data_annotation} and \autoref{sec:object_tracking}, trajectories may be fragmented for certain motorcycle instances over complex underrepresented road infrastructures (e.g., pedestrian crossings, bicycle lanes, traffic signals), see \autoref{fig:annotations} (right). Similarly, some truck variants, such as exceptionally large or specialized trucks that are extremely underrepresented in both the training data and the trajectory dataset, may yield split trajectories or abnormal kinematic profiles, owing to the object detector’s challenges in consistently providing accurate \ac{BB} detections. Users are advised to exercise caution when interpreting trajectory data associated with these specific vehicle instances. Additional fragmentations resulted from necessary mid-hovering video splits during data wrangling (cf.~\autoref{sec:data_wrangling}) and post-processing corrections (cf.~\ref{sec:appendix_validation}) when drones experienced technical issues, naturally creating divided trajectories.

\textbf{2) Vehicle ID ambiguities:} The largest \texttt{Vehicle\_ID} in a given \ac{CSV} file does not necessarily represent the actual number of unique vehicles observed. This is due to (a) the aforementioned trajectory fragmentation, where a single vehicle's trajectory is split into multiple segments, each potentially being assigned a different \texttt{Vehicle\_ID}, (b) the removal of trajectories with 15 or fewer points after the tracking algorithm assigned unique vehicle IDs (cf.~\autoref{sec:songdo_dataset_creation}), and (c) instances of the same vehicle may be observed simultaneously by multiple drones, leading to distinct \texttt{Vehicle\_ID}s. While flight plans were carefully designed to avoid overlapping observations of the same intersection by multiple drones, minor synchronization deviations may have led to occasional overlaps. Nevertheless, the recorded flight logs indicate that all drone flights commenced as planned, with 31 flights (out of 400) concluding slightly after their planned time window, with an average delay of 19.04 ± 5.8 seconds.

\textbf{3) Kinematic estimation limitations:} The speed and acceleration variables in the dataset are derived from raw trajectory data and are subject to inaccuracies due to imperfect and dynamically varying \ac{BB} detections, particularly from shadows and the expansion or contraction of \acp{BB} as vehicles enter or exit the video frame. While the latter effect is indicated by the visibility flag, shadow-induced variations are not explicitly annotated. Additionally, suboptimal track stabilization and applied interpolation and smoothing (cf. \autoref{sec:kinematics}) may further affect accuracy. Consequently, these kinematic estimates are intended primarily for preliminary filtering rather than precise dynamic analyses.

\textbf{4) Vehicle dimension estimation reliability:} As discussed in \autoref{sec:vehicle_dimensions}, vehicle dimension estimates may be unreliable, especially when a vehicle is stationary or not moving parallel to one of the image axes. Additionally, these estimates may systematically overstate actual dimensions due to \acp{BB} capturing protruding parts or shadows.

\textbf{5) Road section and lane assignment inaccuracies:} Due to perspective effects, vehicles with significant heights, such as trucks or buses, may have their centers fall outside the assigned section or lane in the orthophoto, leading to potential mislabeling. This limitation should be considered when analyzing lane-level traffic flows for taller vehicles.

\textbf{6) Occasional pedestrian pair misclassifications:} In rare cases, two pedestrians walking side by side across a crosswalk may be momentarily misclassified as a motorcycle. Such occurrences are extremely brief and, if present, are likely filtered out by the short trajectory filter discussed earlier (cf.~\autoref{sec:songdo_dataset_creation}).

\end{document}